\documentclass[review,3p,authoryear]{elsarticle}

\usepackage[multiple]{footmisc}
\usepackage{enumitem}
\usepackage{lineno}
\usepackage{amssymb}
\usepackage{sidecap}
\usepackage{graphicx}
\usepackage{caption}
\usepackage{subcaption}
\usepackage{multirow}
\usepackage{wrapfig}
\usepackage{lipsum}
\usepackage{nicefrac}
\usepackage{booktabs,tabularx}
\usepackage{adjustbox}
\usepackage{floatrow}
\usepackage{float}
\restylefloat{table}
\newcolumntype{x}{>{\raggedright\arraybackslash}X}
\usepackage{bbding}
\usepackage{amsmath}
\usepackage{romannum}
\usepackage[colorlinks=true,
            linkcolor=magenta,
            urlcolor=gray,
            citecolor=cyan]{hyperref}
\usepackage{slashbox}
\usepackage{arydshln}
\newfloatcommand{capbtabbox}{table}[][\FBwidth]
\usepackage{blindtext}
\usepackage{enumitem}
\usepackage{algpseudocode}
\usepackage{xcolor}
\usepackage{adjustbox}
\usepackage{array}
\usepackage[linesnumbered,ruled,vlined]{algorithm2e}
\usepackage[final]{pdfpages}
\usepackage{multicol}   
\usepackage{xcolor}
\usepackage[tikz]{bclogo}
\usepackage[framemethod=tikz]{mdframed}
\usepackage{lipsum}
\usepackage[many]{tcolorbox}
\usepackage{ctable}

\definecolor{bgblue}{RGB}{245,243,253}
\definecolor{ttblue}{RGB}{91,194,224}

\newcounter{includepdfpage}
\newcounter{currentpagecounter}

\newcolumntype{t}{>{\hsize=0.25\hsize}x}
\newcolumntype{s}{>{\hsize=0.75\hsize}x}
\newcolumntype{X}{>{\hsize=1.5\hsize}x}
\newcolumntype{R}[2]{%
    >{\adjustbox{angle=#1,lap=\width-(#2)}\bgroup}%
    l%
    <{\egroup}%
}

\newcommand{\siunit}[2]{$#1\,\mathrm{#2}$}
\newcommand{\tabincell}[2]{\begin{tabular}{@{}#1@{}}#2\end{tabular}}  
\newcommand\crule[3][black]{\textcolor[HTML]{#1}{\rule{#2}{#3}}}

\usepackage{xparse}
\NewDocumentCommand{\rot}{O{90} O{1em} m}{\makebox[#2][l]{\rotatebox{#1}{#3}}}%

\modulolinenumbers[5]
\newpageafter{author}

\newtcolorbox{myboxii}[1][]{
  breakable,
  freelance,
  title=#1,
  colback=white,
  colbacktitle=white,
  coltitle=black,
  fonttitle=\bfseries,
  bottomrule=0pt,
  boxrule=0pt,
  colframe=white,
  overlay unbroken and first={
  \draw[red!75!black,line width=3pt]
    ([xshift=5pt]frame.north west) -- 
    (frame.north west) -- 
    (frame.south west);
  \draw[red!75!black,line width=3pt]
    ([xshift=-5pt]frame.north east) -- 
    (frame.north east) -- 
    (frame.south east);
  },
  overlay unbroken app={
  \draw[red!75!black,line width=3pt,line cap=rect]
    (frame.south west) -- 
    ([xshift=5pt]frame.south west);
  \draw[red!75!black,line width=3pt,line cap=rect]
    (frame.south east) -- 
    ([xshift=-5pt]frame.south east);
  },
  overlay middle and last={
  \draw[red!75!black,line width=3pt]
    (frame.north west) -- 
    (frame.south west);
  \draw[red!75!black,line width=3pt]
    (frame.north east) -- 
    (frame.south east);
  },
  overlay last app={
  \draw[red!75!black,line width=3pt,line cap=rect]
    (frame.south west) --
    ([xshift=5pt]frame.south west);
  \draw[red!75!black,line width=3pt,line cap=rect]
    (frame.south east) --
    ([xshift=-5pt]frame.south east);
  },
}

\begin{document}

\begin{frontmatter}

%\title{Elsevier \LaTeX\ template\tnoteref{mytitlenote}}

\title{\textsc{VerSe}: A Vertebrae Labelling and Segmentation Benchmark\\ for Multi-detector CT Images}

\author[tum,msb,neuro]{Anjany Sekuboyina}
\ead{anjany.sekuboyina@tum.de}
\author[tum,neuro]{Malek E. Husseini}
\author[tum,neuro]{Amirhossein Bayat}
\author[neuro]{Maximilian L\"offler}
\author[neuro]{Hans Liebl}
\author[tum]{Hongwei Li}
\author[tum]{Giles Tetteh}
\author[ibmi]{Jan Kuka{\v{c}}ka}

% ------------ participants
\author[payer1]{Christian Payer}
\author[payer2]{Darko \v{S}tern}
\author[payer3]{Martin Urschler}
\author[iflytek]{Maodong Chen}
\author[iflytek]{Dalong Cheng}
\author[nlessmann]{Nikolas Lessmann}
\author[huyujin1]{Yujin Hu}
\author[huyujin2]{Tianfu Wang}
\author[yangd05]{Dong Yang}
\author[yangd05]{Daguang Xu}
\author[zib]{Felix Ambellan}
\author[zib]{Tamaz Amiranashvili}
\author[1000shapes]{Moritz Ehlke}
\author[1000shapes]{Hans Lamecker}
\author[1000shapes]{Sebastian Lehnert}
\author[1000shapes]{Marilia Lirio}
\author[1000shapes]{Nicol\'{a}s P\'{e}rez de Olaguer}
\author[1000shapes]{Heiko Ramm}
\author[zib]{Manish Sahu}
\author[zib]{Alexander Tack}
\author[zib]{Stefan Zachow}
\author[alibaba]{Tao Jiang}
\author[alibaba]{Xinjun Ma}
\author[christoph]{Christoph Angerman}
\author[init1]{Xin Wang}
% \author[init2]{Qingyue Wei}
\author[brown2]{Kevin Brown}
% \author[brown1]{Matthias Wolf}
\author[lrde]{Alexandre Kirszenberg}
\author[lrde]{\'{E}lodie Puybareau}
% verse20 authors
\author[deepreasoningai]{Di Chen}
\author[deepreasoningai]{Yiwei Bai}
\author[deepreasoningai]{Brandon H. Rapazzo}
\author[sitp]{Timyoas Yeah}
%\author[sitp]{Xavier Hou}
\author[jdlu1]{Amber Zhang}
\author[jdlu2]{Shangliang Xu}
\author[carpediem1]{Feng Hou}
\author[carpediem2]{Zhiqiang He}
\author[aply]{Chan Zeng}
\author[fakereal1,fakereal2]{Zheng Xiangshang}
\author[fakereal1]{Xu Liming}
\author[superpod]{Tucker J. Netherton}
\author[superpod]{Raymond P. Mumme}
\author[superpod]{Laurence E. Court}
\author[poly1]{Zixun Huang}
\author[poly3]{Chenhang He}
\author[poly1]{Li-Wen Wang}
\author[poly2]{Sai Ho Ling}
\author[lrde]{L\^{e} Duy Hu\`{y}nh}
\author[lrde]{Nicolas Boutry}
\author[ubmi]{Roman Jakubicek}
\author[ubmi]{Jiri Chmelik}
\author[htic1,htic2]{Supriti Mulay}
\author[htic1,htic2]{Mohanasankar Sivaprakasam}
\author[tum]{Johannes C. Paetzold}
\author[tum]{Suprosanna Shit}
\author[tum]{Ivan Ezhov}

% ------------
\author[neuro]{Benedikt Wiestler}
\author[imperial]{Ben Glocker}
\author[neuro]{Alexander Valentinitsch}
\author[fmi]{Markus Rempfler}
\author[tum,uzh]{Bj\"orn H. Menze\tnoteref{mytitlenote}}
\author[neuro]{Jan S. Kirschke\tnoteref{mytitlenote}}

\tnotetext[mytitlenote]{BM and JSK are supervising authors}

% main addresses
\address[tum]{Department of Informatics, Technical University of Munich, Germany.}
\address[msb]{Munich School of BioEngineering, Technical University of Munich, Germany.}
\address[neuro]{Department of Neuroradiology, Klinikum Rechts der Isar, Germany.}
\address[uzh]{Department for Quantitative Biomedicine, University of Zurich, Switzerland.}
\address[fmi]{Friedrich Miescher Institute for Biomedical Engineering, Switzerland}
\address[ibmi]{Institute of Biological and Medical Imaging, Helmholtz Zentrum M\"unchen, Germany}
\address[imperial]{Department of Computing, Imperial College London, UK}
% participant affiliations
\address[payer1]{Institute of Computer Graphics and Vision, Graz University of Technology, Austria}
\address[payer2]{Gottfried Schatz Research Center: Biophysics, Medical University of Graz, Austria}
\address[payer3]{School of Computer Science, The University of Auckland, New Zealand}
\address[iflytek]{Computer Vision Group, iFLYTEK Research South China, China}
\address[nlessmann]{Department of Radiology and Nuclear Medicine, Radboud University Medical Center Nijmegen, The Netherlands}
\address[huyujin1]{Shenzhen Research Institute of Big Data, China}
\address[huyujin2]{School of Biomedical Engineering, Health Science Center, Shenzhen University, China}
\address[yangd05]{NVIDIA Corporation, USA}
\address[zib]{Zuse Institute Berlin, Germany}
\address[1000shapes]{1000shapes GmbH, Berlin, Germany}
\address[alibaba]{Damo Academy, Alibaba Group, China}
\address[christoph]{Department of Mathematics, University of Innsbruck, Austria}
\address[init1]{Department of Electronic Engineering, Fudan University, China}
\address[init2]{Department of Radiology, University of North Carolina at Chapel Hill, USA}
% \address[brown1]{Siemens Healthineers, USA}
\address[brown2]{New York University, USA}
\address[lrde]{EPITA Research and Development Laboratory (LRDE), France}
% verse20 affiliations
\address[deepreasoningai]{Deep Reasoning AI Inc, USA}
\address[jdlu1]{Technical University of Munich, Germany}
\address[jdlu2]{East China Normal University}
\address[sitp]{Chinese Academy of Sciences, China}
\address[carpediem1]{Institute of Computing Technology, Chinese Academy of Sciences, China}
\address[carpediem1]{Lenovo Group, China}
\address[aply]{Ping An Technologies, China}
\address[fakereal1]{College of Computer Science and Technology, Zhejiang University, China}
\address[fakereal2]{Real Doctor AI Research Centre, Zhejiang University, China}
\address[superpod]{The University of Texas MD Anderson Cancer Center, USA}
\address[poly1]{Department of Electronic and Information Engineering, The Hong Kong Polytechnic University, China}
\address[poly3]{Department of Computing, The Hong Kong Polytechnic University, China}
\address[poly2]{The School of Biomedical Engineering, University of Technology Sydney, Australia}
\address[ubmi]{Department of Biomedical Engineering, Brno University of Technology, Czech Republic}
\address[htic1]{Indian Institute of Technology Madras, India}
\address[htic2]{Healthcare Technology Innovation Centre, India\vspace{4em}}

\begin{abstract}

Vertebral labelling and segmentation are two fundamental tasks in an automated spine processing pipeline. Reliable and accurate processing of spine images is expected to benefit clinical decision support systems for diagnosis, surgery planning, and population-based analysis of spine and bone health. However, designing automated algorithms for spine processing is challenging predominantly due to considerable variations in anatomy and acquisition protocols and due to a severe shortage of publicly available data. Addressing these limitations, the \emph{Large Scale Vertebrae Segmentation Challenge} (\textsc{VerSe}) was organised in conjunction with the International Conference on Medical Image Computing and Computer Assisted Intervention (MICCAI) in 2019 and 2020, with a call for algorithms tackling the labelling and segmentation of vertebrae. Two datasets containing a total of 374 multi-detector CT scans from 355 patients were prepared and 4505 vertebrae have individually been annotated at voxel level by a human-machine hybrid algorithm (\url{https://osf.io/nqjyw/}, \url{https://osf.io/t98fz/}). A total of 25 algorithms were benchmarked on these datasets. In this work, we present the results of this evaluation and further investigate the performance variation at the vertebra level, scan level, and different fields of view. We also evaluate the generalisability of the approaches to an implicit domain shift in data by evaluating the top-performing algorithms of one challenge iteration on data from the other iteration. The principal takeaway from \textsc{VerSe}: the performance of an algorithm in labelling and segmenting a spine scan hinges on its ability to correctly identify vertebrae in cases of rare anatomical variations. The \textsc{VerSe} content and code can be accessed at: \url{https://github.com/anjany/verse}.

\end{abstract}

% \begin{keyword}
% spine, vertebrae, segmentation, labelling, computed tomography
% \end{keyword}

\end{frontmatter}

\begin{myboxii}[Note:]
This is a pre-print of the journal article published in \emph{Medical Image Analysis}. If you wish to cite this work, please cite its journal version available here:  \url{https://doi.org/10.1016/j.media.2021.102166}. This work is available under CC-BY-NC-ND license. 
\end{myboxii}

\section{Introduction}
The spine is an important part of the musculoskeletal system, sustaining and supporting the body and its organ structure while playing a major role in our mobility and load transfer. It also shields the spinal cord from injuries and mechanical shocks due to impacts. Efforts towards quantification and understanding of the biomechanics of the human spine include quantitative imaging \citep{loeffler2019}, finite element modelling (FEM) of the vertebrae \citep{anitha2020}, alignment analysis \citep{laouissat2018} of the spine and complex biomechanical models \citep{oxland2016}. Biomechanical alterations can cause severe pain and disability in the short term, and can demonstrate worse consequences in the long term, e.g. osteoporosis leads to an 8-fold higher mortality rate \citep{cauley2000}. In spite of their criticality, spinal pathologies are often under-diagnosed \citep{howlett2020,mueller2008,williams2009}. This calls for computer-aided assistance for efficient and early detection of such pathologies, enabling prevention or effective treatment.

\emph{Vertebral labelling} and \emph{vertebral segmentation} are two fundamental tasks in understanding spine image data. Labelled and segmented spines have diagnostic implications for detecting and grading vertebral fractures, estimating the spinal curve, and recognising spinal deformities such as scoliosis and kyphosis. From a non-diagnostic perspective, these tasks enable efficient biomechanical modelling, FEM analysis, and surgical planning for metal insertions. Vertebral labelling can be performed quickly by a medical expert, on smaller datasets, as it follows clear rules \citep{wigh1980}. But, manually segmenting them is unfeasible owing to the time required for annotating large structures (e.g. 25 objects of interest with a size of $\sim10^4$ voxels each). Moreover, the complex morphology of the vertebra’s posterior elements combined with lower scan resolutions prevents a consistent and accurate manual delineation. Automating these tasks also involves multiple challenges: highly varying fields of view (FoV) across datasets (unlike brain images), large scan sizes, highly correlating shapes of adjacent vertebrae, scan noise, different scanner settings, and multiple anomalies or pathologies being present. For example, the presence of vertebral fractures, metal implants, cement, or transitional vertebrae should be considered during algorithm design. Fig. \ref{fig:datasample} illustrates this diversity using the scans included in the Large Scale Vertebrae Segmentation Challenge (\textsc{VerSe}). 
\subsection{Terminology}
\label{subsec:terminology}
In this section, we introduce three spine-processing terms frequently used in this work: \emph{localisation}, \emph{labelling}, \emph{segmentation}. As used in the rest of the work: \emph{Localisation} is the task of detecting a 3D coordinate on the vertebra and \emph{labelling} is the task of detecting a 3D coordinate on the vertebra as well as identifying the vertebrae. Specifically, labelling superscribes localisation by assigning a 3D coordinate as well as a class to the vertebra (C1-C6, T1-T13, L1-L5, as well as T13 and L6). Unless mentioned otherwise, spine \emph{segmentation} is a voxel-level, multi-class annotation problem, where in each vertebra level has a defined class label (e.g. C1$\rightarrow$1, C2$\rightarrow$2, T1$\rightarrow$8 etc.). It can now be seen that once a vertebra is segmented, its labelling and localisation is implied.

\subsection{Prior Work}
\label{subsec:litrev}
Spine image analysis has received subsistence attention from the medical imaging community over the years. Although computed tomography (CT) is a preferred modality for studying the ‘bone’ part of a spine due to high bone-to-soft-tissue contrast, there are several prior works on the tasks of labelling and segmenting the spine using multiple modalities in addition to CT such as magnetic resonance imaging (MRI), and 2D radiographs. There are works tackling segmentation (most of which inherently include vertebral labelling), and those tackling labelling specifically from a landmark-detection perspective.

\subsubsection{Vertebral Segmentation}
Traditionally, vertebral segmentation was performed using model-based approaches, which loosely involve fitting a shape prior to the spine and deforming it so that it fits the given spine. The incorporated shape priors range from geometric models \citep{vstern2011parametric, ibragimov2014shape,ibragimov2017segmentation}, deformed with Markov random fields (MRF) \citep{kadoury2011automatic,kadoury2013spine}, statistical shape models \citep{rasoulian2013lumbar,pereanez2015accurate,castro2015statistical}, and active contours \citep{leventon2002statistical,athertya2016automatic}. There are also intensity-based approaches such as level sets \citep{lim2014robust} and \emph{a priori} variational intensity models \citep{hammernik2015vertebrae}. Landmark frameworks tackling fully automated vertebral labelling and segmentation from a shape-modelling perspective exist \citep{klinder2009automated,korez2015framework}.       

With the increased adoption of machine learning in image analysis, works incorporating significant data-based learning components have been proposed. \cite{suzani2015deep} propose using a multi-layer perceptron (MLP) to detect the vertebral bodies and employ deformable registration for segmentation. Similar in philosophy, \cite{chu2015fully} propose random forest regression for locating and identifying the vertebrae followed by segmentation performed using random forest classification at a voxel level. Incorporating deep learning, \cite{korez2016model} learn vertebral appearances using 3D convolutional neural networks (CNN) and predict probability maps, which are then used to guide the boundaries of a deformable vertebral model. 

The recent advent of deep learning in image analysis and increased computing capabilities have led to works wherein deformable shape modelling and/or vertebral identification was replaced by data-driven learning of the vertebral shape using deep neural networks. \cite{sekuboyina2017attention} perform a patch-based binary segmentation of the spine using a U-Net \citep{ronneberger2015} (or a fully convolutional network, FCN) followed by denoising the spine mask using a low-resolution heat map. \cite{sekuboyina2017localisation} propose two neural networks for vertebral segmentation in the lumbar region. First, an MLP learns to regress the localisation of the lumbar region, following which a U-Net performs multi-class segmentation. Improving on this, \cite{janssens2018fully} replace the MLP with a CNN, thus performing multi-class segmentation of lumbar vertebrae with two successive CNNs. \cite{lessmann2018iterative} propose a two-staged iterative approach, wherein the first stage involves identifying and segmenting one vertebra after another at a lower resolution, followed by a second CNN to refine the lower-resolution masks. Building on this, \cite{lessmann19} proposed a single-stage FCN which iteratively regresses the vertebrae’s anatomical label and segments it. Once the entire scan is segmented, the vertebral labels are adjusted using a maximum likelihood approach. Approaching the problem from the other end, \cite{payer2020} propose a coarse-to-fine approach involving three stages, spine localisation, vertebra labelling, and vertebrae segmentation, all three utilising purposefully designed FCNs. Note that \citep{payer2020} and \citep{lessmann19} are included in this \textsc{VerSe} benchmark.  

\subsubsection{Vertebral Labelling}
Similar to the segmentation works discussed above, classical works on vertebral labelling also involve deformable shape or pose models \citep{ibragimov2015interpolation,cai2015multi}. Learning from data, \cite{major2013automated} landmark point using probabilistic boosting trees followed by matching local models using MRFs. As such, works transitioned towards incorporating machine learning using hand-crafted features. \cite{glocker2012,glocker2013vertebrae} employ context features to regress vertebral centroids using regression forests and MRFs. \cite{bromiley2016fully} use Haar-like features to identify vertebrae using random forest regression voting. Similarly, \cite{suzani2015fast} employ an MLP to regress the centroid locations. With the incorporation of the ubiquitous CNNs, \cite{chen2015automatic} proposed a joint-CNN as a combination of random forests for candidate selection followed by a CNN for identifying the vertebrae. \cite{forsberg2017detection} employ CNNs to detect the vertebrae followed by labelling them using graphical models. 

Going fully convolutional and regressing on input-sized heatmap responses instead of directly learning the centroid locations (which is a highly non-linear mapping), \cite{yang2017automatic,yang2017deep} propose DI2IN, an FCN, for heatmap regression of the vertebral centroids at lower resolution, followed by correction using message passing and recurrent neural networks (RNN) respectively. Utilising a single network termed Btrfly-Net, \cite{sekuboyina2018,sekuboyina2020} propose labelling sagittal and coronal maximum intensity projections (MIP) of the spine, reinforced by a prior learnt using a generative adversarial network. Using a three-staged approach, \cite{liao2018joint} combine a CNN with a bidirectional-RNN to label and then fine-tune network predictions. Handling close to two hundred landmarks, \cite{mader2019automatically} use multistage, 3D CNNs to regress heatmaps followed by fine-tuning using  regression trees regularised by conditional random fields. \cite{payer2019integrating} propose a two-stream architecture called spatial-configuration net for integrating global context and local detail in one end-to-end trainable network. With a similar motivation of combining long-range and short-range contextual information, \cite{chen2019lsrc} propose combining a 3D localising network with a 2D labelling network.

\begin{figure*}[t!]
  \centering
%   \vspace{-1em}
  \includegraphics[width=0.9\textwidth]{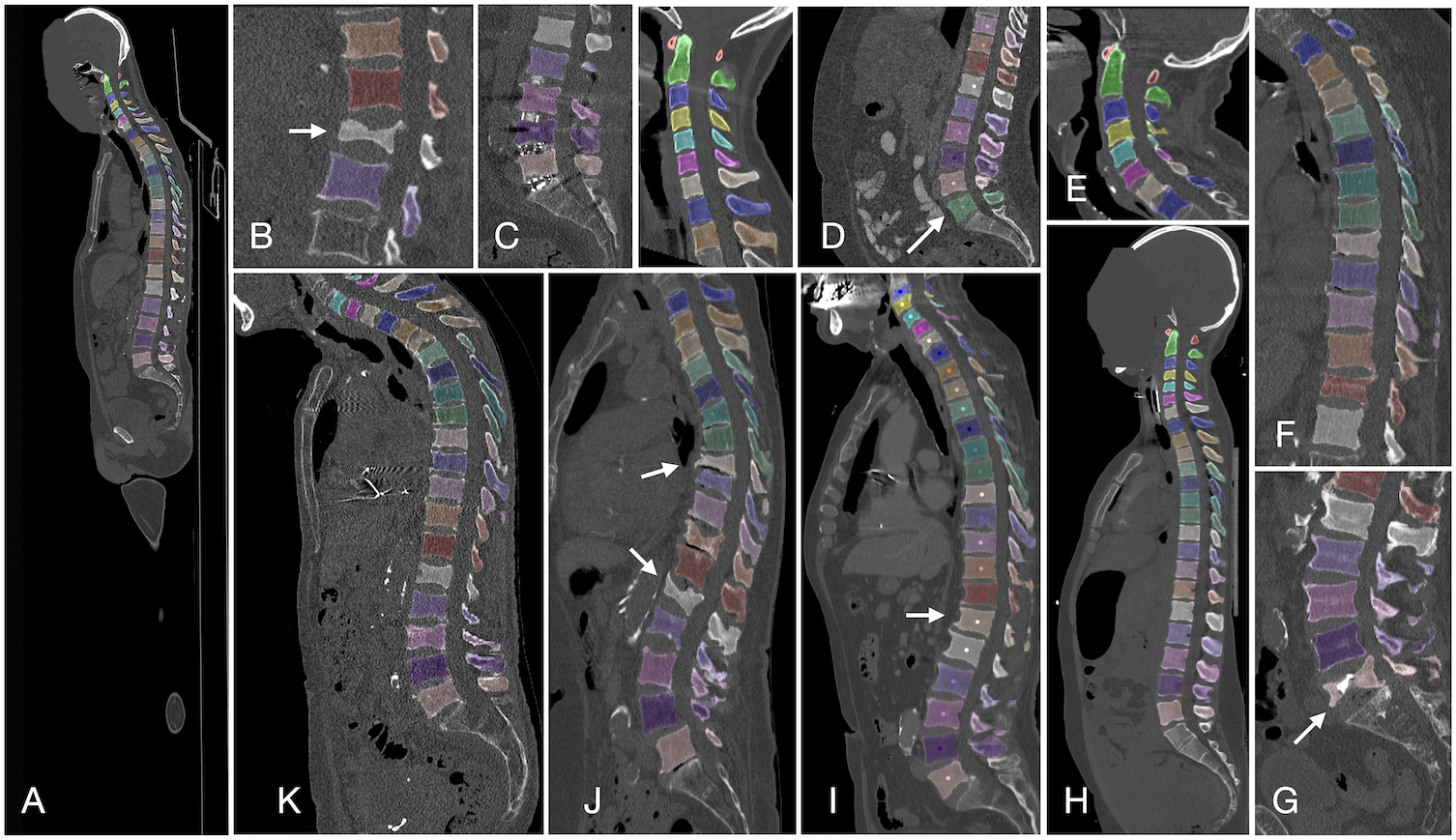}
  \caption{Example scan slices from the \textsc{VerSe} datasets, labelled clockwise. In addition to the wide variation in the fields of view, we illustrate with fractured vertebrae (B, J), metal insertions (C), cemented vertebrae (G), transitional vertebrae (L6 and T13 in D and I respectively), and a noisy scan (K).}
\label{fig:datasample}
\end{figure*}

\begin{table}[t!]
\setlength{\tabcolsep}{0.4em}
\scriptsize
\renewcommand{\arraystretch}{1}

        \begin{tabular}{ c | c : c : c }
        \specialrule{.1em}{0em}{-.1em}
        Dataset & \#train & \#test & Annotations \\[-0.1ex] 
        \specialrule{.05em}{-0.1em}{0em}
        
        CSI-Seg 2014 \citep{yao2012} & 10 & 10 & \textbf{S} \\
        CSI-Label 2014 \citep{glocker2012} & 242 & 60 & \textbf{L} \\
        Dataset-5 \citep{ibragimov2014shape} & 10 & -- & \textbf{S} (Lumbar)\\
        xVertSeg 2016 \citep{korez2015framework} & 15 & 10 & \textbf{S} (Lumbar) \\
        \hline
        \textsc{VerSe} 2019 & 80 & 80 & \textbf{L} + \textbf{S}\\
        \textsc{VerSe} 2020 & 103 & 216 & \textbf{L} + \textbf{S}\\
        
        \specialrule{.1em}{0em}{0em}
        \end{tabular}
        
\caption{Comparing \textsc{VerSe} with other publicly available, annotated CT datasets. In `Annotations', \textbf{L} and \textbf{S} refer to annotations concerning the labelling (3D centroid coordinates) and segmentation tasks (voxel-level labels), respectively.}\label{tab:datasets}
\end{table}

\subsection{Motivation}
Recent spine-processing approaches discussed above are predominantly data-driven, thus requiring annotated data to either learn from (e.g. neural network weights) or to tune and adapt parameters (e.g. active shape model parameters). In spite of this, publicly available data with good-quality annotations is scarce. Eventually, the algorithms are either insufficiently validated or validated in private datasets, preventing a  fair comparison. SpineWeb\footnote{\url{spineweb.digitalimaginggroup.ca}}, an archive for multi-modal spine data, lists a total of four CT datasets with voxel-level or vertebra level annotations: CSI2014-Seg \citep{yao2012,yao2016}, xVertSeg \citep{korez2015framework}, Dataset-5 \citep{ibragimov2014shape}, and CSI2014-Label \citep{glocker2012}. Table \ref{tab:datasets} provides an overview of these public datasets. Except Dataset-5, all datasets were released as part of segmentation and labelling challenges organised as part of the computational spine imaging (CSI) workshop at MICCAI. CSI2014-Seg and Label were made publicly available in conjunction with MICCAI 2014 and xVertSeg with MICCAI 2016. Credit is due to these incipient steps towards open-sourcing data, which have yielded interest in spine processing. A significant portion of the work detailed in Sec.~\ref{subsec:litrev} is benchmarked on these datasets. However, much is to be desired in terms of \emph{data size} and \emph{data variability}. The largest spine CT dataset with voxel-level annotations to date consists of 25 scans, with lumbar annotations only. CSI-Label, even though it is a collection of 302 scans with high data variability, is collected from a single centre (Department of Radiology, University of Washington), possibly inducing a bias.  

With the objective of addressing the need for a large spine CT dataset and to provide a common benchmark for current and future spine-processing algorithms, we prepared a dataset of 374 multi-detector, spine CT (MDCT) scans (an order of magnitude ($\sim$20 times) increase from the prior datasets) with vertebral-level (3D centroids) and voxel-level annotations (segmentation masks). This dataset was made publicly available as part of the \emph{Large Scale \underline{Ver}tebrae \underline{Se}gmentation challenge} (\textsc{VerSe}), organised in conjunction with MICCAI 2019 and 2020. In total, 160 scans were released as part of \textsc{VerSe}`19 and 355 scans for \textsc{VerSe}`20, with a call for fully automated and interactive algorithms for the tasks of \emph{vertebral labelling} and \emph{vertebral segmentation}. 

As part of the \textsc{VerSe} challenge, we evaluated twenty-five algorithms (eleven for \textsc{VerSe}`19, thirteen for \textsc{VerSe}`20, and one baseline). This work presents an in-depth analysis of this benchmarking process, in addition to the technical aspects of the challenge. In summary, the contribution of this work includes:

\begin{itemize}

    \itemsep0em 

    \item A brief description of the setup for the \textsc{VerSe}`19 and \textsc{VerSe}`20 challenges (Sec.~\ref{sec:materials})

    \item A summary of the three top-performing algorithms from each iteration of \textsc{VerSe}, along with a description of the in-house, interactive spine processing algorithm utilised to generate the initial annotation. (Sec. \ref{sec:methods})
    
    \item Performance overview of the participating algorithms and further experimentation provide additional insights into the algorithms. (Sec. \ref{sec:experiments})
    
\end{itemize}

\section{Materials and challenge setup}
\label{sec:materials}
\subsection{Data and annotations}
The entire \textsc{VerSe} dataset consists of 374 CT scans made publicly available after anonymising (including defacing) and obtaining an ethics approval from the institutional review board for the intended use. The data was collected from 355 patients with a mean age of $\sim59(\pm17)$ years. The data is multi-site and was acquired using multiple CT scanners, including the four major manufacturers (GE, Siemens, Phillips and Toshiba). Care was taken to compose the data to resemble a typical clinical distribution in terms of FoV, scan settings, and findings. For example, it consists of a variety of FoVs (including cervical, thoraco-lumbar and cervico-thoraco-lumbar scans), a mix of sagittal and isotropic reformations, and cases with vertebral fractures, metallic implants, and foreign materials. Fig. \ref{fig:datasample} illustrates this variability in the \textsc{VerSe} dataset. Refer to \cite{loeffler2020_verse,liebl2021computed} for further details on the data composition.

The dataset consists of two types of annotations: 1) 3D coordinate locations of the vertebral centroids and 2) voxel-level labels as segmentation masks. Twenty-six vertebrae (C1 to L5, and the transitional T13 and L6) were considered for annotation with labels from 1 to 24, along with labels 25 and 28 for L6 and T13, respectively. Note that partially visible vertebrae at the top or bottom of the scan (or both) were not annotated. Annotations were generated using a human-hybrid approach. The initial centroids and segmentation masks were generated by an automated algorithm (details in Sec.~\ref{sec:methods}) and were manually and iteratively refined. Initial refinement was performed by five trained medical students followed by further refinement, rejection, or acceptance by three trained radiologists with a combined experience of 30 years (ML, HL, and JSK). All annotations were finally approved by one radiologist with 19 years of experience in spine imaging (JSK).

\subsection{Challenge setup}

\textsc{VerSe} was organised in two iterations, first at MICCAI 2019 and then at MICCAI 2020 with a call for algorithms tackling vertebral labelling and segmentation. Both the iterations followed an identical setup, wherein the challenge consisted of three phases: one training and two test phases. In the training stage, participants have access to the scans and their annotations, on which they can propose and train their algorithms. In the first test phase, termed \textsc{Public} in this work, participants had access to the test scans on which they were supposed to submit the predictions. In the second test phase, termed \textsc{Hidden}, participants had no access to any test scans but were requested to submit their code in a docker container. The dockers were evaluated on hidden test data, thus disabling re-training on test data or fine-tuning via over-fitting. Information about the data and its split across the two \textsc{VerSe} iterations is tabulated in Table \ref{tab:data_split}. \textbf{All the 374 scans of} \textsc{VerSe} \textbf{dataset and their annotations are now publicly available, 2019: \url{https://osf.io/nqjyw/} and 2020: \url{https://osf.io/t98fz/}. We have also open-sourced the data processing and the evaluation scripts. All} \textsc{VerSe}\textbf{-content is accessible at \url{https://github.com/anjany/verse}}

\begin{table}[t!]
\setlength{\tabcolsep}{0.4em}
\scriptsize
\renewcommand{\arraystretch}{1}

        \begin{tabular}{ c | c c c c}
        \specialrule{.1em}{0em}{-.1em}
         \textsc{VerSe} & Patients & Scans & Scan split & Vertebrae (Cer/Tho/Lum)\\[-0.1ex] 
        \specialrule{.05em}{-0.1em}{0em}
        
        2019 & 141 & 160 & 80/40/40 & 1725 (220/884/621)\\
        2020 & 300 & 319 & 113/103/103 & 4141 (581/2255/1305)\\
        \hdashline
        Total & 355 & 374 & 141/120/113 & 4505 (611/2387/1507)\\
        \specialrule{.1em}{0em}{0em}
        \end{tabular}
        
\caption{Data split and additional details concerning the two iterations of VerSe. Scan split indicates the split of the data into training/\textsc{Public} test/\textsc{Hidden} test phases. Cer, Tho, and Lum refer to the number of vertebrae from the cervical, thoracic, and lumbar regions, respectively. Note that of the 300 patients in  \text{VerSe}`20, 86 patients are from \textsc{VerSe}`19, resulting in the total patients not being an \emph{ad hoc} sum of the two iterations. \textsc{VerSe}`19 data can be identified by its image ID being less than 500. (Overlap is not absolute owing to the difference in the objectives of the two challenge iterations.)}\label{tab:data_split}
\end{table}

\subsection{Evaluation metrics}
In this work, we employ four metrics for evaluation, two for the task of labelling and two for the task of segmentation. Note that the evaluation protocol employed for ranking the challenge participants builds on the one presented in this work. Please refer to \ref{app:ranking} for an overview of the former. 

\noindent
\textbf{Labelling.} To evaluate the labelling performance, we compute the \emph{Identification Rate} ($id.rate$) and localisation distance ($d_\text{mean}$): Assuming a given scan contains $N$ annotated vertebrae and denoting the true location of the $i^{th}$ vertebra with $x_i$ and it predicted location with $\hat{x}_i$, the vertebra $i$ is correctly \emph{identified} if $\hat{x}_i$ is the closest landmark predicted to $x_i$ among $\{x_j  \forall j \mathrm{~in~} 1, 2, ..., N\}$ and the Euclidean distance between the ground truth and the prediction is less than \siunit{20}{mm}, i.e. $||\hat{x}_i - x_i||_2 <$ \siunit{20}{mm}. For a given scan, $id.rate$ is then defined as the ratio of the correctly identified vertebrae to the total vertebrae present in the scan. Similarly, the localisation distance is computed as $d_\text{mean} = (\sum_{i=1}^{N} ||\hat{x}_i - x_i||_2)/N$, the mean of the euclidean distances between the ground truth vertebral locations and their predictions, per scan. Typically, we report the mean measure over all the scans in the dataset. Note that our evaluation of the labelling tasks slightly deviates from its definition in \citep{glocker2012}, where $id.rate$ and $d_\text{mean}$ are computed not at scan-level but at dataset level.

\noindent
\textbf{Segmentation.} To evaluate the segmentation task, we choose the ubiquitous Dice coefficient (Dice) and Hausdorff distance ($HD$). Denoting the ground truth by $T$ and the algorithmic predictions by $P$, and indexing the vertebrae with $i$, we compute the mean Dice score across the vertebrae as follows:
\begin{equation}
\text{Dice}(P,T) = \frac{1}{N}\sum_{i=1}^{N} \frac{2|P_i \cap T_i|}{|P_i| + |T_i|}.
\end{equation}

As a surface measure, we compute the mean Hausdorff distance over all vertebrae as:
\begin{equation}
HD(P,T) = \frac{1}{N}\sum_{i=1}^{N} \max\left\{\sup\limits_{p\in\mathcal{P}_i} \inf\limits_{t\in\mathcal{T}_i} d(p,t), \sup\limits_{t\in\mathcal{T}_i} \inf\limits_{p\in\mathcal{P}_i} d(p,t)\right\},
\end{equation}
where $\mathcal{P}_i$ and $\mathcal{T}_i$ denote the surfaces extracted from the voxel masks of the $i^\text{th}$ vertebra and $d(p,t) = ||p - t||_2$, i.e a Euclidean distance between the points $p$ and $t$ on the two surfaces.

\noindent
\emph{Outliers.} In multi-class labelling and segmentation, there will be cases where the prediction of an algorithm will contain fewer vertebrae than the ground truth. In such cases, $d_\text{mean}$ and $HD$ are not defined for the missing vertebrae. For the sake of analysis in this work, we ignore such vertebrae while computing the averages. This way, we still get a picture of the algorithm's performance on the rest of the correctly predicted vertebrae. The missing vertebrae are anyway clearly penalised by the other two metrics, viz. $id.rate$ and Dice.

\section{Methods}
\label{sec:methods}

In this section, we present Anduin, our spine processing framework that enabled the medical experts to generate voxel-level annotations at scale. Then, we present details of select participating algorithms.     

\begin{figure}[t!]
  \centering
%   \vspace{-1em}
  \includegraphics[width=0.5\textwidth]{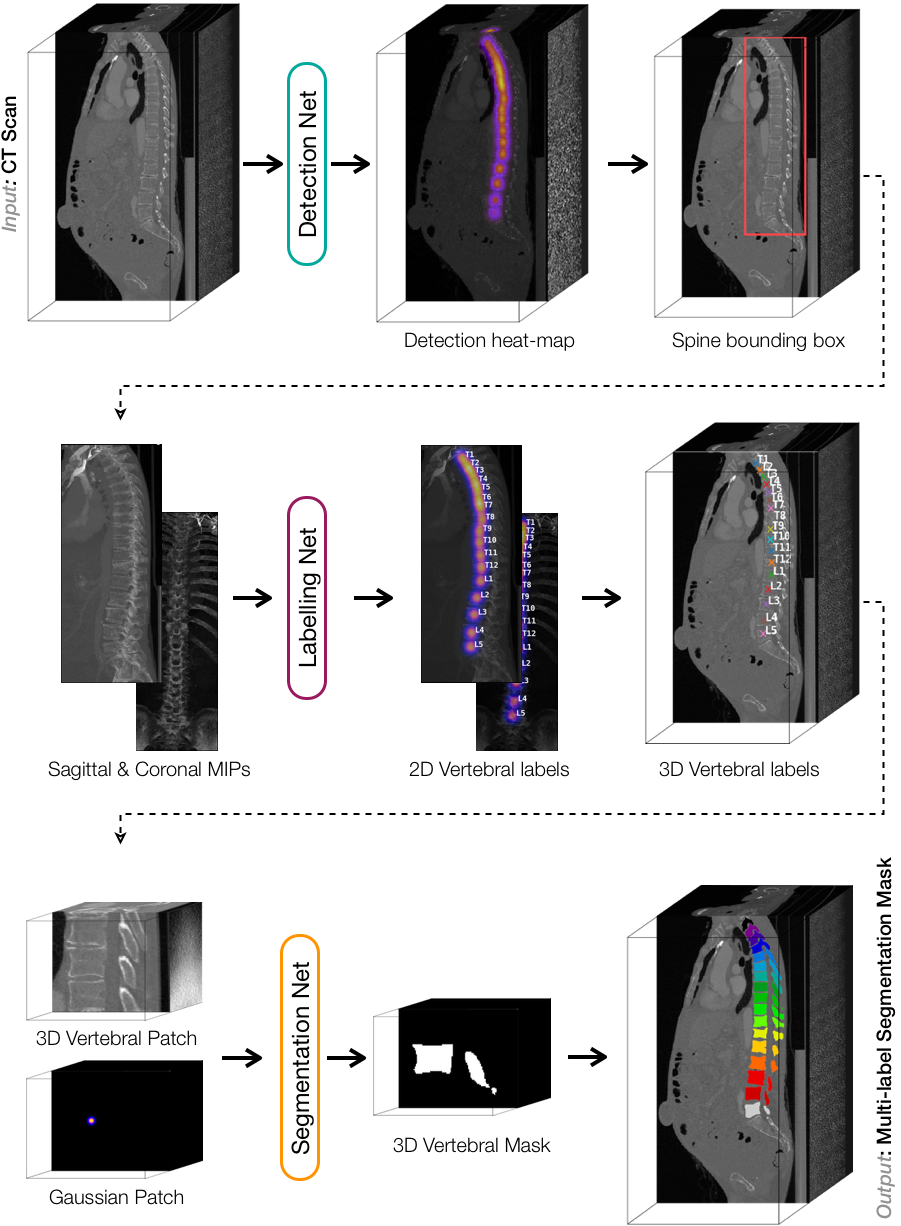}
  \caption{\textbf{Our interactive spine-processing pipeline}: Schematic of the semi-automated and interactive spine processing pipeline developed in-house. The bold lines indicate automated steps. The dotted lines indicate a \emph{possibly} interactive step.}
\label{fig:pipeline}
\end{figure}

\subsection{Anduin: Semi-automated spine processing framework}

Anduin is a semi-automated, interactive processing tool developed in-house, which was employed to generate the \emph{initial} annotations for more than 4000 vertebrae. It is a three-staged pipeline consisting of: 1) \emph{Spine detection}, performed by a light-weight, FCN predicting a low-resolution heatmap over the spine location, 2) \emph{Vertebra labelling}, based on the Btrfly Net \citep{sekuboyina2018} architecture working on sagittal and coronal MIPs of the localised spine region, and finally, 3) \emph{Vertebral segmentation}, performed by an improved U-Net \citep{ronneberger2015,roy2018} to segment vertebral patches, extracted at 1mm resolution, around the centroids predicted by the preceding stage. Fig. \ref{fig:pipeline} gives a schematic of the entire framework. Importantly, the detection and labelling stages offer interaction, wherein the user can alter the bounding box predicted during spine detection as well as the vertebral centroids predicted by the labelling stage. Such \emph{human-in-loop} design enabled the collection of accurate annotations with minimal human effort.  We made a web-version of \emph{Anduin} publicly available to the research community that can be accessed at \url{anduin.bonescreen.de}. Refer to \ref{app:anduin_details} for further details on \emph{Anduin} (at the time of this work) such as network architecture, training scheme, and post-processing steps. Furthermore, without human-interaction, \emph{Anduin} is fully automated. We include this version of \emph{Anduin} in the benchmarking process as `Sekuboyina A.'. We note that since the ground-truth segmentation masks are generated with \emph{Anduin}-predictions as initialisation, there exists a bias. However, the bias is not as strong for the labelling task as the centroid annotations are sparse and have a high intra- and inter-rater variability.

\begin{table*}[!htbp]
\tiny
\centering
% \begin{adjustbox}{angle=90}
\begin{tabular}{l : l | l}% p{7cm}  } 
\specialrule{.1em}{0em}{-.1em}
 & \textbf{Team}~/~\textbf{Ref. Author} & \multicolumn{1}{c}{\textbf{Method Features}} \\ [0.5ex] 

\specialrule{.05em}{-0.1em}{0em}

%  & \crule[f29e8e]{0.25cm}{0.25cm} anduin / Sekuboyina A. & \tabincell{l}{It is a three-stage pipeline consisting  of: 1) Spine detection by a light-weight network predicting a \\ low-resolution heatmap, 2) Vertebra labelling, based on the Btrfly Net \citep{sekuboyina2018} working \\on sagittal and coronal MIPs, and 3) Vertebral segmentation by an improved U-Net at 1mm resolution.} \\
% \cline{2-3}
\parbox[t]{1mm}{\multirow{25}{*}{\rotatebox[origin=c]{90}{\textsc{VerSe}`19}}} & \crule[ff0000]{0.25cm}{0.25cm} zib / Amiranashvili T. & 
\tabincell{l}{Multi-stage, shape-based approach. Multi-label segmentation with arbitrary labels for vertebrae. Unique label \\ assignment for based on shape templates. Landmark positions are derived as centres of fitted model.}\\
%\hline
\cline{2-3}
 %\hline
& \crule[b00000]{0.25cm}{0.25cm} christoph / Angermann C. & \tabincell{l}{Single-staged, slice-wise approach. One 2.5D U-Net \citep{angermann19} and two 2D U-Nets are employed.\\ The first network generates 2D projections containing 3D information. Then, one 2D U-Net segments the\\ projections, one segments the 2D slices. Labels are obtained as centroids of segmentations.} \\
 %\cline{2-5}
\cline{2-3}
 & \crule[870000]{0.25cm}{0.25cm} brown / Brown K. & \tabincell{l}{A 3D bounding box around the vertebra is predicted by regressing on a set of canonical landmarks. Each\\ vertebra is segmented using a residual U-Net and labelled by registering to a common atlas.} \\
 %\cline{2-5}
\cline{2-3}
 & \crule[550000]{0.25cm}{0.25cm} iflytek / Chen M. & \tabincell{l}{A three-staged approach. Spine localisation and multi-label segmentation are based on a 3D U-Net. Using the \\predicted segmentation mask, the third stage employs a RCNN-based architecture to label the vertebrae.} \\
 %\cline{2-5}
\cline{2-3}
 & \crule[e4e400]{0.25cm}{0.25cm} yangd05 / Dong Y. & \tabincell{l}{Single-staged approach. A 3D U-Net based on neural-architecture search is employed to segment vertebrae as \\26-class problem. Vertebral-body centre are located using iterative morphological erosion.} \\ 
 %\cline{2-5}
\cline{2-3}
 & \crule[baba00]{0.25cm}{0.25cm} huyujin / Hu Y. & \tabincell{l}{Single-staged, patch-based approach. Based on the nnU-Net \citep{isensee2019}. All three networks are used:\\ a 3D-UNet at high resolution, a 3D U-Net at low resolution, and a 2D U-Net.}\\
 %\cline{2-5}
\cline{2-3}
 & \crule[878700]{0.25cm}{0.25cm} alibabadamo / Jiang T. & \tabincell{l}{Single-staged approach, employing a V-Net \citep{milletari2016v} backbone with two heads, one for binary-\\segmentation and the other for vertebral-labelling. Vertebrae C2, C7, T12, and L5 are identified and the rest\\ are inferred from these.}\\
 %\cline{2-5}
\cline{2-3}
 & \crule[545400]{0.25cm}{0.25cm} lrde / Kirszenberg A. & \tabincell{l}{Multi-stage, shape-based approach. A combination of three 2D U-Nets generate 3D binary mask of spine. \\ Anchor points on a skeleton obtained from this mask are used for template matching. Five vertebrae are \\ chosen for matching, and one with highest score is chosen as a match. } \\
 %\cline{2-5}
\cline{2-3}
 & \crule[00ff00]{0.25cm}{0.25cm} diag / Lessmann N. & \tabincell{l}{Single-staged, patch-based approach. A 3D U-Net \citep{lessmann19} iteratively identifies and segments\\ the bottom-most visible vertebra in extracted patches, eventually crawling the spine. An additional network is\\ trained to detect first cervical and thoracic vertebrae.} \\
 %\cline{2-5}
\cline{2-3}
 & \crule[00b000]{0.25cm}{0.25cm} christian\_payer / Payer C. & \tabincell{l}{Multi-staged, patch-wise approach. A 3D U-Net regresses a heatmap of the spinal centre line. Individual vert-\\ebrae are localized and are identified with the SpatialConfig-Net \citep{payer2020}. Each vertebra is then \\independently segmented as a binary segmentation.} \\
%\cline{2-5}
\cline{2-3}
 & \crule[008700]{0.25cm}{0.25cm} init / Wang X. & \tabincell{l}{Multi-staged-approach. A single-shot 2D detector is utilised to localise the spine. A modified Btrfly-Net\\ \citep{sekuboyina2018} and a 3D U-Net are employed to address labelling and segmentation respectively.} \\
\hline

 \parbox[t]{1mm}{\multirow{28}{*}{\rotatebox[origin=c]{90}{\textsc{VerSe}`20}}} & \tabincell{l}{\crule[005500]{0.25cm}{0.25cm}{\null} deepreasoningai\_team1 \\ / Chen D.} & \tabincell{l}{Multi-staged, patch-based approach. A 3D U-Net coarsely localises the spine. Then, a U-Net performs binary\\ segmentation, patchwise. Lastly, a 3D Resnet-model identifies the vertebral class taking the vertebral mask\\ and CT-image segmented vertebra.}\\
 %\cline{2-5}
\cline{2-3}
& \crule[00ffff]{0.25cm}{0.25cm}{\null} carpediem / Hou F. & \tabincell{l}{Multi-staged approach. First, the spine position is located with 3D U-Net. Second the vertebrae are labelled\\ in the cropped patches. Lastly, U-Net segments individual vertebrae from background using centroids labels.}\\
 %\cline{2-5}
\cline{2-3}
& \crule[00b0b0]{0.25cm}{0.25cm}{\null} poly / Huang Z. & \tabincell{l}{Single-staged, patch-based approach. A U-Net with feature-aggregation and squeeze \& exictation module is\\ proposed.Contains two task-specific heads, one for vertebrae labelling and the other for segmentation.}  \\
 %\cline{2-5}
\cline{2-3}
& \crule[008787]{0.25cm}{0.25cm}{\null} lrde / Hu\`{y}nh L. D. & \tabincell{l}{A single model with two-stages, a Mask-RCNN-inspired model incorporating RetinaNet is proposed. First \\stage detects and classifies vertebral RoIs. Second stage outputs a binary segmentation for each of the RoIs.}\\
 %\cline{2-5}
\cline{2-3}
& \crule[005555]{0.25cm}{0.25cm}{\null} ubmi / Jakubicek R. & \tabincell{l}{Multi-staged, semi-automated approach \citep{jakubicek2020tool}. Stages include: spine-canal tracking,\\ localising and labelling the inter-vertebral disks, and then labelling the vertebrae. Segmentation is based on\\ graph-cuts.}\\
 %\cline{2-5}
\cline{2-3}
& \crule[b0b0ff]{0.25cm}{0.25cm}{\null} htic / Mulay S. & \tabincell{l}{Single-staged approach. A 2D Mask R-CNN withcomplete IoU loss performs slice-wise segmentation.}\\ 
 %\cline{2-5}
\cline{2-3}
& \crule[8484ff]{0.25cm}{0.25cm}{\null} superpod / Netherton T. J. & \tabincell{l}{Multi-staged approach. Combines a 2D FCN for coarse spinal canal segmentation, a multi-view X-Net\\ \citep{netherton2020evaluation} for labelling, and a U-Net++ architecture for vertebral segmentation.}\\
 %\cline{2-5}
\cline{2-3}
& \crule[4949ff]{0.25cm}{0.25cm}{\null} rigg / Paetzold J.& A naive 2D U-Net performs multi-class segmentation of sagittal slices.  \\
 %\cline{2-5}
\cline{2-3}
& \crule[0000ff]{0.25cm}{0.25cm}{\null} christian\_payer / Payer C. & \tabincell{l}{Similar to Payer C.'s 2019 submission. Different from it, Markov Random fields are employed for post-\\processing the localisation stage's output. Additionally, appropriate floating-point optimisation of network\\ weights scans into patches.}\\
 %\cline{2-5}
\cline{2-3}
& \crule[ff00ff]{0.25cm}{0.25cm}{\null} fakereal / Xiangshang Z. & \tabincell{l}{Both tasks are handled individually. A modified Btrfly-Net \citep{sekuboyina2018} detects vertebral key\\ points. An nnU-Net \citep{isensee2019} performs multi-class segmentation.} \\
 %\cline{2-5}
\cline{2-3}
& \crule[b000b0]{0.25cm}{0.25cm}{\null} sitp / Yeah T.& \tabincell{l}{Two-staged approach containing two 3D U-Nets. First one performs coarse localisation of the spine\\ at low-resolution. Second one performs multi-class segmentation of the vertebra at a higher resolution.}\\
 %\cline{2-5}
\cline{2-3}
& \crule[870087]{0.25cm}{0.25cm}{\null} aply / Zeng C.& \tabincell{l}{Multi-staged approach. First stage detects five key-points on the spine using a HRNet. Second,\\ improved Spatialconfig-Net \citep{payer2019integrating} performs the labelling. Segmentation is now a binary problem.}\\
 %\cline{2-5}
\cline{2-3}
& \crule[550055]{0.25cm}{0.25cm}{\null} jdlu / Zhang A. & \tabincell{l}{A four-step approach. A patch-based V-Net is used to regress the spine center-line. A key-point localization V-\\Net predicts potential vertebral candidates. A three-class vertebrae segmentation network obtains main class of\\ each vertebrae. Final labels are obtained using a rule-based postprocessing.}  \\ [1ex]

 \specialrule{.1em}{0em}{-.1em}
 \end{tabular}
\caption{Brief summary of the participating methods in \textsc{VerSe} benchmark, ordered alphabetically according to referring author.}
% \end{adjustbox}
\label{tab:teams}
\end{table*}

\subsection{Participating methods}
Over its two iterations, \textsc{VerSe} has received more than five hundred data download requests. Forty teams uploaded their submissions onto the leaderboards. Of these, eleven and thirteen teams were evaluated for \textsc{VerSe}`19 and \textsc{VerSe}`20, respectively. Table \ref{tab:teams} provides a brief synopsis of all the participating teams. Below, we present the algorithms proposed by the best and the second-best-performing teams in each iteration of the challenge. \ref{app:teams} provides the details of the remaining algorithms.

% -- payer ----------------------------
\subsection*{\crule[00b000]{0.25cm}{0.25cm} Payer C. et al.: Vertebrae localisation and segmentation with SpatialConfiguration-net and U-net \textsc{[VerSe`19]}}
\label{desc:payerc}

% \begin{wrapfigure}{r}{0.5\textwidth}
%   \centering
%   \vspace{-0.25in}
%   \includegraphics[width=0.45\textwidth]{figures/appendix/payer.png}
%   \caption{The three processing stages in \emph{Payer C.} for localisation, identification, and segmentation of vertebrae.}
% \label{fig:payer}

\begin{figure}%{\textwidth}
  \centering
%   \vspace{-0.25in}
  \includegraphics[width=0.4\textwidth]{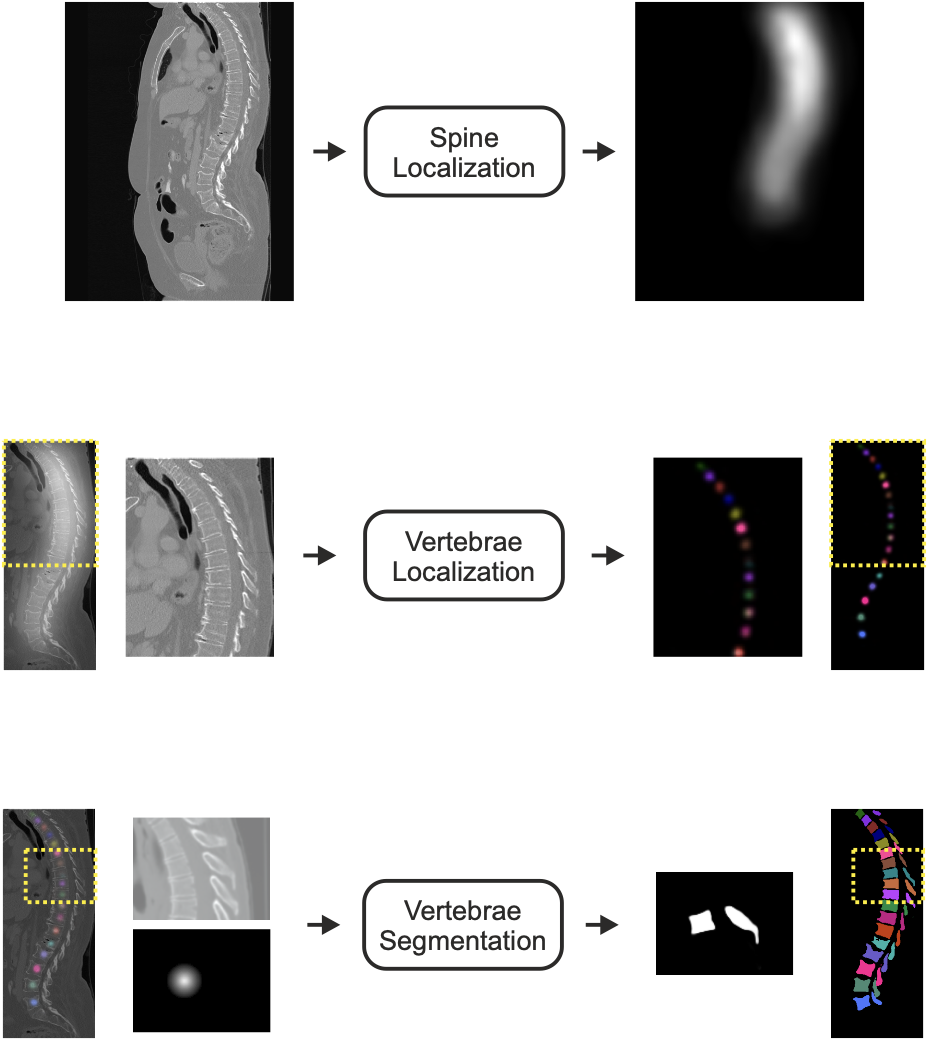}
  \caption{The three processing stages in \emph{Payer C.} for localisation, identification, and segmentation of vertebrae.}
\label{fig:payer}

 %\vspace{-110pt} % This removes the white box on the second page
\end{figure}
Vertebrae localisation and segmentation are performed in a three-step approach: spine localisation, vertebrae localisation and identification, and finally binary segmentation of each located vertebra (cf. Fig. \ref{fig:payer}). The results of the individually segmented vertebrae are merged into the final multi-label segmentation.

\noindent
\emph{Spine Localisation.} To localise the approximate position of the spine, a variant of the U-Net was used to regress a heatmap of the spinal centreline, i.e. the line passing through vertebral centroids, with an $\ell_2$ loss. The heatmap of the spinal centreline is generated by combining Gaussian heatmaps of all individual landmarks. The input image is resampled to a uniform voxel spacing of \siunit{8}{mm} and centred at the network input.

\noindent
\emph{Vertebra Localisation \& Identification.} The SpatialConfiguration-Net \citep{payer2020} is employed to localise centres of the vertebral bodies. It effectively combines the local appearance of landmarks with their spatial configuration. Please refer to \citep{payer2020} for details on architecture and loss functions. Every input volume is resampled to have a uniform voxel spacing of \siunit{2}{mm}, while the network is set up for inputs of size $96\times96\times128$. As some volumes have a larger extent in the cranio-caudal axis and do not fit into the network, these volumes are processed as follows: During training, sub-volumes are cropped at a random position at the cranio-caudal axis. During inference, volumes are split at the cranio-caudal axis into multiple sub-volumes that overlap for 96 pixels and processed them one after another. Then, the network predictions of the overlapping sub-volumes are merged by taking the maximum response over all predictions.

Final landmark positions are obtained as follows: For each predicted heatmap volume, multiple local heatmap maxima are detected that are above a certain threshold. Then, the first and last vertebrae that are visible on the volume are determined by taking the heatmap with the largest value that is closest to the volume top or bottom, respectively. The final predicted landmark sequence is then the sequence that does not violate the following conditions: consecutive vertebrae may not be closer than 12.5 mm and further away than 50 mm, and a subsequent landmark may not be above a previous one.

\noindent
\emph{Vertebra Segmentation.}
To create the final vertebrae segmentation, a U-Net is set up with a sigmoid cross-entropy loss for binary segmentation to separate individual vertebrae. The entire spine image is cropped to a region around the localised centroid such that the vertebra is in the centre of the image. Similarly, the heatmap image of the vertebral centroid is also cropped from the prediction of the vertebral localisation network. Both the cropped vertebral image and vertebral heatmap are used as an input for the segmentation network. Both input volumes are resampled to have a uniform voxel spacing of 1 mm. To create the final multi-label segmentation result, the individual predictions of the cropped inputs are resampled back to the original input resolution and translated back to the original position.

% -- lessmann ----------------------------

\subsection*{\crule[00ff00]{0.25cm}{0.25cm} Lessmann et al.: Iterative fully convolutional neural networks \textsc{[VerSe`19]}}
\label{desc:lessmannn}

The proposed approach largely depends on iteratively applied fully convolutional neural networks \citep{lessmann19}. Briefly, this method relies on a U-net-like 3D network that analyses a $128\times128\times128$ region of interest (RoI). In this region, the network segments and labels only the bottom-most visible vertebra and ignores other vertebrae that may be (partly) visible within the RoI. The RoI is iteratively moved over the image by placing it at the centre of the detected piece of the vertebra after each segmentation step. If only part of a vertebra was detected, moving the RoI to the centre of the detected fragment ensures that a larger part of the vertebra becomes visible for the next iteration. Once the
entire vertebra is visible in the RoI, the segmentation and labeling results are stored in a memory component. This memory is a binary mask that is an additional input to the network and is used by the network to recognise and ignore already segmented vertebrae. By repeating the process of searching for a piece of vertebra and following this piece until the whole vertebra is visible in the region of interest, all vertebrae are segmented and labeled one after the other. When the end of the scan is reached, the predicted labels of all detected vertebrae are combined in a global maximum likelihood model to determine a plausible labeling for the entire scan, thus avoiding duplicate labels or gaps. Please refer to \citep{lessmann19} for further details. Note that two publicly available datasets were also used for training: CSI-Seg 2014 \citep{yao2012} and the xVertSeg 2016 datasets \citep{korez2015framework}. The approach is supplemented with minor changes over \citep{lessmann19} so that: anatomical labelling of the detected vertebra is optimised by minimising a combination of $\ell_1$ and $\ell_2$ norms; the loss for the segmentation network is a combination of the proposed segmentation error and a cross-entropy loss.  

\noindent
\emph{Rib Detection.} In order to improve the labeling accuracy, a second network is trained to predict whether a vertebra is a thoracic vertebra or not. As input, this network receives the final image patch in which a vertebra is segmented and the corresponding segmentation mask as a second channel. The network has a simple architecture based on $3\times3\times3$ convolutions, batch normalisation and max-pooling. The final layer is a dense layer with a sigmoid activation function. At inference time, the first thoracic vertebra and the first cervical vertebra identified by this auxiliary network had a stronger influence on the label voting. Their vote counted three times as much as that of other vertebrae.

\noindent
\emph{Cropping at Inference.} Note that if the first visible vertebra is not properly detected, the whole iterative process might fail. Therefore, at inference time, an additional step is added which crops the image along the z-axis in steps of 2.5\% from the bottom if no vertebra was found in the entire scan. This helps in case the very first, i.e. bottom-most, vertebra is only visible with a very small fragment. This small element might be too small to be detected as a vertebra but might prevent the network from detecting any vertebra above as the bottom-most vertebra.

\noindent
\emph{Centroid Estimation.} Instead of the vertebral centroids provided as training data, the centroids of the segmentation masks were utilised to estimate the `actual' centroids. This was done by estimating the offset between the centroids measured from the segmentation mask ($\hat{v}_i$) and the expected centroids ($v_i$). For every vertebra individually, an offset ($\delta$) was determined by minimising $\sum_{i} \hat{v}_i - v_i + \delta$.

\subsection*{\crule[005500]{0.25cm}{0.25cm} Chen D. et al.: Vertebrae Segmentation and Localisation via Deep Reasoning \textsc{[VerSe`20]}}
\label{desc:chend}
The authors propose deep reasoning approach as a multi-stage scheme. First, a simple U-Net model with a coarse input resolution identifies the approximate location of the entire spine in the CT volume to identify the area of interest. Secondly, another U-Net with a higher resolution is used, zoomed in on the spinal region, to perform binary segmentation on each individual vertebra (bone vs. background). Lastly, a CNN is employed to perform multi-class classification for each segmented vertebra obtained from the second step. The results of the classification and the segmentation are merged into the final multi-class segmentation, which is then used to compute the corresponding centroids for each vertebra. 
\\
\emph{Spine Localisation.} Considering the large volume of whole-body CT scan, the original CT image is down-sampled to a coarse resolution and fed to a shallow 3D-UNet to identify the rough location of the visible spine. The network has the following number of feature maps for both the sequential down and up sampling layers: 8, 16, 32, 64, 128, 64, 32, 16, 8. This is similar to Payer C. \emph{et al.'s} method for \textsc{VerSe}`19 in Section \ref{desc:payerc}. The authors replaced batch normalisation with instance normalisation and ReLU activation with leaky ReLU (leak rate of 0.01), similar to \cite{payer2020}.
\\
\emph{Vertebrae Segmentation.} 
The authors train a 3D U-Net model to solely perform binary segmentation (vertebrae bone vs. background) at a resolution of 1mm. Given the natural sequential structure of the vertebrae, inspired by \cite{lessmann2018iterative}, the authors train a model to perform an iterative vertebrae segmentation process along the spine. That is, the model is given the mask of the previous vertebra and the CT scan as input, and mask for the next vertebrae is predicted. The input is restricted to a a small-sized patch obtained from the spine localisation step. A 3D U-Net with the following number of kernels for both the sequential down and up sampling layers is used: 64, 128, 256, 512, 512, 512, 256, 128, 64.
\\
\emph{Vertebrae Classification.}
A 3D ResNet-50 model is used to predict the class of each vertebra. As input, this model takes the segmentation mask obtained in the vertebral segmentation step, as well as the corresponding CT volume, and outputs a single class for the entire vertebrae. Given the prior knowledge of the anatomical structure of the spine and its variations, it can be ensured that the predictions are anatomically valid. 
\\
\emph{Deep Reasoning Module}
Given the biological setting of this computer vision challenge, the task is very structured and the proposed models use reasoning to leverage the anatomical structure and prior knowledge. Using the Deep Reasoning framework \citep{chen2020deep}, the authors were able to encode and constrain the model to produce results that are anatomically correct in terms of the sequence of vertebrae, as well as only produce vertebral masks that are anatomically possible.

\subsection*{\crule[0000ff]{0.25cm}{0.25cm} Payer C. et al.: Improving Coarse to Fine Vertebrae Localisation and Segmentation with SpatialConfiguration-Net and U-Net \textsc{[VerSe`20]}}
The overall setup of the algorithm stays the same as Payer et al.'s approach for \textsc{VerSe}`19 \citep{payer2020}: a three-stage approach consisting of: spine localisation, vertebrae localisation and identification, and finally binary segmentation of each located vertebra. 

This approach, however, differs in its post-processing after the localisation and identification stage, due to an increased variation in the \textsc{VerSe}`20 data. For all vertebrae $i \in \{\text{C1 ... L6}\}$, the authors generate multiple location candidates and identify the ones that maximises the following function of the graph with vertices $\mathcal{V}$ and edges $\mathcal{E}$ modelling an MRF,
\begin{equation}
\label{eq:payerc}
\sum\limits_{i\in\mathcal{V}}\mathcal{U}\left(v_i^k\right) + \sum\limits_{i,j\in\mathcal{E}}\mathcal{P}\left(v_i^k, v_j^l\right),
\end{equation}
where $\mathcal{U}$ describes the unary weight of candidate $k$ of vertebrae $i$, and $\mathcal{P}$ describes the pairwise weight of the edge from candidate $k$ of vertebrae $i$ to candidate $l$ of vertebrae $j$. An edge from $i$ to $j$ exists in the graph if $v_i$ and $v_j$ are possible subsequent neighbors in the dataset. 

The unary terms are set to the heatmap responses plus a bias, i.e. $u\left(v_i^k\right) = \lambda h_i^k + b$, where $h_i^k$ is the heatmap response of the candidate $k$ of vertebra $i$, $b$ is the bias, and $\lambda$ is the weighting factor. The pairwise terms penalise deviations from the average vector from vertebrae $i$ to $j$ and are defined as 
\begin{equation}
\mathcal{P}\left(v_i^k, v_j^l\right) = (1-\lambda)\left(1 - \left|\left|2\frac{\overline{d_{i,j}} - d_{i,j}^{k,l}}{||\overline{d_{i,j}}||_2}\right|\right|^2\right),
\end{equation}
with $d_{i,j}$ being the mean vector from vertebra $i$ to $j$ in the ground truth, $d^{k,l}_{i,j}$ being the vector from $v^k_i$ and $v^l_j$, and $||\cdot||$ denoting the Euclidean norm. 

The bias is set to 2.0 and also encourages the detection of vertebrae, for which the unary and pairwise terms would be slightly negative. The weighting factor $\lambda$ set 0.2 to encourage the MRF to more rely on the direction information. For the location candidates of vertex $v_i$, the authors take the local maxima responses of the predicted heatmap with a heatmap value larger than 0.05. Additionally, as the authors observed that the networks often confuse subsequent vertebrae of the same type, the authors add to the location candidates of a vertebra also the candidates of the previous and following vertebrae of the same type. For these additional candidates from the neighbors, heatmap response is penalised by multiplying it by a factor of 0.1 such that the candidates from the actual landmark are still preferred. Function~\ref{eq:payerc} is solved by creating the graph and finding the shortest negative path from a virtual start to a virtual end vertex.

Another minor change involves usage of mixed-precision networks. The memory consumption of training the networks is drastically reduced due to 16-bit floating-point intermediate outputs, while the accuracy of the networks stays high due to the network weights still being represented as 32-bit floating-point values.

\section{Experiments}
\label{sec:experiments}

In this section, we report the performance measures of the participating algorithms in the \emph{labelling} and \emph{segmentation}  tasks. Following this, we present a dissected analysis of the algorithms over a series of experiments that help understand the tasks as well as the algorithms.

\begin{table*}[!t]
\setlength{\tabcolsep}{0.2em}
\scriptsize
\renewcommand{\arraystretch}{1}
\begin{subtable}{\linewidth}\centering
 {
        \begin{tabular}{ c | c : c | c : c | c : c | c : c}
        \specialrule{.1em}{0em}{-.1em}
        \multirow{3}{*}{\rule{0pt}{2.5ex}\textbf{Team}} & \multicolumn{4}{c}{Labelling} & \multicolumn{4}{c}{Segmentation} \\[-1ex] 
        
        & \multicolumn{2}{c}{\textsc{Public}} & \multicolumn{2}{c}{\textsc{Hidden}} & \multicolumn{2}{c}{\textsc{Public}} & \multicolumn{2}{c}{\textsc{Hidden}}\\
        &  $id.rate$ & $d_\text{mean}$ & $id.rate$ & $d_\text{mean}$ & Dice &$HD$ & Dice &$HD$ \\ [0.25ex]
        \specialrule{.05em}{-0.1em}{0em}
        
        Payer C. & 95.65 \tiny{(100.0)} & 4.27 \tiny{(3.29)} & \textbf{94.25} \tiny{(100.0)} & \textbf{4.80} \tiny{(3.37)} & 90.90 \tiny{(95.54)} & \textbf{6.35} \tiny{(4.62)} & \textbf{89.80} \tiny{(95.47)} & \textbf{7.08} \tiny{(4.45)}\\
        
        Lessmann N. & 89.86 \tiny{(100.0)} & 14.12 \tiny{(13.86)} & 90.42 \tiny{(100.0)} & 7.04 \tiny{(5.3)} & 85.08 \tiny{(94.25)} & 8.58 \tiny{(4.62)} & 85.76 \tiny{(93.86)} & 8.20 \tiny{(5.38)}\\
        
        Chen M. & \textbf{96.94} \tiny{(100.0)} & \textbf{4.43} \tiny{(3.7)} & 86.73 \tiny{(100.0)} & 7.13 \tiny{(3.81)} & \textbf{93.01} \tiny{(95.96)} & 6.39 \tiny{(4.88)} & 82.56 \tiny{(96.5)} & 9.98 \tiny{(5.71)}\\
        
        Amiranashvili T. & 71.63 \tiny{(100.0)} & 11.09 \tiny{(4.78)} & 73.32 \tiny{(100.0)} & 13.61 \tiny{(4.92)} & 67.02 \tiny{(90.47)} & 17.35 \tiny{(8.42)} & 68.96 \tiny{(91.41)} & 17.81 \tiny{(8.62)}\\
        
        Dong Y. & 62.56 \tiny{(60.0)} & 18.52 \tiny{(17.71)}  & 67.21 \tiny{(71.40)} & 15.82 \tiny{(14.18)} & 76.74 \tiny{(84.15)} & 14.09 \tiny{(11.10)} & 67.51 \tiny{(66.05)} & 26.46 \tiny{(28.18)}\\
        
        Angermann C. & 55.80 \tiny{(57.19)} & 44.92 \tiny{(15.29)} & 54.85 \tiny{(57.18)} & 19.83 \tiny{(16.79)} & 43.14 \tiny{(43.44)} & 44.27 \tiny{(35.75)} & 46.40 \tiny{(47.98)} & 41.64 \tiny{(36.27)}\\
        
        Kirszenberg A. & 0.0 \tiny{(0.0)} & 155.42 \tiny{(126.24)} & 0.0 \tiny{(0.0)} & 1000 \tiny{(1000.0)} & 13.71 \tiny{(0.01)} & 77.48 \tiny{(86.83)} & 35.64 \tiny{(0.09)} & 65.51 \tiny{(60.27)}\\
        
        Jiang T. & 89.82 \tiny{(100.0)} & 7.39 \tiny{(4.67)} & * & * & 82.70 \tiny{(92.62)} & 11.22 \tiny{(8.1)} & * & * \\
        
        Wang X. & 84.02 \tiny{(100.0)} & 12.40 \tiny{(8.13)} & * & * & 71.88 \tiny{(84.65)} & 24.59 \tiny{(18.58)} & * & *\\
        
        Brown K. & $\star$ & $\star$ &  * & * & 62.69 \tiny{(85.03)} & 35.90 \tiny{(29.58)} & * & * \\
        
        Hu Y. & $\star$ & $\star$ & $\star$ & $\star$ & 84.07 \tiny{(91.41)} & 12.79 \tiny{(11.66)} & 81.82 \tiny{(90.47)} & 29.94 \tiny{(20.33)}\\
        
        \hline
        
        Sekuboyina A. & 89.97 \tiny{(100.0)} & 5.17 \tiny{(3.96)} & 87.66 \tiny{(100.0)} & 6.56 \tiny{(3.6)} & 83.06 \tiny{(90.93)} & 12.11 \tiny{(7.56)} & 83.18 \tiny{(92.79)} & 9.94 \tiny{(7.22)}\\
        
        \specialrule{.1em}{0em}{0em}
        \end{tabular}
    }
    \caption{\textsc{VerSe}`19}
    \label{tab:verse19}
    \end{subtable}
    
    \medskip
        
 \begin{subtable}{\linewidth}\centering
 {
        \begin{tabular}{ c | c : c | c : c | c : c | c : c}
        \specialrule{.1em}{0em}{-.1em}
        \multirow{3}{*}{\rule{0pt}{2.5ex}\textbf{Team}} & \multicolumn{4}{c}{Labelling} & \multicolumn{4}{c}{Segmentation} \\[-1ex] 
        
        & \multicolumn{2}{c}{\textsc{Public}} & \multicolumn{2}{c}{\textsc{Hidden}} & \multicolumn{2}{c}{\textsc{Public}} & \multicolumn{2}{c}{\textsc{Hidden}}\\
        & $id.rate$ & $d_\text{mean}$ & $id.rate$ & $d_\text{mean}$ & Dice &$HD$ & Dice &$HD$ \\ [0.25ex]
        \specialrule{.05em}{-0.1em}{0em}
        
         Chen D. & \textbf{95.61} \tiny{(100.0)} & \textbf{1.98} \tiny{(0.65)}  &   \textbf{96.58} \tiny{(100.0)} & \textbf{1.38} \tiny{(0.59)} & \textbf{91.72} \tiny{(95.52)}  & 6.14 \tiny{(4.22)} &  \textbf{91.23} \tiny{(95.21)}  & 7.15 \tiny{(4.30)} \\
        
        Payer C. & 95.06 \tiny{(100.0)} & 2.90 \tiny{(1.62)}  &   92.82 \tiny{(100.0)} & 2.91 \tiny{(1.54)}  & 91.65 \tiny{(95.72)}  & \textbf{5.80} \tiny{(4.06)} &  89.71 \tiny{(95.65)}  &  \textbf{6.06} \tiny{(3.94)} \\

         Zhang A. & 94.93 \tiny{(100.0)} & 2.99 \tiny{(1.49)}  &   96.22 \tiny{(100.0)} & 2.59 \tiny{(1.27)}  & 88.82 \tiny{(92.90)}  & 7.62 \tiny{(5.28)} &  89.36 \tiny{(92.77)}  &  7.92 \tiny{(5.52)} \\
             
        Yeah T. & 94.97 \tiny{(100.0)} & 2.92 \tiny{(1.38)}  &   94.65 \tiny{(100.0)} & 2.93 \tiny{(1.29)}  & 88.88 \tiny{(92.93)}  & 9.57 \tiny{(5.43)} &  87.91 \tiny{(92.76)}  & 8.41 \tiny{(5.91)}  \\
        
        Xiangshang Z. & 75.45 \tiny{(92.86)} & 22.75 \tiny{(5.88)}  &   82.08 \tiny{(93.75)} & 17.09 \tiny{(4.79)} & 83.58 \tiny{(92.69)}  & 15.19 \tiny{(9.76)} &  85.07 \tiny{(93.29)}  &  12.99 \tiny{(8.44)} \\
        
        Hou F. & 88.95 \tiny{(100.0)} & 4.85 \tiny{(1.97)}  &  90.47 \tiny{(100.0)} & 4.40 \tiny{(1.97)} & 83.99 \tiny{(90.90)}  & 8.10 \tiny{4.52} &  84.92 \tiny{(94.21)}  &   8.08 \tiny{(4.56)} \\
        
        Zeng C. & 91.47 \tiny{(100.0)} & 4.18 \tiny{(1.95)}  &  92.82 \tiny{(100.0)} & 5.16 \tiny{(2.17)} & 83.99 \tiny{(90.90)}  & 9.58 \tiny{6.14} &  84.39 \tiny{(91.97)}  &  8.73 \tiny{(5.68)} \\
        
        Huang Z. & 57.58 \tiny{(62.5)} & 19.45 \tiny{(15.57)}  &  3.44 \tiny{(0.0)} & 204.88 \tiny{(155.75)}  & 80.75 \tiny{(88.83)}  & 34.06 \tiny{(27.36)} &  81.69 \tiny{(89.85)}  & 15.75 \tiny{(11.58}  \\
        
        Netherton T. & 84.62 \tiny{(100.0)} & 4.64 \tiny{(1.67)}  &  89.08 \tiny{(100.0)} & 3.49 \tiny{(1.6)}  & 75.16 \tiny{(86.74)}  & 13.56 \tiny{(6.8)} &  78.26 \tiny{(87.44)}  & 14.06 \tiny{(7.05)}  \\
        
        Huynh L. & 81.10 \tiny{(88.23)} & 10.61 \tiny{(5.66)}  &  84.94 \tiny{(90.91)} & 10.22 \tiny{(4.93)}  & 62.48 \tiny{(66.02)}  & 20.29 \tiny{(16.23)} &  65.23 \tiny{(69.75)}  & 20.35 \tiny{(16.48)}  \\
        
        Jakubicek R.$^\dagger$ & 63.16 \tiny{(80.0)} & 17.01 \tiny{(13.73)} & 49.54 \tiny{(56.25)} & 16.59 \tiny{(13.87)}   & 73.17 \tiny{(85.15)}  & 17.26 \tiny{(12.80)} &  52.97 \tiny{(63.56)}  & 20.30 \tiny{(19.45)} \\
        
        Mulay S. & 9.23 \tiny{(0.0)} & 191.02 \tiny{(179.26)}  &   *  & *   &  58.18 \tiny{(64.96)}  & 99.75 \tiny{(95.60)} &  *  & *  \\
        
        Paetzold J. & $\star$ & $\star$ & $\star$ & $\star$ & 10.60 \tiny{(4.79)} & 166.55 \tiny{(265.16)}  &  25.49 \tiny{24.55} & 240.61 \tiny{191.29}  \\
        
        \hline
        
        Sekuboyina A. & 82.68 \tiny{(93.75)} & 6.66 \tiny{(3.87)}  &  86.06 \tiny{100.0} & 5.71\tiny{(3.51)}  & 78.05 \tiny{(85.09)}  & 10.99 \tiny{(6.38)} &  79.52 \tiny{(85.49)}  & 11.61 \tiny{(7.76)}  \\
        
        \specialrule{.1em}{0em}{0em}
        \end{tabular}
    }
    \caption{\textsc{VerSe}`20}
    \label{tab:verse20}
    \end{subtable}
        
\caption{Benchmarking \textsc{VerSe}: Overall performance of the submitted algorithms for the tasks of labelling and segmentation over the two test phases. The table reports mean and median (in brackets) measures over the dataset. The teams are ordered according to their Dice scores on the \textsc{Hidden} set. Dice and $id.rate$ are reported in \% and $d_\text{mean}$ and $HD$ in mm. $\star$ indicates that the team's algorithm did not predict the vertebral centroids. * indicates a non-functioning docker container. $\dagger$ Jakubicek R. submitted a semi-automated method for \textsc{Public} and a fully automated docker for \textsc{Hidden}.}
\label{tab:verse_overall}
\end{table*}

\begin{figure*}[!htbp]
    \centering
         \includegraphics[width=0.95\textwidth]{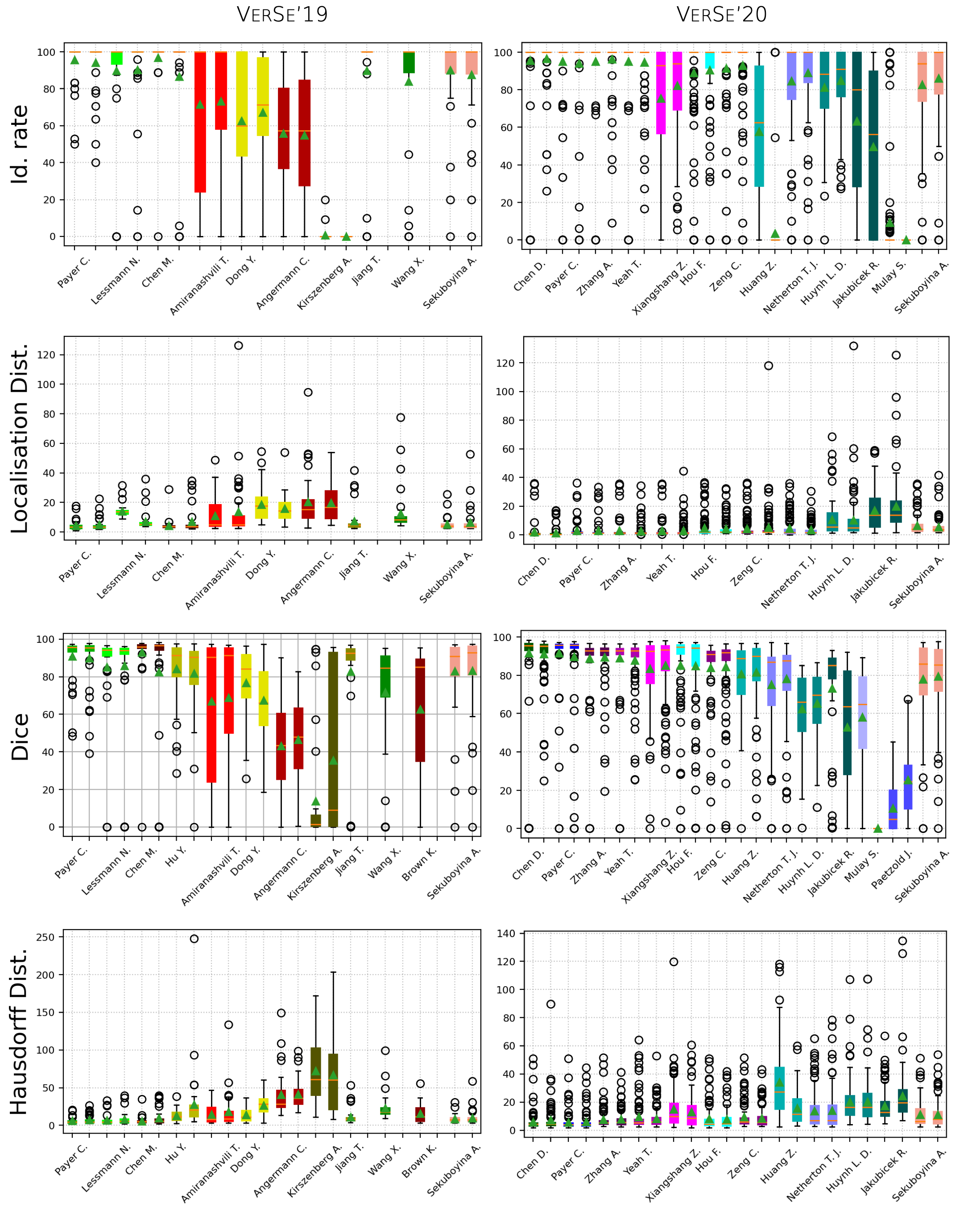}
    \caption{\textbf{Overall performance}: Box plots comparing all the submissions on the four performance metrics. The plots also show the mean (green triangle) and median (orange line) values of each measure. The two boxes for every team correspond to the performance on the \textsc{Public} and \textsc{Hidden} data. Note that Dice and $id.rate$ are on a scale of 0 to 1 while Hausdorff distance ($HD$) and localisation distance (d$_{mean}$) are plotted in mm.  }
\label{fig:teamwise_results}
\end{figure*}

\subsection{Overall performance of the algorithms}
The overall performance of the evaluated algorithms for \textsc{VerSe}`19 and `20 is reported in Tables \ref{tab:verse19} and \ref{tab:verse20}, respectively. We report the mean and the median values of all four evaluation metrics, viz. identification rate ($id.rate$) and localisation distance ($d_\text{mean}$) for the labelling task and Dice and Hausdorff distance ($HD$) for segmentation. Note that the algorithms are arranged according to their performance on the corresponding challenge leaderboards. Of the evaluated algorithms in \textsc{VerSe`19}, the highest $id.rate$ and Dice in the \textsc{Public} phase were 96.9\% and 93.0\%, both by Chen M. On the \textsc{Hidden} data, these are 94.3\% and 89.8\%, by Payer C. Similarly, for \textsc{VerSe`20}, Chen D. achieved the highest mean $id.rate$ and Dice on both the test phases: 95.6\% and 91.7\% in \textsc{Public} and  96.6\% and 91.2\% in \textsc{Hidden} phase. Fig. \ref{fig:teamwise_results} illustrates the mean and other statistics pertaining to the algorithms' performance as box plots for the four evaluation metrics. Of importance: At least four methods in \textsc{VerSe}`19 achieve a median $id.rate$ of 100\%. In \textsc{VerSe}`20, this is achieved by seven teams, a majority of the submissions.

\begin{table}[t!]
\setlength{\tabcolsep}{0.4em}
\scriptsize
\renewcommand{\arraystretch}{1}

        \begin{tabular}{ c c | c c : c c}
        \specialrule{.1em}{0em}{-.1em}
        & \multirow{3}{*}{\rule{0pt}{2.5ex}\textsc{VerSe}} & \multicolumn{2}{c}{\textsc{Public}} & \multicolumn{2}{c}{\textsc{Hidden}}\\
        & & All & Top-5 & All & Top-5\\[-0.1ex] 
        \specialrule{.05em}{-0.1em}{0em}
        
        \parbox[t]{1mm}{\multirow{2}{*}{\rotatebox[origin=c]{90}{$id.rate$}}} & 2019 & 61.4$\pm$44.5 & 83.3$\pm$30.7 & 61.6$\pm$43.6 & 82.4$\pm$31.6 \\
        & 2020 & 72.5$\pm$39.3 & 93.9$\pm$21.0 & 68.6$\pm$42.1 & 94.4$\pm$17.5 \\
        \hline
        \parbox[t]{1mm}{\multirow{2}{*}{\rotatebox[origin=c]{90}{Dice}}} & 2019 & 71.2$\pm$33.7 & 82.5$\pm$25.9 & 71.3$\pm$32.6 & 78.9$\pm$28.4 \\
        & 2020 & 75.2$\pm$28.5 & 89.3$\pm$17.9 & 71.1$\pm$32.2 & 88.8$\pm$16.7 \\
        \specialrule{.1em}{0em}{0em}
        \end{tabular}
        
\caption{Mean performance ($id.rate$ and Dice, in \%) of all the evaluated algorithms in both the \textsc{VerSe} iterations. `Top-5' indicates that the mean was computed on the five top-performing algorithms in that year's leaderboard. `All' considers all submitted algorithms.}\label{tab:mean_perf_annual}
\end{table}

Table~\ref{tab:mean_perf_annual} provides a bigger picture, reporting the mean performance of all the evaluated algorithms as well as the five top-performing algorithms. In 2019, the performance of all methods (incl. Top 5) is consistent between the \textsc{Public} and \textsc{Hidden} phases, except for a slight drop in Dice in 2019's \textsc{Hidden} phase. However, in 2020, we see that the mean performance of all teams drops, while that of only the top-5 stays relatively consistent. Additionally, observe that the mean $id.rate$ and Dice score increased from 2019 to 2020 (for both \emph{All} and \emph{Top-5}). These observations can be attributed to: 1) Supervised algorithms fail to generalise to out-of-distribution cases (L6 in \textsc{VerSe}`19) when their percentage of occurrence in the dataset is consistent with their low clinical prevalence. 2) With the availability of large, public data with an over-representation of out-of-distribution cases (as in \textsc{VerSe}`20), makes better algorithm design and learning feasible.

\begin{figure*}[t!]
    \centering
         \includegraphics[width=0.83\textwidth]{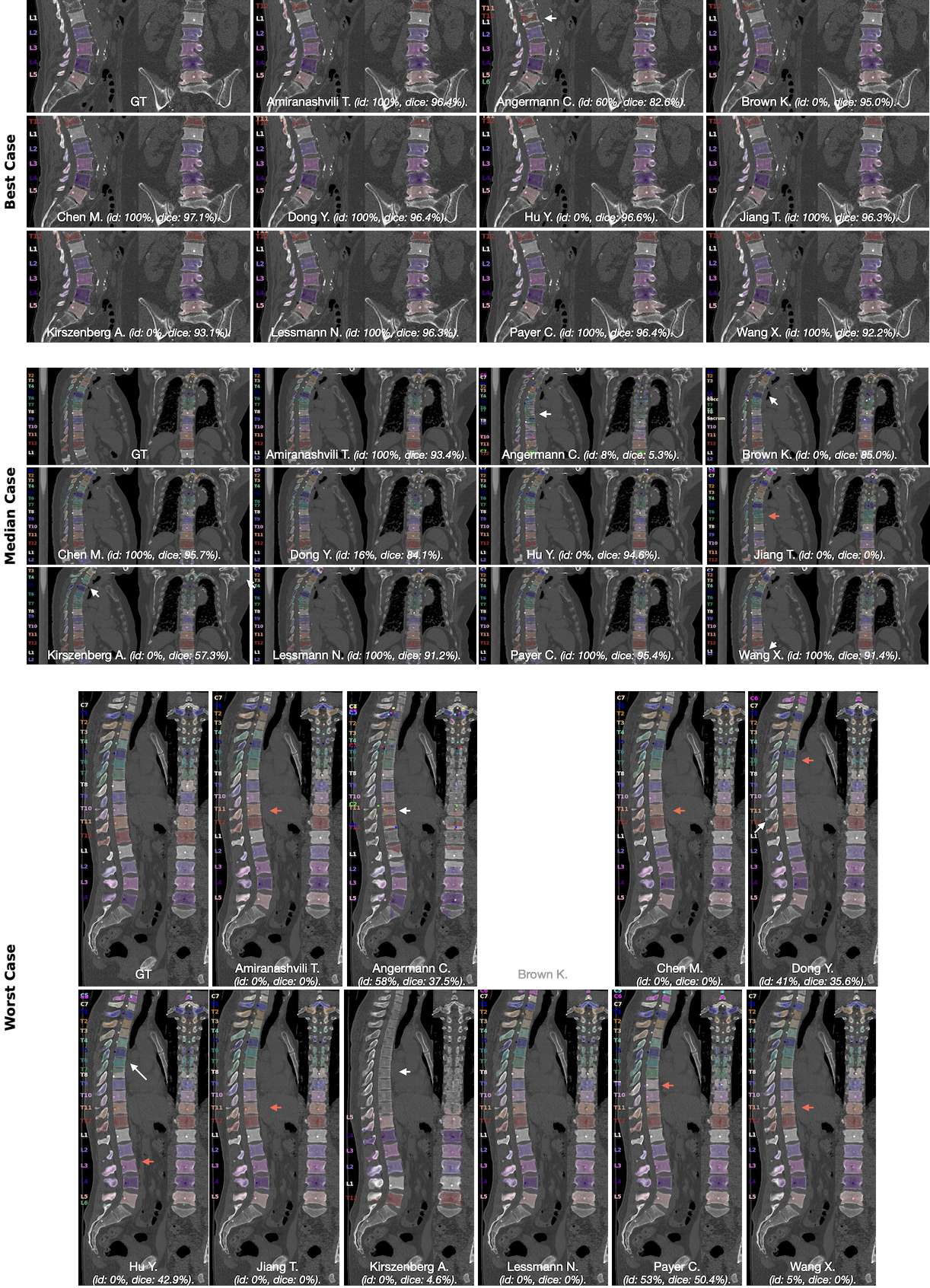}
    \caption{\textsc{VerSe}`19: Qualitative results of the participating algorithms on the \emph{best}, \emph{median}, and \emph{worst} cases, determined using the mean performance of the algorithms on all cases. We indicate erroneous predictions with arrows. A red arrow indicates mislabelling with a \emph{one-label shift}. From Brown K., the prediction for the worst case was missing.}
\label{fig:qual_verse19}
\end{figure*}

\begin{figure*}[t!]
    \centering
         \includegraphics[width=0.84\textwidth]{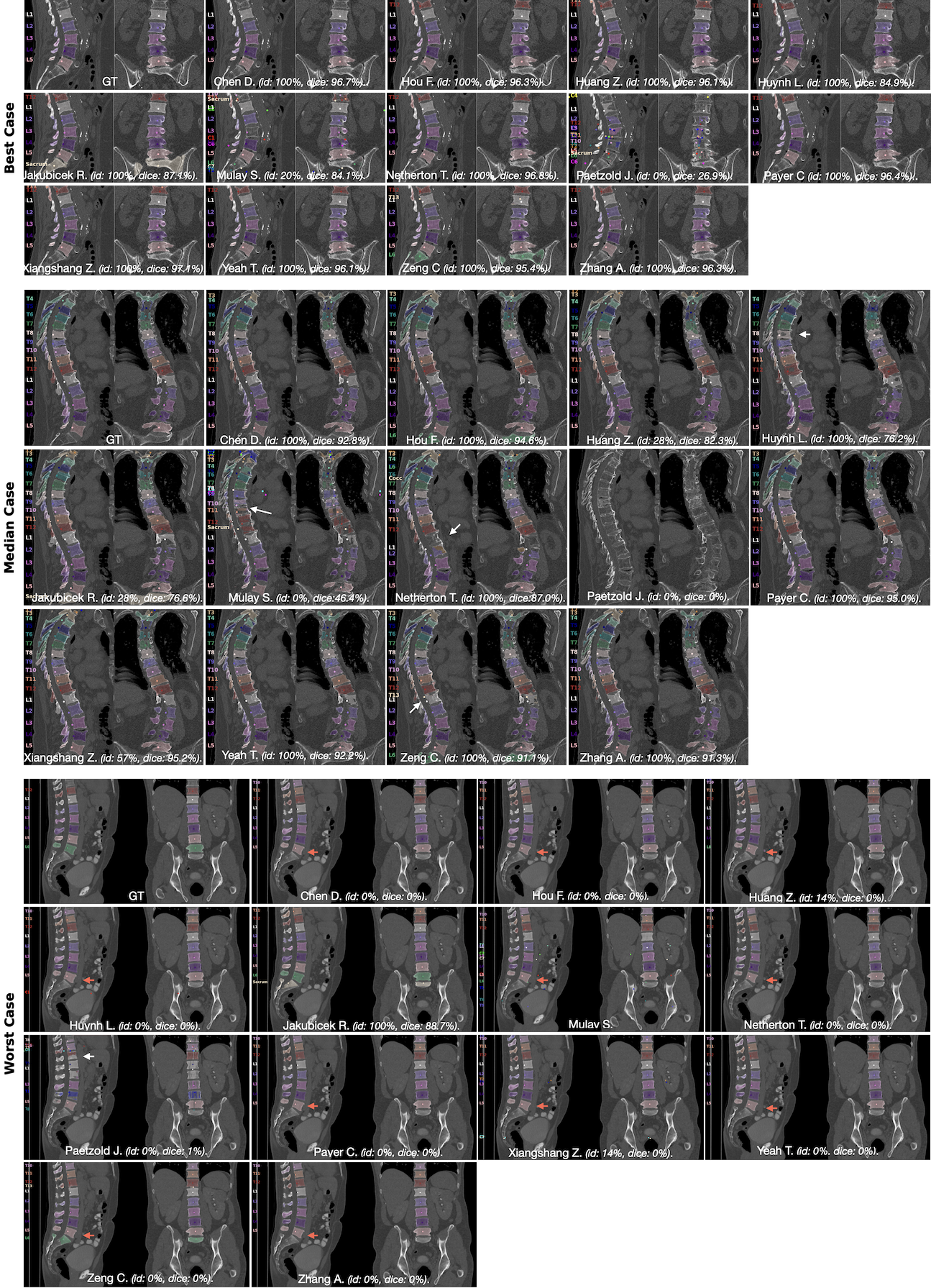}
    \caption{\textsc{VerSe}`20: Qualitative results of the participating algorithms on the \emph{best}, \emph{median}, and \emph{worst} cases, determined using the mean performance of the algorithms on all cases. We indicate erroneous predictions with arrows. A red arrow indicates mislabelling with a \emph{one-label shift}}
\label{fig:qual_verse20}
\end{figure*}

In Figs.~\ref{fig:qual_verse19} and \ref{fig:qual_verse20}, we show predictions of the algorithms on the \emph{best}, \emph{median}, and \emph{worst} scans, ranked by the average performance of all the algorithms on every scan. In \textsc{VerSe}`19, the \emph{best} scan, a lumbar FoV, is segmented correctly by all the algorithms. The \emph{median} scan, a thoracic FoV with a fracture, is erroneously segmented by a few teams, due to mislabelling (Jiang T., Kirszenberg A., and Wang X.) or stray segmentation (Angermann C., Brown K. and Dong Y.). The \emph{worst}-case scan, interestingly, is an anomalous one, wherein L5 is absent. Seemingly, the lumbar-sacral junction is a strong anatomical pointer for labelling and hence almost every algorithm wrongly labels an L4 as an L5. Medical experts, on the other hand, use the last rib (attached to T12) to identify the vertebrae and hence would arrive at the correct spine labels. Similarly, in \textsc{VerSe}`20, the \emph{best} case is a lumbar scan. The \emph{median} case is a thoracolumbar scan with severe scoliosis. In spite of this, the majority of the algorithms identify and segment the scan correctly. The \emph{worst} case again occurs due to an anomaly at the lumbar-sacral junction, here due to the presence of a transitional L6 vertebra. Interestingly, the semi-automated approach of Jakubicek R. succeeds in identifying this anomaly correctly.

\begin{figure*}[!t]
    \centering
         \includegraphics[width=0.98\textwidth]{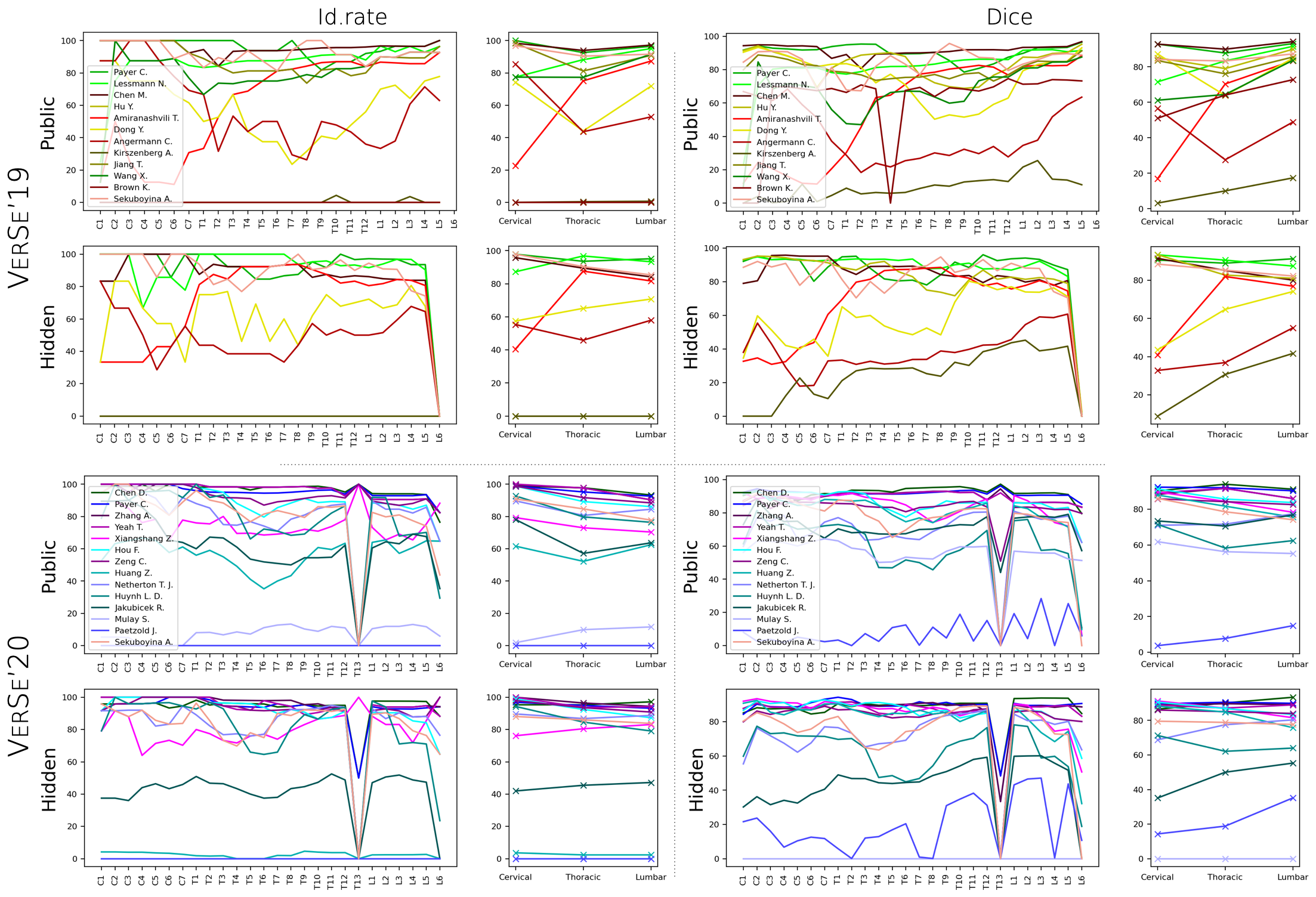}
    \caption{\textbf{Vertebra-wise and region-wise performance}: Plot shows the mean labelling and segmentation performance of the submitted algorithms at a vertebra level (left) and at a spine-region level (right), viz. cervical, thoracic, and lumbar regions.}
\label{fig:teamwise_results_region_and_vert}
\end{figure*}

\subsection{Vertebrae-wise and region-wise evaluation}
\label{subsec:regionwise}
In Fig.~\ref{fig:teamwise_results_region_and_vert}, we illustrate the mean labelling at segmentation capabilities of the submitted methods at a vertebra-level and region-level (cervical, thoracic, and lumbar). 

At a vertebra level, we observe a sudden performance drop in the case of transitional vertebrae (T13 and L6). Concerning L6, None of the methods in \textsc{VerSe}’19 identified the presence of L6. However, in \textsc{VerSe}`20, almost all algorithms identify at least a fraction of the L6 vertebrae. On the other hand, for T13, except for Xiangshang Z., the identification rate widely varies between the \textsc{Public} and \textsc{Hidden} phases for all teams.

Looking at the region-specific performance, \textsc{VerSe}`19 shows a trend of performance-drop in the thoracic region. This could be expected as mid-thoracic vertebrae have a very similar appearance, making them indistinguishable without external anatomical reference. Of course, such a reference (as T12/L1 or C7/T1 junctions) was present in all scans, but apparently not considered by most algorithms. This drop is not observed in \textsc{VerSe}`20. We hypothesise this to be a consequence of better algorithm design because the condition of identifying transitional vertebrae required accurate identification at a local level and reliable aggregation of labels at a global level. We further investigate this behaviour in the following sections.   

% \begin{figure}[!t]
%     \centering
%          \includegraphics[width=0.9\textwidth]{figures/scanwise_fractions.png}
%     \caption{\textbf{Performance at scan-level}: Fraction of scans from the dataset, $n$, with an $id.rate$ or Dice score higher than a threshold, $\tau$. The fraction is computed over scans in both the test phases.}
% \label{fig:success_percentile}
% \end{figure}

\begin{figure*}[t]
    \centering
  
  \begin{subfigure}[b]{0.48\textwidth}
        \includegraphics[width=\textwidth]{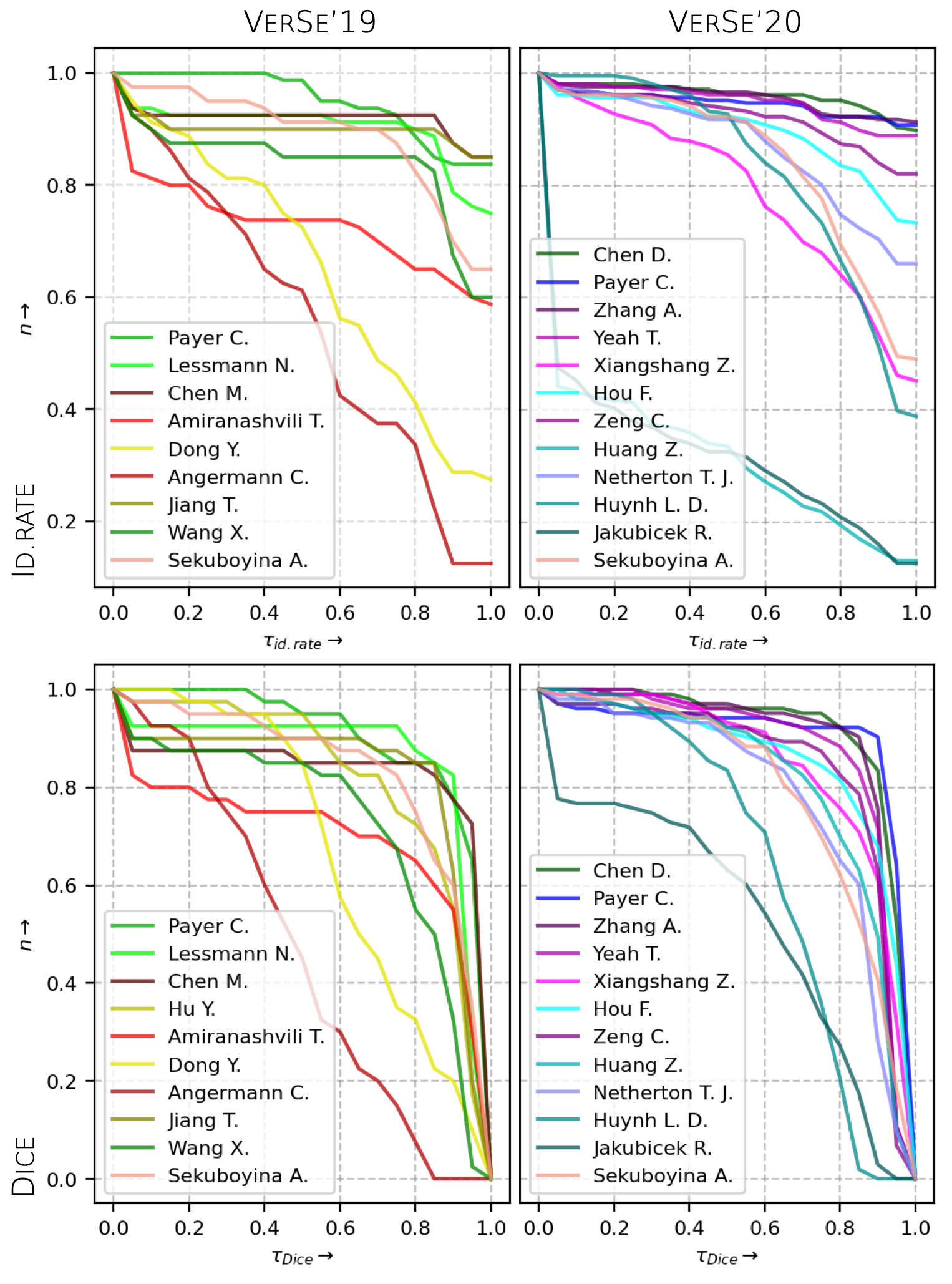}
        \caption{Performance at scan level}
    \end{subfigure}
  ~
     \begin{subfigure}[b]{0.48\textwidth}
        \includegraphics[width=\textwidth]{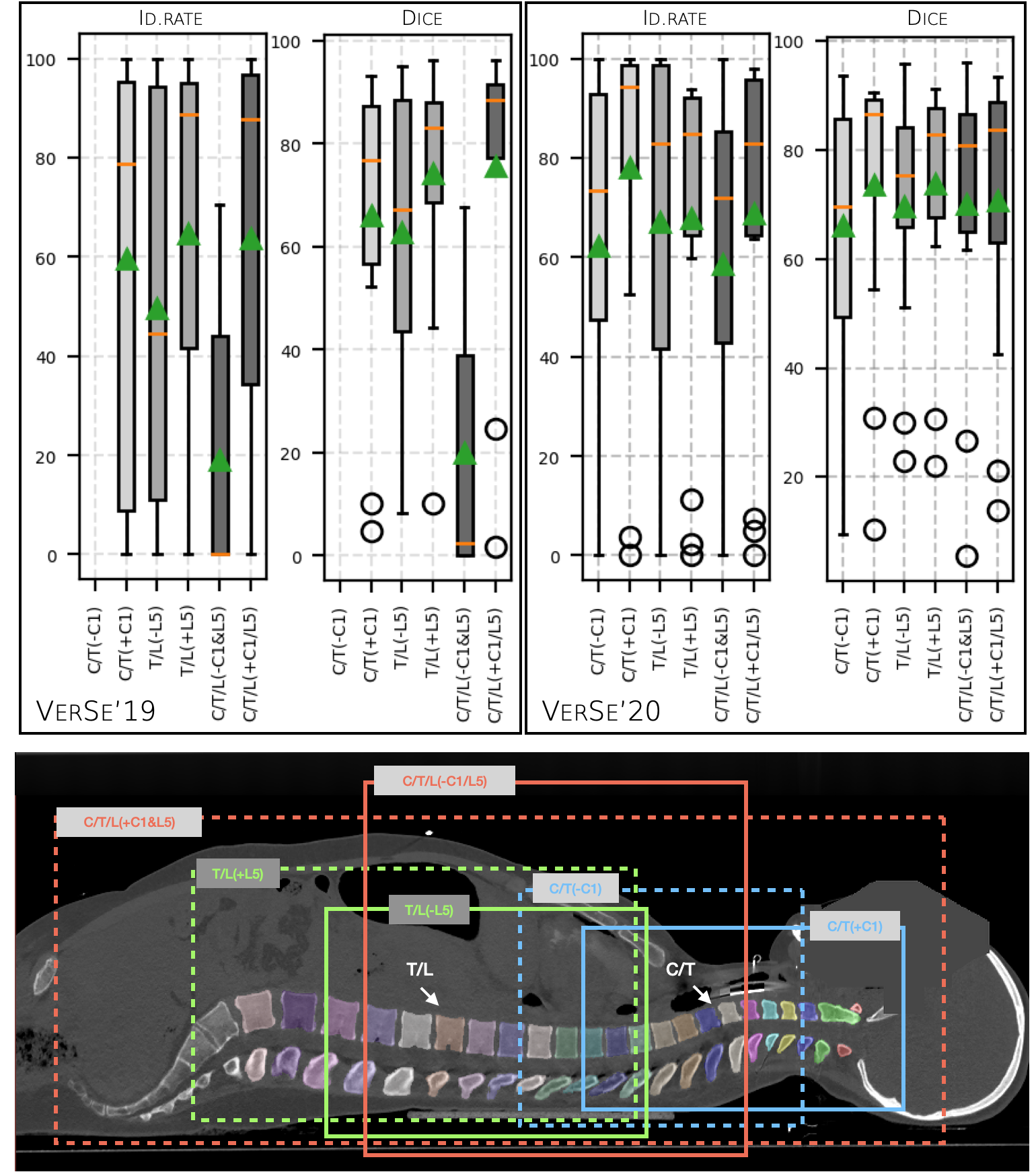}
        \vspace{1em}
        \caption{Effect of the field of view}
    \end{subfigure}

    \caption{(a) Fraction of scans, $n$, with an $id.rate$ or Dice higher than a threshold, $\tau$. The fraction is computed over scans in both the test phases. Uninformative dockers with lines hugging the axes are not visualised (Kirszenberg A., Brown K., Mulay S., and Paetzold J.). Hu Y. is not included in the $id.rate$ experiment due to missing centroid predictions. (b) Performance measures of scans grouped according to their field of view. Scans are binned into six categories of FoVs. Please refer to Sec.~\ref{subsec:fov} for details.}
	\label{fig:scan_level_and_fov_effect}
	
\end{figure*}

\begin{table*}[t]
\setlength{\tabcolsep}{0.23em}
\scriptsize
\renewcommand{\arraystretch}{1}
	\begin{tabular}{ c c | c | c | c | c | c | c | c | c | c | c | c | c || c | c | c | c | c | c | c | c | c | c | c | c | c | c}
	
% 	\specialrule{.1em}{0em}{0.05em}
     & \multicolumn{13}{c}{\textsc{VerSe}`19} & \multicolumn{14}{c}{\textsc{VerSe}`20} \\
     
	  & $<5\%$ & \rot{Payer C.} & \rot{Lessmann N.} & \rot{Chen M.} & \rot{Amiranashvili T.} & \rot{Dong Y.} & \rot{Angermann C.} & \rot{Kirszenberg A.} & \rot{Jiang T.} & \rot{Wang X.} & \rot{Brown K.} & \rot{Hu Y.}  &  \rot{Sekuboyina A.} & \rot{Chen D.} & \rot{Payer C.} & \rot{Zhang A.} & \rot{Yeah T.} & \rot{Xiangshang Z.} & \rot{Hou F.} & \rot{Zeng C.} & \rot{Huang Z.} & \rot{Netherton T.} & \rot{Huynh L.} & \rot{Jakubicek R.} &  \rot{Mulay S.} & \rot{Paetzold J.} &  \rot{Sekuboyina A.}  \\ 
	\specialrule{.1em}{0em}{0.05em}
	\textsc{Public} & $id.rate$ & 0 & 3 & 1 & 6 & 3 & 2 & 38 & 3 & 3 & -- & -- & 1 & 4 & 3 & 4 & 4 & 5 & 5 & 4 & 16 & 4 & 1 & 77 & 82 & -- & 3\\
	& Dice & 0 & 3 & 1 & 7 & 0 & 2 & 28 & 4 & 4 & 8 & 0 & 1 & 3 & 2 & 3 & 3 & 1 & 4 & 3 & 1 & 4 & 1 & 16 & 6 & 52 & 2 \\ 
	\hline
	\textsc{Hidden} & $id.rate$ & 0 & 2 & 4 & 8 & 1 & 4 & 40 & -- & -- & -- & -- & 1 & 0 & 3 & 0 & 0 & 0 & 3 & 2 & 99 & 2 & 0 & 31 & -- & -- & 3\\ 
	& Dice & 0 & 3 & 5 & 7 & 0 & 1 & 14 & -- & -- & -- & 1 & 1 & 0 & 3 & 0 & 0 & 1 & 3 & 3 & 0 & 2 & 0 & 23 & -- & 6 & 1\\
	\specialrule{.1em}{0em}{0em}
	\end{tabular}
\caption{\small \small Number of scans in each subset of \textsc{VerSe} with an $id.rate$ or Dice score less than 5\%. Reported values are absolute number of scans from a maximum of: 40 scans each for \textsc{VerSe}`19's \textsc{Public} and \textsc{Hidden} sets, and 103 scans each for \textsc{VerSe}`20's test sets.}
\label{tab:success_failure}
\end{table*}

\subsection{Labelling and segmentation at a scan level}
When an algorithm is deployed in a clinical setting, minimal manual intervention is desired. Therefore, it is of interest to peruse the \emph{effort} needed for correction. As a proxy, we analyse the number of scans in the dataset that were \emph{successfully} processed. We define \emph{success} using a threshold $\tau$, wherein a scan is said to be successfully \emph{identified} if its $id.rate$ is above $\tau_{id.rate}$. Similarly, successful segmentation is defined using $\tau_{Dice}$. The fraction of scans successfully processed is denoted by $n$. In Fig.~\ref{fig:scan_level_and_fov_effect}a, we show the behaviour of $n$ at varying thresholds. The best-case scenario for both the tasks is $n=1, \forall\tau$. The methods in \textsc{VerSe}`20 are closer to this behaviour than \textsc{VerSe}`19, the latter showing more spread over the grid. Especially, Chen D., Payer C., Zhang A., and Yeah T. perfectly identify ($id.rate$=100\%) close to 90\% of the scans. In 2019, this number was closer to 80\% for Chen M., Payer C., and Jiang T. Looking at the Dice curves in 2020, given a vertebra is labelled correctly, its segmentation seems trivial, with the majority of the methods attaining scores of 80-90\% on at least 80\% of the scans. In 2019, only three methods indicate this performance. 

Looking specifically at `failed' scans, we log the number of scans which resulted in less than 5\% $id.rate$ or Dice in Table~\ref{tab:success_failure}. When seen in tandem with Fig.~\ref{fig:teamwise_results}, this table provides an idea of scan-level failures. Interestingly, in \textsc{VerSe}`20, numerous methods do not show absolute failure in the \textsc{Hidden} phase, e.g. Chen D., Zhang A., Yeah T., and Huynh L.   

% \begin{figure}[!t]
%     \centering
%          \includegraphics[width=0.99\textwidth]{figures/fov_measures.png}
%     \caption{\textbf{Effect of FoV}: Performance measures one scans grouped according to their field-of-view (FoV). Scan are binned into the following six categories of FoVs. Please refer to the text for the details.}
% \label{fig:fov_effect}
% \end{figure}

\subsection{Effect of field of view on performance}
\label{subsec:fov}
Delving deeper into the region-wise performance of the methods, we ask the question: \emph{What landmark in a scan most aids labelling and segmentation?}. For this, we identify four landmarks on the spine: the cranium (if C1 exists), the cervico-thoracic junction (if C7 and T1 coexist), the thoraco-lumbar junction (if T12/T13 and L1 coexist), and lastly the sacrum (if L5 or L6 exists). Based on this, we divide the scans into six categories, namely:
\begin{enumerate}
    \itemsep0em 
    \item $C/T(+C1)$: Cranium and the cervico-thoracic junction are present. Thoraco-lumbar junction absent.
    \item $C/T(-C1)$: Cervico-thoracic junction present. Thoraco-lumbar junction absent.
    \item $T/L(+L5)$: Sacrum and the thoraco-lumbar junction are present. Cervico-thoracic junction absent. 
    \item $C/T(-L5)$: Thoraco-lumbar junction present. Sacrum and cervico-thoracic junction absent.
    \item $C/T/L(+C1\&L5)$: Full spines. Both cervico-thoracic and thoraco-lumbar junctions are present.
    \item $C/T/L(-C1/L5)$: Cervico-thoracic and thoraco-lumbar junctions are present. Either cranium or both cranium and sacrum are absent. (\textsc{VerSe} did not contain any scan with cranium and without sacrum)
\end{enumerate}
Note that in the categories above, L5 refers to the last lumbar vertebra, which could be L4 or L6 as well. Fig.~\ref{fig:scan_level_and_fov_effect}b shows an example of a full spine scan with crops that would fall into one of these categories. Once every scan in the dataset is assigned the appropriate category, we compute the mean identification rate and Dice score of every method for every category (cf. Fig.~\ref{fig:scan_level_and_fov_effect}b). In \textsc{VerSe}`19, we observe that scans with all lumbar vertebra are easier to process compared to cervical ones ($T/L$ or $C/T/L$ with L5). For a similar FoV, we see a large drop when cases do not contain L5 or C1. This shows the reliance of the \textsc{VerSe}`19 methods on the cranium and sacrum. Interestingly, the reliance on L5 is not as drastic in \textsc{VerSe}`20 (refer to categories $-C1\&L5$ and $-L5$). However, the cranium seems to still be a strong reference. Essentially, the median segmentation performance (Dice) coefficient of the methods is $\sim80\%$ in thoracic and lumbar regions for a variety of FoVs, where at least one of the four landmarks mentioned above is visible. Nonetheless, for cervical (-thoracic) scans, there is room for improvement for FoVs without the cranium.

\begin{figure}[!t]
    \centering
         \includegraphics[width=0.8\textwidth]{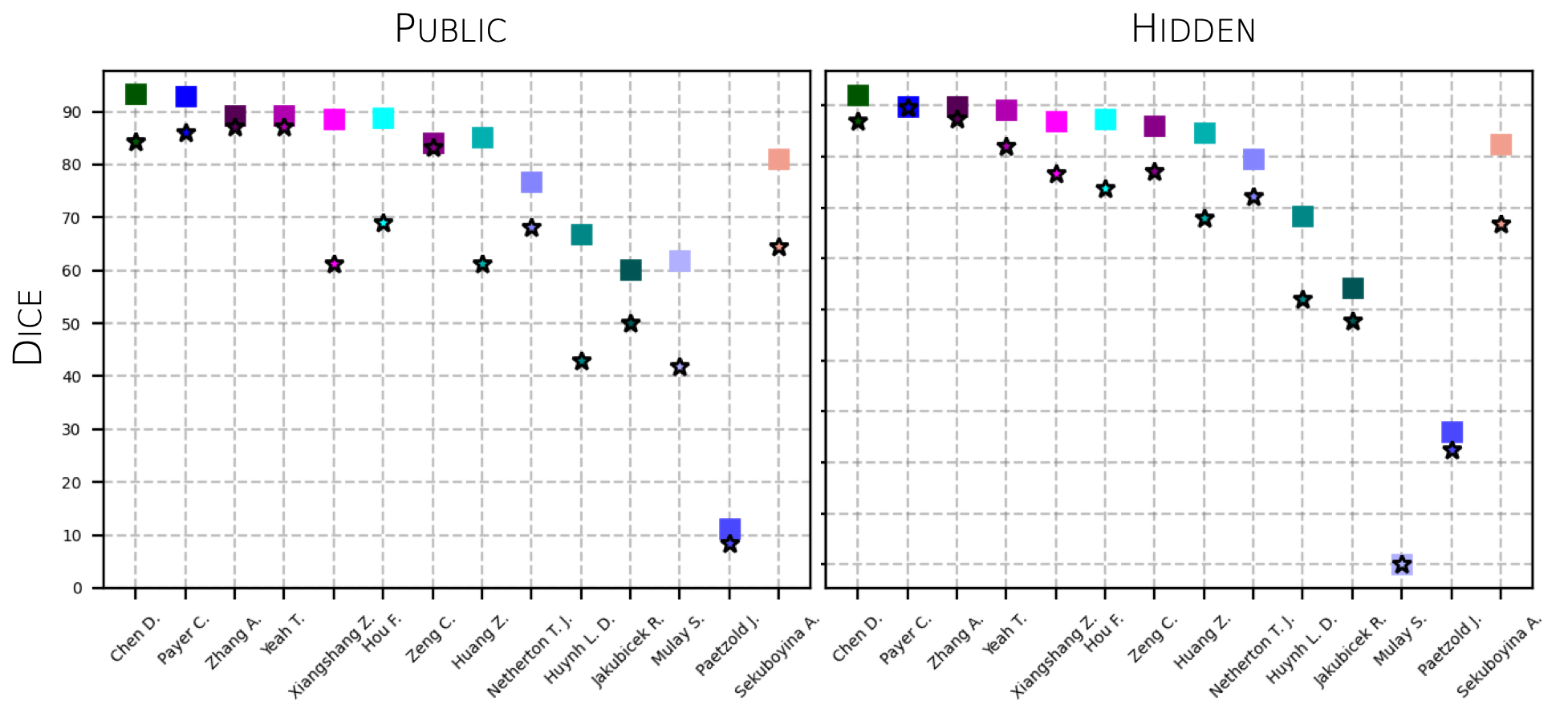}
    \caption{\textbf{Performance on transitional vertebrae}: Dice scores of the \textsc{VerSe}`20 algorithms computed on anatomically rare scans with transitional vertebrae ($\star$), i.e. T13 and L6, and the normal scans without them ($\blacksquare$).}
\label{fig:transitional_analysis}
\end{figure}

\subsection{Performance on anatomically rare scans vs. normal scans}
As stated earlier, \textsc{VerSe}`20 was rich in rare anatomical anomalies in the form of transitional vertebrae, viz. T13 and L6. In Fig~\ref{fig:transitional_analysis}, we illustrate the difference in performance of the submitted algorithms between a normal scan and a scan with transitional vertebrae. As expected, we observe a superior performance on normal anatomy when compared to that on rare anatomy. The difference in performance, however, is of interest. In \textsc{Public}, Yeah T., Zhang A., and Zeng C. have a small drop in performance, with the first two approaches showing a better performance on the rare cases compared to the two top performers, Payer C. and Chen D. In \textsc{Hidden}, Payer C. does not show any drop in performance, and outperforms the rest on the rare cases. Arguably, algorithms that either show a stable performance across anatomies or those that identify (and skip processing) a rare case are preferred in a clinical routine.

\begin{figure}
\begin{floatrow}
\capbtabbox[0.35\textwidth]{%
\scriptsize
       \begin{tabular}{ c  c }
        \specialrule{.1em}{0em}{-.1em}
        \multicolumn{2}{c}{V`19 approaches on V`20 data}\\
        \hline
        Payer C. & \textbf{85.21} \\
        Lessmann N. & 66.96 \\
        Chen M. & 65.21 \\
        \specialrule{.1em}{0em}{0em}
        \multicolumn{2}{c}{V`20 approaches on V`19 data}\\
        \hline
        Chen D. & \textbf{86.44} \\
        Payer C. & 84.11 \\
        Zhang A. & 85.42 \\
        \specialrule{.1em}{0em}{0em}
    \end{tabular}
    \vspace{5em}
    \label{tab:cross_verse}
    \caption{Mean Dice (\%) of running the three of the top-performing dockers of one \textsc{VerSe} iteration on \textsc{Hidden} set of the other iteration.}
    
}
{%
\label{tab:cross_verse}
%   \caption{Dice (in \%) of running the three of the top-performing dockers of one \textsc{VerSe} iteration on the \textsc{Hidden} set of the other iteration.}%
}
\hfill
\ffigbox[0.63\textwidth]{%
  \includegraphics[width=0.63\textwidth]{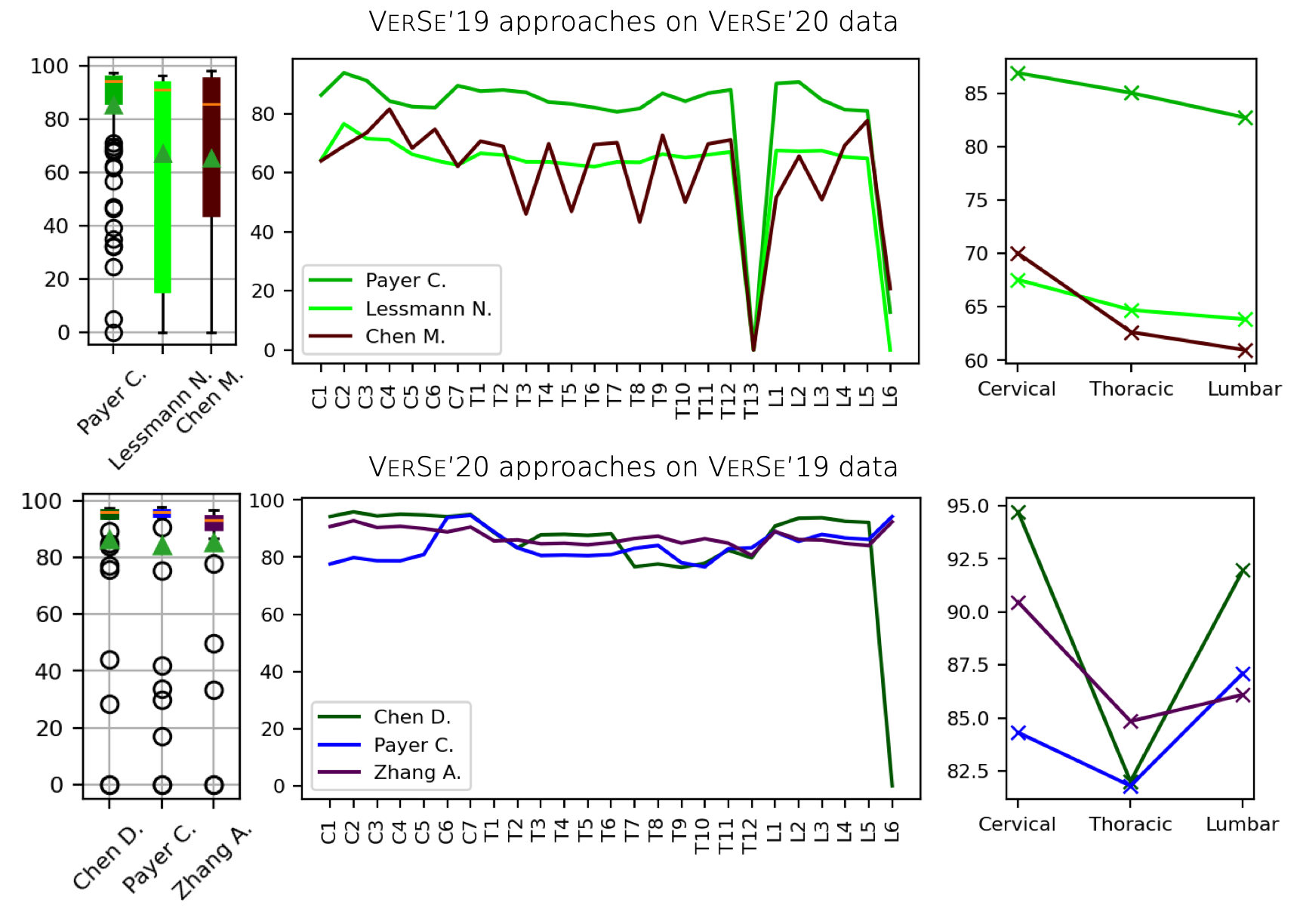}
  \label{fig:cross_verse}
  \caption{(Left) Teamwise overall Dice scores of the approaches from one \textsc{VerSe} iteration run on the \textsc{Hidden} set of the other iteration. (Center and right) Mean vertebrae-wise, and region-wise Dice scores of the same.}
}{%
  \label{fig:cross_verse}
}

\end{floatrow}
\end{figure}

\subsection{Generalisability of the algorithms}
Owing to the \textsc{Hidden} test phase in both iterations of \textsc{VerSe}, we have access to the docker containers that can be deployed on any spine scan. The only prerequisite for this being that the scan conforms to the Hounsfield scale (as in \textsc{VerSe} data). Exploring the dockers' ability at clinical translation, we deploy three of the top-performing dockers of \textsc{VerSe}`19 on the \textsc{Hidden} set of \textsc{VerSe}`20, and vice versa. Table~\ref{tab:cross_verse} and Fig.~\ref{fig:cross_verse} report the cross-iteration performance of these dockers. 

Recall that the \textsc{VerSe}`20 data has some overlap with \textsc{VerSe}`19. Therefore, the approaches trained on \textsc{VerSe}`20 perform reasonably well on the \textsc{VerSe}`19 data. There is a drop of $\sim3\%$, which can be attributed a domain shift between the datasets. Note that Payer C. and Zhang A. succeed in identifying L6, while none of the methods in 2019 do, owing to the over-representation of L6 in \textsc{VerSe}`20. This underpins our motivation for the second \textsc{VerSe} iteration. 

On the other hand, the setting of \textsc{VerSe}`19 methods on \textsc{VerSe}`20 data is more interesting. In addition to a domain shift (due to multi-scanner, multi-centre data in 2020), there are also unseen anatomies. Understandably, we see a drop in performance for Lessmann N. and Chen M. Interestingly, the performance drop is not as large for Payer C. This can be attributed to the way these approaches arrive at the final labels. Lessmann N. depends on identifying the last vertebra. In cases with L6, this affects the entire scan. We assume a similar behaviour for Chen M. In case of Payer C., the presence of L6 was not as detrimental, as the rest of the vertebrae were identified and segmented correctly and the final labels depended prediction confidences during the post-processing stage. Vertebra T13, however, can be ignored due to its absence in \textsc{VerSe}`19.

%%%%%%%%%%%%%%%%%%%%%%%%%%%%%%%%%%%%%%

\section{Discussion}
\label{section:discussion}

\subsection{Algorithm design}
In this section, we comment on the design of the submitted approaches. Brief descriptions of the evaluated algorithms are provided in Table~\ref{tab:teams}, Sec.~\ref{sec:methods}, and \ref{app:teams}. We look into the following design decisions: pure deep-learning (DL) \emph{vs.} hybrid models, 3D patch-based \emph{vs.} 2D slice-wise approach, and a single model \emph{vs.} a multi-staged approach.\\

\noindent
\textbf{Deep learning \emph{vs.} hybrid.} Out of the twenty-four algorithms benchmarked in this work, twenty-one are purely deep-learning-based, albeit with minor pre- (e.g. intensity-based filtering) and post-processing components (e.g. connected components or morphological operations). Three algorithms: \hyperref[desc:amiranashvilit]{Amiranashvili T.}, \hyperref[desc:kirszenberga]{Kirszenberg A}, and \hyperref[desc:jakubicekr]{Jakubicek R.} employ statistical shape models. The first two approaches use such models for identifying the vertebrae. The third approach uses it for segmentation using elastic registration. Unlike learning-based approaches, atlases incorporate reliable prior information, thus preventing anatomically implausible results. However, in this benchmark, we see a clear superiority of data-driven, DL approaches compared to the hybrid ones. This is understandable, given the size of \textsc{VerSe}. Better integration of shape-based and learning-based ones is of interest, thus enabling segmentation with anatomical guarantees.\\

\noindent
\textbf{3D patch-based \emph{vs.} 2D slice-wise segmentation.} Common among all the algorithms is the motivation that a clinical spine scan's size is large for current-generation GPU memory. We can draw two lines of algorithms among the benchmarked ones: First, those performing 2D slice-wise segmentation (e.g. \hyperref[desc:angermannc]{Angermann C.}, \hyperref[desc:kirszenberga]{Kirszenberg A.}, \hyperref[desc:supritim]{Mulay S.}, \hyperref[desc:paetzoldj]{Paetzold J.}). Second, which form the majority, are the approaches that perform patch-wise segmentation in 3D using architectures such as 3D U-Net \citep{cciccek20163d}, V-Net \citep{milletari2016v}, or nnU-Net \citep{isensee2019}. The second category can further be split into approaches performing multi-label segmentation, and those performing binary segmentation. 

Observe that, in general, 3D processing is preferable naive 2D slice-wise segmentation. More so, when compared to 2D slice-wise multi-label segmentation. This is expected because slice-wise processing, in spite of offering a larger FoV and memory efficiency, ignores crucial 3D context for an anatomically large structure such as a spine. Moreover, labelling the vertebrae becomes noisy as not every vertebra is visible in every slice.\\

\noindent
\textbf{Single model \emph{vs.} multi-staged.} 
One principal categorisation of the benchmarked algorithms is into two categories based on the number of stages they employ to tackle the tasks of labelling and segmentation, as demonstrated by some representative algorithms listed below: 

\begin{enumerate}
    \item Single-stage: \hyperref[desc:lessmannn]{Lessmann N.}, \hyperref[desc:jiangt]{Jiang T.}, \hyperref[desc:huangz]{Huang Z.}, and \hyperref[desc:huynhl]{Hu\`{y}nh D.}
    \item{Multi-staged:} \hyperref[desc:chend]{Chen D.}, \hyperref[desc:payerc]{Payer C.}, \hyperref[desc:AmberZhang]{Zhang A.}, and \hyperref[desc:nethertont]{Netherton T.}
\end{enumerate}

Typically, single-staged models work with 3D patches. The likes of Lessmann N. perform iterative identification and segmentation and determine a label arrangement using maximum likelihood estimation. Jiang T. and Huang Z. propose dedicated architectures with multiple heads, one each for the labelling and segmentation tasks, thus exploiting their interdependency. nnU-Net or 3D-UNet-based multi-label classification followed by final labelling is also a recurring theme. 

On the other hand, numerous sequential frameworks have also been proposed. Payer C., for instance, perform labelling and segmentation in three stages of localisation, then labelling, and finally binary vertebral segmentation. Zhang A. propose a four-stage approach involving spine-centerline detection, vertebral candidate prediction, and a three-class segmentation of the localised spine. Following this, final labels are identified based on certain spine-centric rules.

As evidenced by the performance, one cannot propose a `winner' among the two categories. Both categories equally span the upper regions of the leaderboards. The first category could possibly result in numerous inferences of large patches per scan (resulting in longer inference times), while the second approach could be prone to errors compounding from a preliminary stage of the sequence. 

\subsection{On rare anatomical variations: transitional vertebrae}
\textsc{VerSe}`19 included two cases with L6 in the train set, a proportion resembling its clinical occurrence. We observed that almost every algorithm fails to segment the one L6 in the \textsc{Hidden} set. A major motivation for the second iteration of \textsc{VerSe}, was hence, to increase number of anatomically anomalous cases. \textsc{VerSe}`20 included six cases with T13 (2/2/2 in \textsc{Train/Public/Hidden}) and 47 cases with an L6 (15/15/17). The effect of this increase in transitional vertebrae can be seen in Fig.~\ref{fig:teamwise_results_region_and_vert}, with L6 now being detected and segmented, at least in some cases. Surprisingly, T13, if occurring only twice is successfully identified by some methods. Note that Xiangshang Z. is the only approach that successfully identifies all T13 instances in both test phases. 

This contradictory behaviour of better performance of approaches in the case of T13 compared to L6, in spite of higher numbers gives us some insights into the task at hand. For T13, the sequence of vertebral labels gives a strong prior. In the case of L6, which itself acts as a strong prior due to the sacrum, its reliable detection doesn't seem as consistent. \cite{hanaoka2017automatic}, for example, recognise this issue and work towards directly predicting such abnormal numbers. Nonetheless, the improved behaviour of the approaches in such anatomical variations brings us closer to realising automated algorithms in clinical settings. 

\subsection{Limitations of our study}
The scale, clinical similitude, data and anatomical variability are the strengths of the \textsc{VerSe} benchmark. In this section, we identify some limitations of this study. 

Foremost among the limitations is the lack of inter-rater annotations. Owing to the effort involved in creating the voxel-level annotations for a multitude of vertebrae, the hierarchical process of okaying an annotation, and the use of a machine in the annotation process, the decision of having multiple-raters was delegated to future challenge iterations. This would eventually enable algorithms to predict uncertainty, inter-rater variability studies, and learning annotator biases.

Putting aside the insufficiency of the Dice metric for evaluating segmentation performance \citep{taha2015metrics}, the metrics in the spine literature have a major short-coming: one-label shift, where the labels of the predicted mask are \emph{off} by one label (cf. Fig.~\ref{fig:qual_verse20}, Worst Case). One-label shift penalises the current metrics more than label mixing, which results in unusable masks. The drastic drop in performance of Chen M. between the \textsc{Public} and \textsc{Hidden} phases (Table~\ref{tab:verse19}) was due to this issue. Therefore, research towards better domain-specific evaluation metrics is of interest, more so for differentiable variants enabling neural network optimisation.

%%%%%%%%%%%%%%%%%%%%%%%%%%%%%%%%%%%%%%

\section{Conclusions}
The Large Scale Vertebrae Segmentation Challenge (\textsc{VerSe}) was organised in two iterations in conjunction with MICCAI 2019 and 2020. \textsc{VerSe}, publicly made available 374 CT scans from 355 patients, the largest spine dataset to date with accurate centroid and voxel-level annotations. On this data, twenty-five algorithms (twenty-four participating algorithms, one baseline) are evaluated for the tasks of vertebral labelling and segmentation. This work describes the challenge setup, summarises the baseline and the participating algorithms, and benchmarks them with each other.The best algorithm in terms of mean performance in \textsc{VerSe'19} achieves identification rate of 94.25\% and a Dice score of 89.80\% (Payer C.) on the \textsc{Hidden} test set. In \textsc{VerSe'20}, these numbers are 96.6\% ($id.rate$) and 91.72\% (Dice), achieved by Chen D. Based on the statistical ranking method chosen for evaluating \textsc{VerSe} challenges, Payer C.'s approach led the leaderboard due to its better and relatively consistent performance on healthy as well as the anatomically rare cases. 

Aimed at understanding the algorithms' behaviour, we present an in-depth analysis in terms of the spine region, fields of view, and manual effort. We make the following key observations: (1) The performance of algorithms, on average, increased from \textsc{VerSe}`19 to \textsc{VerSe}`20, in spite of the data being more multi-centred and anomalous, (2) Spine processing, for now, is better approached in 3D, either as large patches or in a appropriately designed sequence of stages, and (3) Transitional vertebrae (T13 and L6) can be efficiently handled given sufficient data and post-processing. We hope that the \textsc{VerSe} dataset and benchmark will enable researchers to contribute towards more accurate and reliable clinical translation of their spine algorithms.

As stated, future directions could include the incorporation of multi-raters, inter-rater variability, and spine-centred evaluation measures. Additionally, modelling the sacrum is of interest for load analysis. Lastly, in spite of labelling and segmentation being inter-dependent, our motivation for having two tasks was to enable participation in individual tasks. However, our experience shows this to be redundant. Moreover, the \textsc{VerSe} challenges did not explicitly require the participating algorithms to be optimised for run time. Including this as an objective could bring in added insights into algorithm design. We bring these observations to the attention of future attempts at benchmarking.

%%%%%%%%%%%%%%%%%%%%%%%%%%%%%%%%%%%%%%

\section{Acknowledgements}
This work is supported by the European Research Council (ERC) under the European Union's `Horizon 2020' research \& innovation programme (GA637164--iBack--ERC--2014--STG). We acknowledge NVIDIA Corporation's support with the donation of the GPUs used for this research.

FA from  Zuse Intitute Berlin is funded by  the Deutsche Forschungsgemeinschaft (DFG, German Research Foundation)
under Germany’s Excellence Strategy -- The Berlin Mathematics Research Center MATH+ (EXC-2046/1, project ID: 390685689); TA from Zuse Intitute Berlin is funded by the German Ministry of Research and Education (BMBF) Project Grant 3FO18501 (Forschungscampus MODAL). 

%%%%%%%%%%%%%%%%%%%%%%%%%%%%%%%%%%%%%%

%\bibliographystyle{elsarticle-num}
\bibliographystyle{model5-names}
\bibliography{bibliography}

%%%%%% Appendix

\clearpage

\appendix

% %%%%%%%%%%%%%%%%%%%%%%%%%%%%%%%%%%%%%%%%%%%%%%%%%%%%%%%%%%%%%%%%%%%%%%%%%%%%%%%%%%%

\section{Challenge Evaluation and Ranking}
\label{app:ranking}
\subsection{Statistical Tests \& Points}

In this technical report (Sec. \ref{sec:experiments}), we reported four performance measures, $id.rate$, $d_\text{mean}$, Dice, and $HD$. However, recall from \ref{sec:materials}, that $HD$ and $d_\text{mean}$ are undefined in the case of missing vertebral predictions. Therefore, to rank the teams in \textsc{VerSe}`19, the missing predictions for vertebrae were substituted by a maximum Euclidean distance of \siunit{1000}{mm} for $d_\text{mean}$ and \siunit{100}{mm} for $HD$. Expecting more missed predictions in \textsc{VerSe}'20, and in order to avoid inducing a bias due to such substitution, $HD$ and $d_\text{mean}$ were not used to rank the teams in \textsc{VerSe}`20.

Once computing the performance measures, we compare them. Inspired by \cite{maier_hein2018} and \cite{menze2015}, the comparison and ranking of the participating algorithms were based on a scheme based on statistical significance. The value of the performance measure obtained from each scan in the cohort was treated as a sample from a distribution and the Wilcoxon signed-rank test with a `greater' or `less' hypotheses testing (as appropriate for the performance metric)  was employed to test the significance of the difference in performance between a pair of participants. A $p -$value of 0.001 was chosen as the threshold to ascertain a significant difference. Following this, a \emph{point} was assigned to the better team. All possible pairwise comparisons were performed for every performance measure. Each comparison awards a point to a certain team unless the difference is not statistically significant. For every measure, the points are aggregated at a team level and normalised with the total number of participating teams in the experiment to obtain a score between 0 and 1.

Lastly, for every team, the normalised points across the measures are combined as described in the next section, which describes particulars of point-computation for the ranking pertaining to the challenge.

\begin{figure}
  \centering
  %\vspace{-15pt}
  \includegraphics[width=0.85\textwidth]{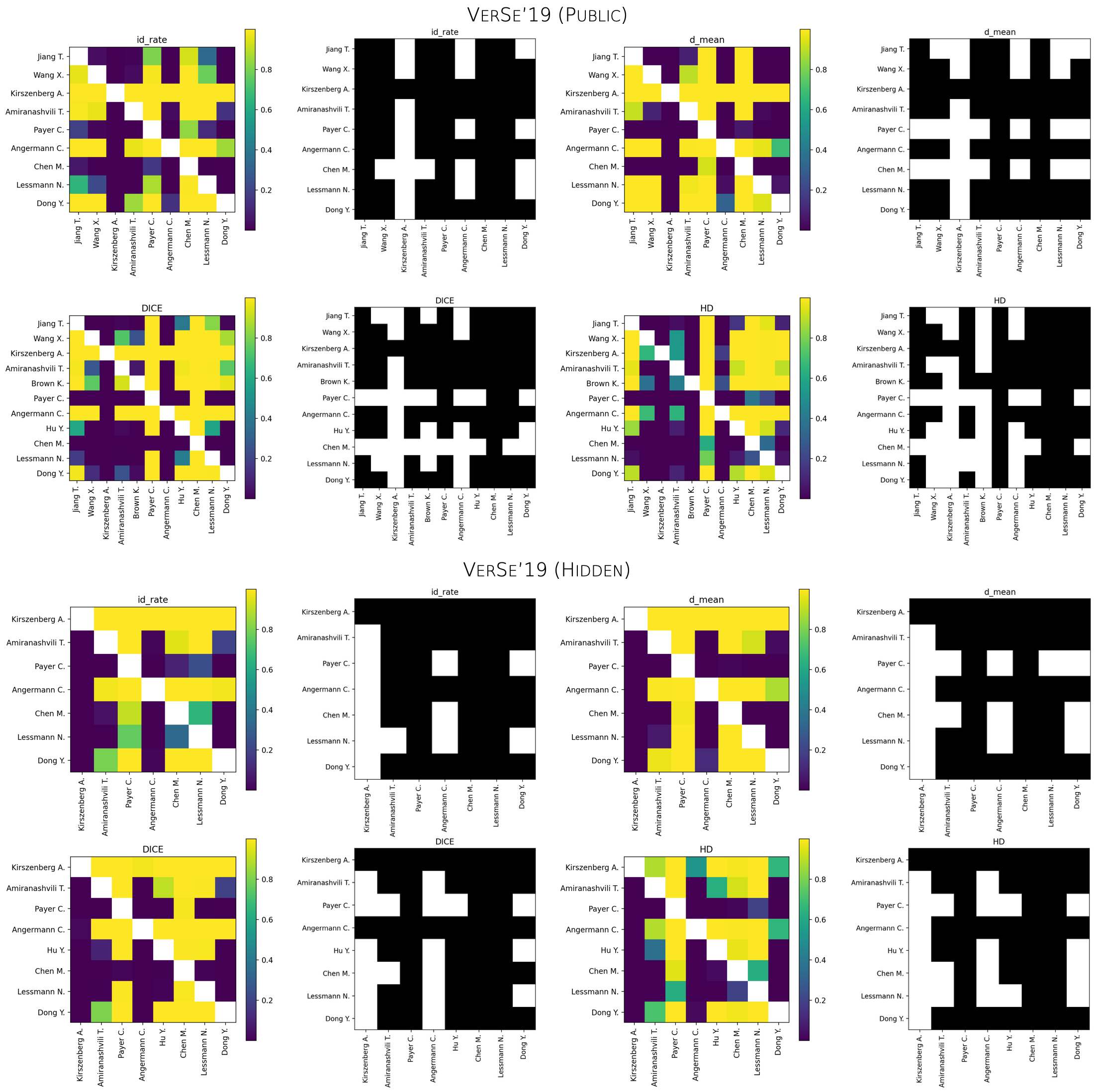}
  \caption{\textsc{VerSe}`19 points: Illustrating the $p-$value matrices and their binarised versions for every metric used.}
\label{fig:points_verse19}
 %\vspace{-50pt} % This removes the white box on the second page
\end{figure}

\begin{figure}
  \centering
  %\vspace{-15pt}
  \includegraphics[width=0.85\textwidth]{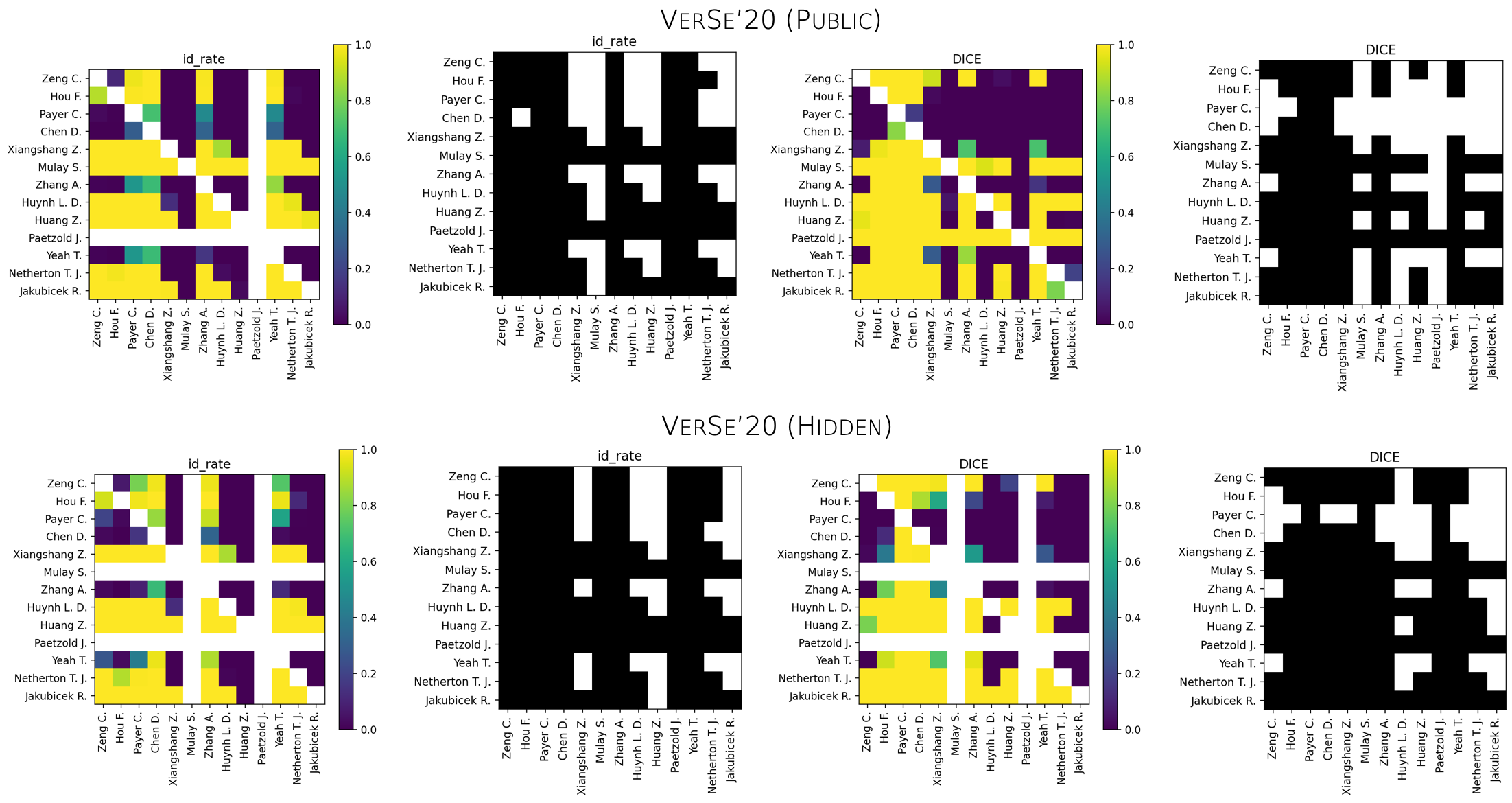}
  \caption{\textsc{VerSe}`20 points: Illustrating the $p-$value matrices and their binarised versions for every metric used.}
\label{fig:points_verse20}
 %\vspace{-50pt} % This removes the white box on the second page
\end{figure}

The points scored by each team are reported in Tables \ref{tab:points_verse19} and \ref{tab:points_verse20} respectively. Illustrated in Figs. \ref{fig:points_verse19} and \ref{fig:points_verse20} are the $p$--values of the significance as well as their binarised versions (thresholded at $p=0.001$) that ensue from the pairwise comparisons.

\subsection{Final Ranking: Combining all the scores}
\begin{figure}
  \centering
  %\vspace{-15pt}
   \begin{subfigure}[b]{0.48\textwidth}
        \includegraphics[width=0.95\textwidth]{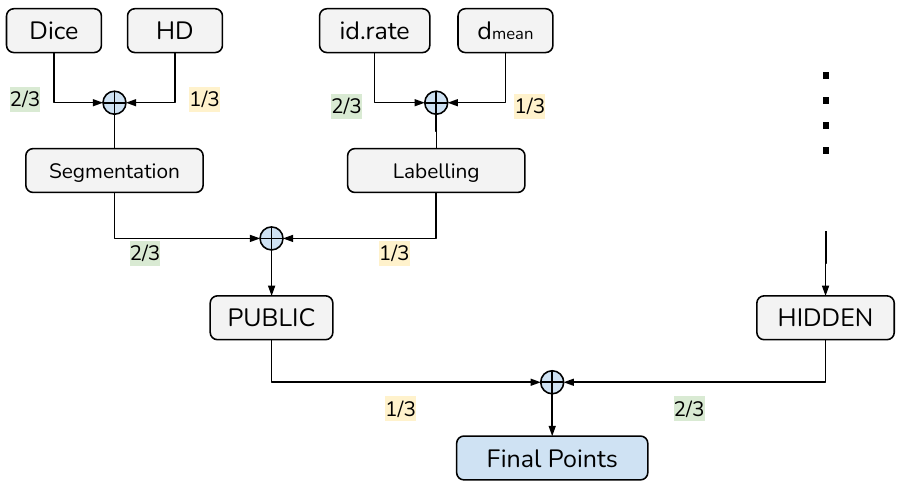}
        \caption{\textsc{VerSe'19}}
    \end{subfigure}
    ~
    \begin{subfigure}[b]{0.48\textwidth}
        \includegraphics[width=0.95\textwidth]{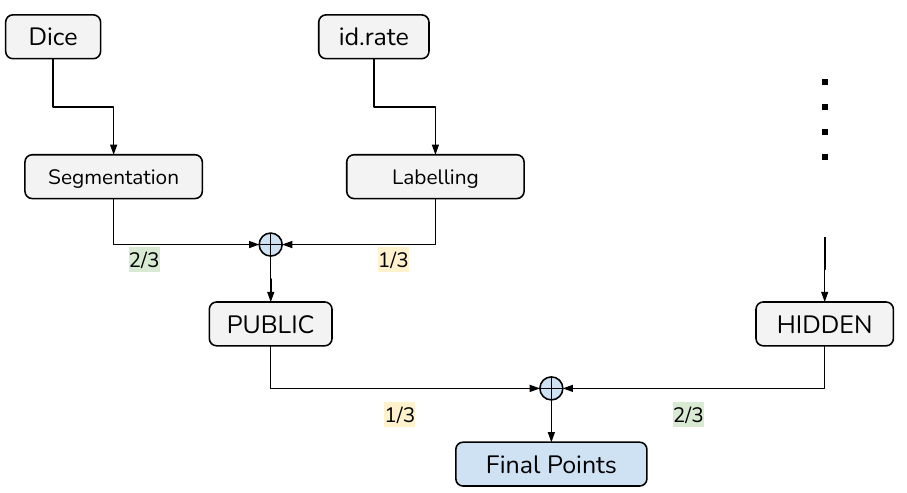}
        \caption{\textsc{VerSe'20}}
    \end{subfigure}
\caption{\textbf{Protocol for obtaining the final ranking}: Flow diagram of the weights assigned to each stage of the \textsc{VerSe} evaluation, in order to obtain the final point count.} %\vspace{-50pt} % This removes the white box on the second page
\label{fig:final_point_flow}
\end{figure}

Fig.~\ref{fig:final_point_flow} illustrates how the performance of the algorithms over the multiple stages was combined to construct one ranking scheme. Tables \ref{tab:points_verse19} and \ref{tab:points_verse20} also report the normalised points. The rationale in choosing this presented scheme was as follows:
\begin{itemize}
\item
$d_\text{mean}$ and $HD$, compared to $id.rate$ and Dice, are weighted at a ratio of $1:2$ in order to de-emphasise the contribution of the upper bounds chosen on the former measures in case of missing predictions. (This does not apply to \textsc{VerSe}`20.)
\item
\textsc{Hidden} has twice the weight as \textsc{Public} as it was evaluated on a completely hidden dataset, thus nullifying the chance of over-fitting or retraining on the test set.
\item
Lastly, the segmentation task has twice the weight of the labelling task as the latter can possibly be a consequence of the former, as was the final goal of this challenge.
\end{itemize}

\begin{table*}[!t]
\setlength{\tabcolsep}{0.2em}
\scriptsize
\renewcommand{\arraystretch}{1}
\begin{subtable}{\linewidth}\centering
 {
        \begin{tabular}{ c | c | c : c | c : c | c : c | c : c}
        \specialrule{.1em}{0em}{-.1em}
        \multirow{3}{*}{\rule{0pt}{2.5ex}\textbf{Team}} & \multirow{3}{*}{\rule{0pt}{2.5ex}\textbf{Normalised Points}}  &  \multicolumn{4}{c}{Labelling} & \multicolumn{4}{c}{Segmentation} \\[-1ex] 
        
        & & \multicolumn{2}{c}{\textsc{Public}} & \multicolumn{2}{c}{\textsc{Hidden}} & \multicolumn{2}{c}{\textsc{Public}} & \multicolumn{2}{c}{\textsc{Hidden}}\\
        & & $id.rate$ & $d_\text{mean}$ &   $id.rate$ & $d_\text{mean}$ & Dice &$HD$ & Dice &$HD$ \\ [0.25ex]
        \specialrule{.05em}{-0.1em}{0em}
        
        Payer C.& 0.691 & 3 & 7 & 3 & 5  & 8 & 8 & 5 & 5 \\
        
        Chen M.& 0.597 & 5 & 7 & 2 & 4 & 10 & 8 & 3 & 4 \\
        
        Lessmann N.& 0.496 & 3 & 1 & 4 & 3 & 4 & 5 & 3 & 5 \\
        
        Hu Y. & 0.279 & $\star$ & $\star$ & $\star$ & $\star$ & 4 & 4 & 3 & 3\\
        
        Dong Y. & 0.216 & 1 & 1  & 1 & 1 & 2 & 4 & 2 & 1\\
        
        Amiranashvili T. & 0.215 & 1 & 1 & 1 & 1 & 1 & 3 & 2 & 2\\
        
        Jiang T. & 0.140 & 3 & 5 & * & * & 4 & 4 & * & *\\        
        
        Angermann C. & 0.107 & 1 & 1 & 1 & 1 & 1 & 2 & 0 & 1 \\
        
        Wang X. & 0.084 & 2 & 3 & * & * & 2 & 3 & * & *\\
        
        Brown K. & 0.022 & $\star$ & $\star$ &  * & * & 1 & 1 & * & *\\
        
        Kirszenberg A. & 0.007 & 0 & 0 & 0 & 0 & 0 & 1 & 0 & 0\\
        
        \specialrule{.1em}{0em}{0em}
        \end{tabular}
    }
    \caption{\textsc{VerSe}`19}
    \label{tab:points_verse19}
    \end{subtable}
    
    \medskip
        
 \begin{subtable}{\linewidth}\centering
 {
        \begin{tabular}{ c | c | c | c | c | c }
        \specialrule{.1em}{0em}{-.1em}
        \multirow{3}{*}{\rule{0pt}{2.5ex}\textbf{Team}} & \multirow{3}{*}{\rule{0pt}{2.5ex}\textbf{Normalised Points}} & \multicolumn{2}{c}{Labelling} & \multicolumn{2}{c}{Segmentation} \\[-1ex] 
        
        & & \textsc{Public} & \textsc{Hidden} & \textsc{Public} & \textsc{Hidden}\\
        & &   $id.rate$ &  $id.rate$ & Dice & Dice \\ [0.25ex]
        \specialrule{.05em}{-0.1em}{0em}
        
        Payer C. & 0.675 & 6 & 4  & 11 & 10  \\
        
        Chen D. & 0.581 & 7 & 5 &  10 & 7\\
        
        Yeah T. & 0.453 & 6 & 5 &  7 &  5\\
        
        Zhang A. & 0.453 & 6 & 5 &   7 & 5\\
        
        Hou F. & 0.393 & 5 & 4 &  7 &  4\\
        
        Zeng C. & 0.333 & 6 & 4 &  5 & 3\\
        
        Xiangshang Z. & 0.316 & 2 & 2 &  6 & 4\\
        
        Netherton T. & 0.222 & 3 & 3&  3 & 2\\
        
        Huang Z. & 0.171 & 1 & 0 &  4 & 2\\
        
        Huynh L. & 0.119  & 3 & 2 & 1 & 2 \\
        
        Jakubicek R.$^\dagger$ & 0.085  & 1 & 1 & 3 & 0   \\
        
        Mulay S. & 0.017 &  0  & *  &   1  & *\\
        
        Paetzold J. & 0.0 & $\star$ & $\star$ & 0 & 0\\
        
        \specialrule{.1em}{0em}{0em}
        \end{tabular}
    }
    \caption{\textsc{VerSe}`20}
    \label{tab:points_verse20}
    \end{subtable}
        
\caption{Point counts of the submitted approaches of (a) \textsc{VerSe}`19 and (b) \textsc{VerSe}`20, based on the proposed pairwise, statistical comparison. * indicates a non-functioning docker container. $\dagger$ Jakubicek R. submitted a semi-automated method for \textsc{Public} and a fully-automated docker for \textsc{Hidden}.}
\label{tab:verse_overall}
\end{table*}

\section {Description of \emph{Anduin}}
\label{app:anduin_details} 

\begin{figure}[t!]
    \centering
         \includegraphics[width=0.85\textwidth]{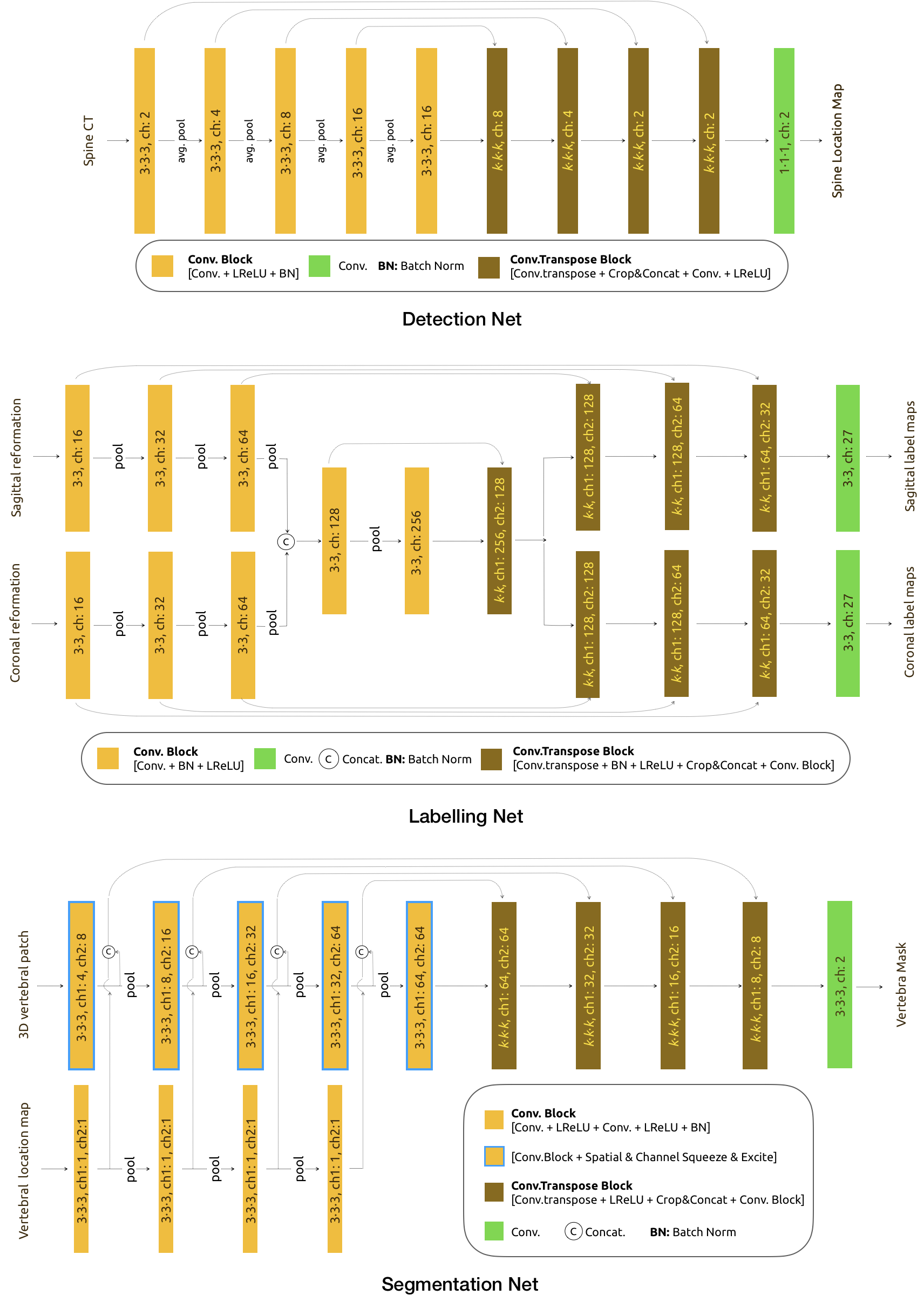}
    \caption{\textbf{Architectures}: Detailed network architectures of the three stages in \emph{Anduin}: the spine detection, vertebrae labelling, and the vertebra segmentation stages.}
\label{fig:architectures}
\end{figure}

The \emph{Anduin} framework was used to assist the data team in the creation of the ground truth. Please refer to \cite{loeffler2020_verse} for an overview of annotation-creation for \textsc{VerSe}. Given the CT scan of a spine, our framework aims to predict accurate voxel-level segmentation of the vertebrae by splitting the task in to three sub-tasks: spine detection, vertebrae labelling, and vertebrae segmentation. In the following section, the network architectures, loss functions, and training and inference details of each of these modules is elaborated. Fig. \ref{fig:pipeline} gives an overview of the proposed framework and Fig. \ref{fig:architectures} details the architectures of the networks employed in the three sub-tasks.\\    

\subsection{Notation.} The input CT scan is denoted by $\mathbf{x} \in \mathbb{R}^{h\times w \times d}$ where $h$, $w$, and $d$ are the  height, width, and depth of the scan respectively. The annotations available are, (1) the vertebral centroids, denoted by $\{\mu_i \in \mathbb{R}^3\}$ for $i \in \{1, 2, \dots N\}$. These are used to construct the ground truth for the detection and labelling tasks, denoted by $\mathbf{y}_{d}$ and $\mathbf{y}_{l}$, respectively.  (2) the multi-label segmentation masks, denoted by $\mathbf{y}_{s} \in \mathbb{Z}^{h\times w \times d}$.

\subsection{Spine Detection}
To detect the spine, we propose a parametrically-light, 3D, FCN operating at an isotropic resolution of \siunit{4}{mm}. This network regresses a 3D volume consisting of Gaussians at the vertebral locations as shown in Fig. \ref{fig:architectures}. The Gaussian heatmap is generated at a resolution of \siunit{1}{mm} with a standard deviation, $\sigma=8$, and then downsampled to a resolution of \siunit{4}{mm}. Additionally, spatial squeeze and channel excite blocks (SSCE) are employed to increase the network's performance-to-parameters ratio. Specifically, the probability of each voxel  being a \emph{spine voxel} or a \emph{non-spine} one is predicted by optimising a combination of $\ell_2$ and binary cross-entropy losses as shown:
\begin{equation}
\mathcal{L}_{\text{detect}} = ||\mathbf{y}_{d} - \tilde{\mathbf{y}}_{d}||_2 - H\left(\sigma(\mathbf{y}_{d}), \sigma( \tilde{\mathbf{y}}_{d})\right) 
\end{equation}   
where $\mathbf{y}_{d}$ is constructed by concatenating the Gaussian location map with a background channel obtained by subtracting the foreground from 1, $\tilde{\mathbf{y}}_{d}$ denotes the prediction of whose foreground channel represents the desired location map, and $\sigma(\cdot)$ and $H(\cdot)$ denote the softmax and cross-entropy functions.

\subsection{Stage 2: Vertebrae Labelling}
To label the vertebrae, we adapt and improve the Btrfly net \citep{sekuboyina2018,sekuboyina2020} that works on two-dimensional sagittal and coronal MIP. By virtue of the spine's extent obtained from the previous component, MIPs can now be extracted from a region focused on the spine, thus eliminating occlusions from ribs and pelvic bones. Cropping the scans to the spine region also makes the input to the labelling stage more uniform, thus improving the training stability. The labelling module works at \siunit{2}{mm} isotropic resolution and is trained by optimizing the loss function that is a combination of the sagittal and coronal components, $\mathcal{L}_{\text{label}} = \mathcal{L}_{\text{label}}^{\text{sag}} + \mathcal{L}_{\text{label}}^{\text{cor}}$, where the loss of each view is given by: 
\begin{equation}
\mathcal{L}_{\text{label}}^{\text{sag}} = ||\mathbf{y}_{l}^{\text{sag}} - \tilde{\mathbf{y}}_{l}^{\text{sag}}||_2+ \omega H\left(\sigma(\mathbf{y}_{l}^{\text{sag}}),\sigma(~\tilde{\mathbf{y}}_{l}^{\text{sag}})\right),
\end{equation}   
where $\tilde{\mathbf{y}}_{l}^{\text{sag}}$ is the prediction of the net's sagittal-arm of the Btrfly net and $\omega$ denotes the median frequency weight map giving a higher weight to the loss originating from less frequent vertebral classes.

\subsection{Stage 3: Vertebral Segmentation}
Once the vertebrae are labelled, their segmentation is posed as a binary segmentation problem. This is done by extracting a patch around each vertebral centroid predicted in the earlier stage and segmenting the vertebra of interest. An architecture based on the U-Net working at a resolution of \siunit{1}{mm} is employed for this task. Additionally, SSCE blocks are incorporated after every convolution and upconvolution block. Importantly, as there will be more than one vertebra within a patch, a vertebra-of-interest (VOI) arm is used to point the segmentation network to delineate the vertebra of interest. The VOI arm is an encoder parallel to the image encoder as shown in Fig. \ref{fig:architectures}, processing a 3D Gaussian heatmap centred at the vertebral location predicted by the labelling stage. The feature maps of the VOI arm are concatenated to those of the image encoder at every resolution. The segmentation network is trained using a standard binary cross-entropy as a loss.

\begin{algorithm}[h!]
\small
\SetAlgoLined
\KwIn{$\mathbf{x}$, a 3D MDCT spine scan}
\KwOut{Vertebral centroids \& segmentation masks\\ \textsc{~Detection}}
\texttt{$\mathbf{x}_d$ = resample\_to\_4mm}($\mathbf{x}$)\\
$\mathbf{y}_d$ = \texttt{predict\_spine\_heatmap}($\mathbf{x}_d$)\\
$bb$ = \texttt{construct\_bounding\_box}($\mathbf{y}_d$, \texttt{threshold}=$T_d$)\\
\underline{Possible interaction}: Alter $bb$ by \emph{mouse-drag} action.\linebreak
\textsc{Labelling}\\
\texttt{$\mathbf{x}_l$ = resample\_to\_2mm}($\mathbf{x}$)\\
$bb$ = \texttt{upsample\_bounding\_box}($bb$, \texttt{from}=4mm, \texttt{to}=2mm)\\
$\mathbf{x}_{sag}$, $\mathbf{x}_{cor}$ = \texttt{get\_localised\_mips}($\mathbf{x}_l$, $bb$)\\
$\mathbf{y}_{sag}$, $\mathbf{y}_{cor}$ = \texttt{predict\_vertebral\_heatmaps}($\mathbf{x}_{sag}$, $\mathbf{x}_{cor}$)\\
$\mathbf{y}_l$ = \texttt{get\_outer\_product}($\mathbf{y}_{sag}$, $\mathbf{y}_{cor}$)\\
\texttt{\textbf{centroids}} = \texttt{heatmap\_to\_3D\_coordinates}($\mathbf{y}_l$, \texttt{threshold}=$T_l$)\\
\underline{Interaction}: Insert missing vertebrae, delete spurious predictions, drag incorrect predictions.\linebreak
\textsc{Segmentation}\\
$\mathbf{x}_s$ =  \texttt{resample\_to\_1mm}($\mathbf{x}$); $mask$ = \texttt{np.zeros\_like}($\mathbf{x}_s$)\\
\For{every \texttt{centroid} in \texttt{centroids}}{
\texttt{$p$} = \texttt{get\_3D\_vertebral\_patch}($\mathbf{x}_s$, \texttt{centroid})\\
\texttt{$p_{mask}$} = \texttt{binary\_segment\_vertebra\_of\_interest}($p$)\\
\texttt{$p_{mask}$} = \texttt{index\_of}($mask$, \texttt{centroid})$*p_{mask}$\\
$mask$ = \texttt{put\_vertebrae\_in\_mask}($p_{mask}$)\\
}
\caption{Pseudocode for inference on \emph{Anduin}}
\label{algo:inference}
\end{algorithm}

\subsection{Inference \& Interaction}
Simplifying the flow of control throughout the pipeline, Algo. \ref{algo:inference} describes the inference routine given a spine CT scan and various points where medical experts can interact with the results, thus improving its overall performance.

\section {Participating Algorithms}
\label{app:teams}

% ================== VERSE 2019  ================== 
% -- Amiranashvili T. ----------------------------

\subsection*{\crule[ff0000]{0.25cm}{0.25cm} Amiranashvili T. et al.: Combining Template Matching with CNNs for Vertebra Segmentation and Identification}
\label{desc:amiranashvilit}

\begin{figure}%{\textwidth}
  \centering
  \includegraphics[width=0.6\textwidth]{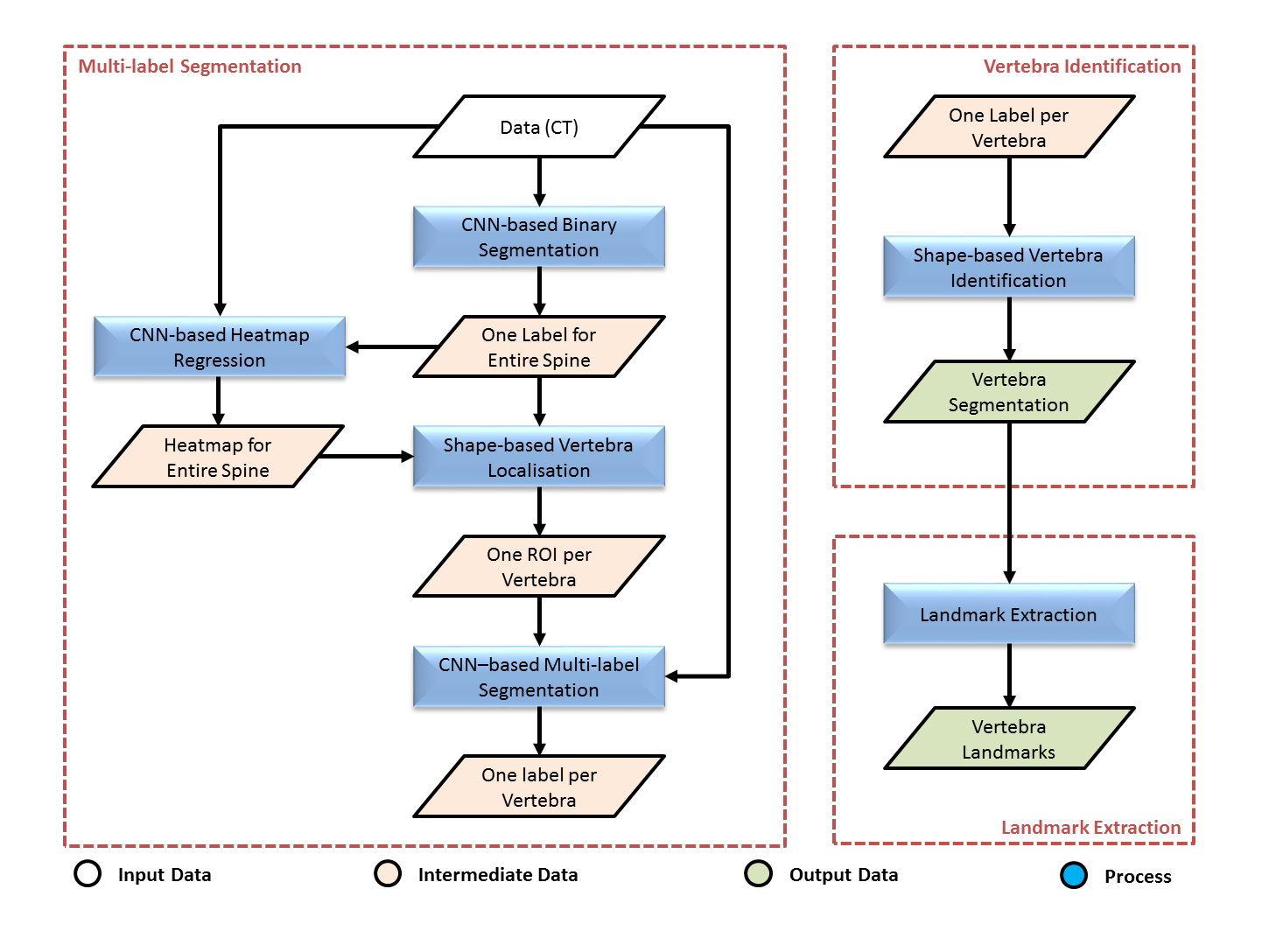}
  \caption{Multiple stages involved in the algorithm proposed by \emph{Amiranashvili T.}}
\label{fig:zib}
 %\vspace{-110pt} % This removes the white box on the second page
\end{figure}

A multi-stage approach is adopted to label and segment the vertebrae as illustrated in Fig. \ref{fig:zib}: 1. Multi-label segmentation with arbitrary, but separate labels for each vertebra based on local regions of interest in the image. 2. Unique label-assignment to segmented vertebral masks based on shape, while globally regularising
over the entire CT field of view. 3. Derive landmark positions from the multi-label segmentation masks by applying a shape-based approach.\\

\noindent
\emph{Multi-label Segmentation.} This stage includes creating a first, rough binary segmentation of the overall spine followed by localising regions of interests around each vertebra and performing voxel-level, refined segmentation of each vertebra. Binary segmentation separating the
spine from the background is achieved through a U-Net employed on 2D sagittal slices. For each slice, neighboring slices are included as
additional channels in the input to provide a larger context. The network is trained on fixed-size, random crops from original slices. Following this, the number of vertebra and their rough positions are computed based on the binary segmentation by combining shape-based fitting via generalised Hough transform (GHT) \citep{seim08} with a CNN-based heat-map regression for localising vertebra in the spinal column. Put to use in the fitting procedure were manually generated GHT templates of the lumbar (L1-L5), lower thoracic (T10-T12), mid-thoracic (T5-T9), upper-thoracic (T1-T4), lower-to-mid cervical (C3-C5), and upper-cervical (C2-C1) spine. The Butterfly network \citep{li2018butterfly} was trained on mean and maximum intensity projections in anterior-posterior and lateral directions of the CTs. Finally, multi-label segmentation is performed based on the rough locations from the previous step by deriving a region of interest for each visible vertebra. Individual vertebrae are then segmented via a U-Net based on 2D sagittal slices cropped to the corresponding regions of interests while including neighboring slices as additional input channels. The segmentation masks resulting from the cropped images are then combined into a multi-label segmentation mask.\\

% \begin{wrapfigure}{r}{0.6\textwidth}
%   \centering
%   \includegraphics[width=0.58\textwidth]{figures/appendix/zib.png}
%   \caption{Multiple stages involved in the algorithm proposed by \emph{Amiranashvili T.}.}
% \label{fig:zib}
%  %\vspace{-110pt} % This removes the white box on the second page
% \end{wrapfigure}

\noindent
\emph{Vertebra Identification.} Vertebra identification is performed based on shape through template fitting along with explicit global regularisation over the whole visible spine. For each vertebra, shape templates are fitted non-rigidly to the given labels via the iterative closest points (ICP) algorithm using the six templates introduced above. This results in a table containing a fitting score for each template and each detected vertebra. Then, optimisation for a set of unique vertebra types is performed such that the combined score from the table
is maximised while maintaining the consistent ordering of vertebra (e.g. L4 must follow L5). The multi-label segmentation of the previous stage is then re-labelled according to the determined ordering, resulting in a
segmentation with uniquely identified labels for each vertebra.\\

\noindent
\emph{Landmark Extraction.} After segmentation and identification, the positions of the landmarks are identified by re-fitting a template of the body of each vertebra to the unique labels followed by extracting the template's centre point which forms the landmark.

% -- angermann ----------------------------

\subsection*{\crule[b00000]{0.25cm}{0.25cm} Angermann C. et al.: A Projection-based 2.5D U-net Architecture for \textsc{VerSe}`19. \citep{angermann19}}
\label{desc:angermannc}

For the task of a fully-automated technique for volumetric spine segmentation, a combination of a 2D slice-based approach and a projections-based approach is proposed with two tasks: 1. 3D spine segmentation with one output channel denoting the probability of a voxel belonging to a vertebra, followed by assignment of a label from C1 to L6. 2. Using the multi-label segmentation mask, weighted centroid computation for each label for the task of vertebra labelling. Please refer to \citep{angermann19} for details on the 3D segmentation procedure.

% \begin{wrapfigure}{r}{0.4\textwidth}
%   \centering
%   \includegraphics[width=0.38\textwidth]{figures/appendix/angermann.png}
%   \caption{Maximum intensity projections of a 3D spine scan with directions $\{k\times30 \text{~degrees} | k = 0, . . . , 5\}$}. 
%   \label{fig:angermann}
%  %\vspace{-110pt} % This removes the white box on the second page
% \end{wrapfigure}

\begin{figure}%{\textwidth}
  \centering
  \includegraphics[width=0.5\textwidth]{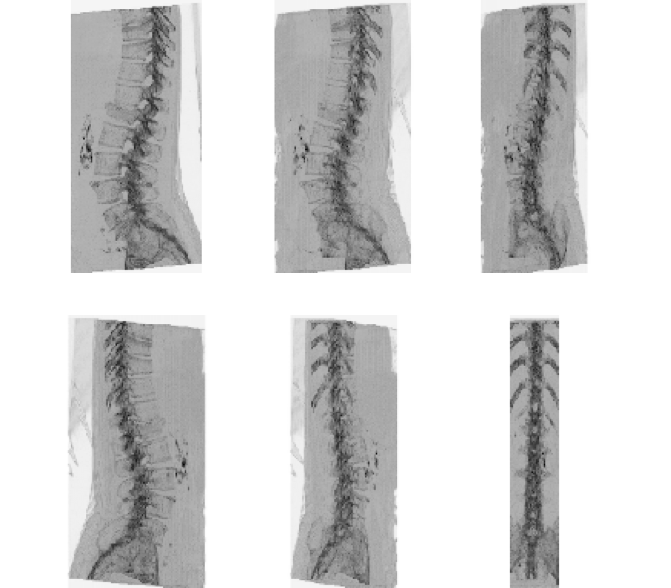}
  \caption{Maximum intensity projections of a 3D spine scan with directions $\{k\times30 \text{~degrees} | k = 0, . . . , 5\}$}. 
  \label{fig:angermann}
 %\vspace{-110pt} % This removes the white box on the second page
\end{figure}

\noindent
\emph{Vertebra Segmentation.} This is a two-step approach working with images of size $224\times224\times224$, obtained by zooming the array such that the longest axis is size 224 and padding the other axes with zeros. In the first step, whose output is a one channel segmentation mask (vertebra as foreground), a 2.5D U-net \citep{angermann19} and two 2D U-net are employed. The former network takes the 3D array as input and generates 2D projections containing full 3D information. Here the MIPs are employed (cf. Fig \ref{fig:angermann}). These 2D projections are propagated through a 2D U-net and lifted back to a volume using a trainable reconstruction algorithm (cf. Eq 3.1,  \citep{angermann19}). Due to the non-convex nature of vertebrae, this segmentation is combined with that of a 2D slice-based U-net in the probability space. In the second step, the binary segmentation mask is assigned multiple labels. For this, A 2D U-Net working on six MIPs per scan is employed. Each of the MIPs is obtained at an angle in $\{0^o, 10^o, 80^o, 90^o, 100^o, 170^o\}$, as in Fig.~\ref{fig:angermann}. As output, six labelled MIP segmentation masks are obtained. From these, the 3D labelled mask is obtained by back-projection, wherein each 2D MIP mask is multiplied by a rotated 3D binary segmentation from the previous step, rotated according to the angle corresponding to the MIP mask in question.\\ 

\noindent
\emph{Vertebra Labelling.}
Since the vertebrae are already labelled in the segmentation stage, the vertebral centroids are obtained by just weighing the edges of the vertebra and computing the centroid. The edge-weight is set empirically and is same across the vertebrae.

% -- brown ----------------------------

\subsection*{\crule[870000]{0.25cm}{0.25cm} Brown K. et al.: Spine Segmentation with Registration}
\label{desc:brownk}

\begin{figure}%{\textwidth}
  \centering
%   \vspace{-75pt}
  \includegraphics[width=0.5\textwidth]{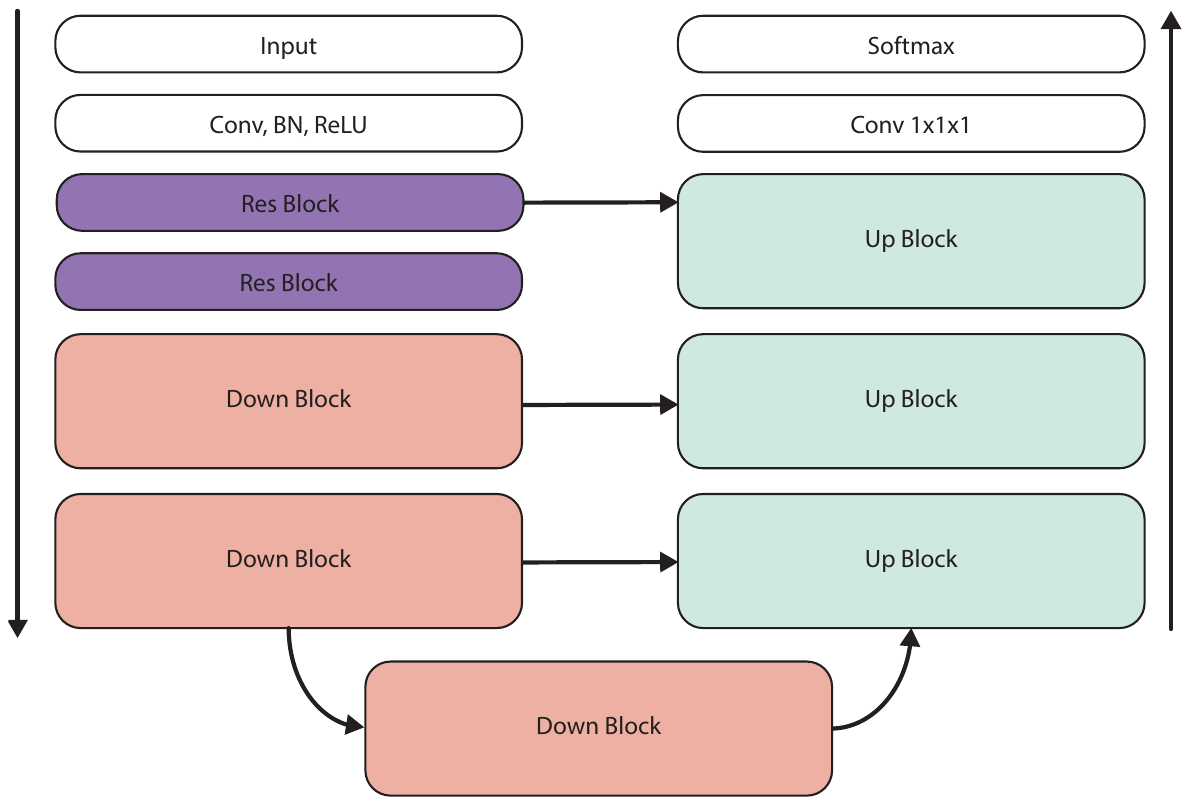}
  \caption{The residual U-Net employed for segmentation in \emph{Brown K.}'s approach.}
  \label{fig:brown}
 \vspace{-50pt} % This removes the white box on the second page
\end{figure}

Segmentation of the vertebrae is performed by extracting a bounding box around each vertebra and segmenting this box with a residual U-net. The bounding box around the vertebra is identified via a regressed set of canonical landmarks. Each vertebra is then registered to a common `atlas’ space via these landmarks. For segmentation, the employed residual U-net works with inputs of size $64\times64\times64$ voxels with a depth of five blocks (cf. Fig.~\ref{fig:brown}).\\

% \begin{wrapfigure}{r}{0.45\textwidth}
%   \centering
%   \vspace{-75pt}
%   \includegraphics[width=0.4\textwidth]{figures/appendix/brown.pdf}
%   \caption{The residual U-Net employed for segmentation in \emph{brown}'s approach.}
%   \label{fig:brown}
%  \vspace{-50pt} % This removes the white box on the second page
% \end{wrapfigure}

\noindent
\emph{Objective Function.} A network is trained to minimise a combination of Dice coefficient ($L_D$) and a weighted  false-positive/false-negative loss ($L_{FPFN}$), described as: $L = L_D + \alpha L_{FPFN}$ ($\alpha=0.5$ in this work). Specifically, the dice coefficient measures the
degree of overlap between two sets. For two binary sets ground truth (G) and predicted class membership (G) with (N) elements each, the dice coefficient can be written as   
\begin{equation}
D = \frac{2 \sum_i^N p_i g_i}{\sum_i^Np_i + \sum_i^N g_i},
\end{equation}
where each $p_i$ and $g_i$ are binary labels. In this case, $p_i$ is set to [0, 1] from the softmax layer representing the probability that the $i^{th}$ voxel is in the foreground class. Each $g_i$ is obtained from a one-hot encoding of the ground-truth-labelled volume of tissue class. Additionally, the weighted false-positive/false-negative loss term is included to provide smoother
convergence. It is defined as:
\begin{equation}
L_{FPFN} = \sum_{i\in I} w_i p_i (1 - g_i) + \sum_{i \in I} w_i (1 - p_i) g_i,
\end{equation}

where the weight, $w_i = \gamma_e exp (-d_i^2/\sigma) + \gamma_c f_i$, with $d_i$ being the euclidean distance to the nearest class boundary and $f_i$ the frequency of the ground truth class at voxel $i$. In this work, $\sigma$ is chosen to be 10 voxels, and the parameters $\gamma_e$ and $\gamma_c$ are set to 5 and 2, respectively.

% -- iflytek ----------------------------

\subsection*{\crule[550000]{0.25cm}{0.25cm} Chen M.: An Automatic Multi-stage System for Vertebra Segmentation and Labelling}
\label{desc:chenm}

A three-stage strategy is applied to solve the task of vertebral segmentation and labelling. The first two stages are based on a U-Net architecture for multi-label segmentation. Utilising the predicted segmentation mask, the third stage employs an RCNN-based architecture \citep{frcnn1,frcnn2} to label the vertebrae.

% \begin{wrapfigure}{r}{0.5\textwidth}
%   \centering
%   \includegraphics[width=0.48\textwidth]{figures/appendix/iflytek.pdf}
%   \caption{Procedure for label correction after Stage 2.}
% \label{fig:iflytek}
%  %\vspace{-110pt} % This removes the white box on the second page
% \end{wrapfigure}

\begin{figure}%{\textwidth}
  \centering
  \includegraphics[width=0.4\textwidth]{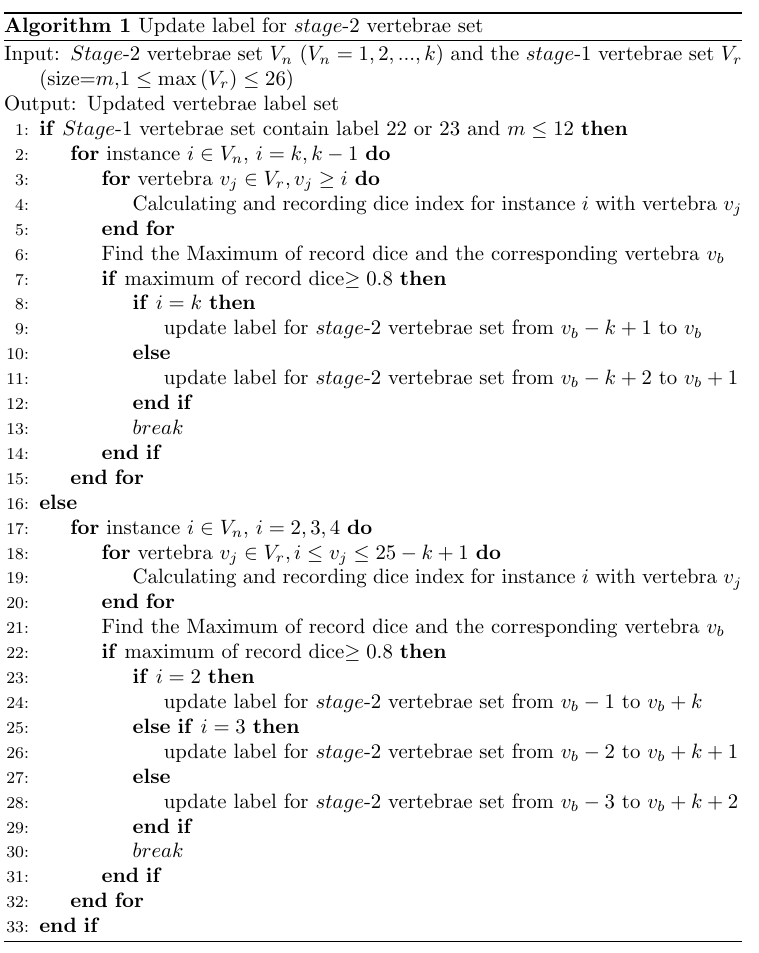}
  \caption{Procedure for label correction after Stage 2 of \emph{Chen M.}'s approach.}
  \label{fig:iflytek}
 %\vspace{-110pt} % This removes the white box on the second page
\end{figure}

\noindent
\emph{Segmentation (Stages 1 \& 2).} The first stage consists of a 3D
U-Net working on randomly extracted patches of size $224\times160\times128$. The network is trained to predict 25 labels, ignoring the rare L6 label. It is observed that the segmentation Stage 1 performs well in regions close to C1 and L5. However, in the other regions, the vertebral labels are mixed with each other due to a similarity in their shapes. To resolve this problem, a second \emph{refinement network} is introduced with an architecture similar to the first stage but with a major difference in the training regime. For this, patches are extracted covering the spine in the middle and extending 1.5 times in the \emph{slice} direction. These patches are padded to $128\times128\times128$ with zeroes if necessary. The network is trained to predict a binary label only for the mid-vertebra. The combination is trained as follows: All the labelled Stage 1 masks are combined into a binary mask, indicating the foreground. Each of these masks (corresponding to each vertebral label) is used to generate a patch for Stage 2. This prediction is believed to be accurate at instance level and filled back into the binary foreground. If the foreground is not filled sufficiently, new patches will be selected from the not-filled regions for Stage 2 recursively till convergence. Because the well-segmented instances in Stage 1 and Stage 2 mostly overlap, it is operable to assign labels based on both the stages by comparing the Dice of the pairs. With the constraint on the label continuity of neighboring spines, this process can be performed using the matching algorithm presented in Fig. \ref{fig:iflytek}.\\

\noindent
\emph{Labelling.} An RCNN-based architecture with a 3D ResNet-50 is used as the backbone for the vertebra labelling task. RoI pooling is performed on the features of the feature map at stride 4 to regress the deviation of the vertebra centre to the RoI box's centre in the coordinate space of the box. This network works with inputs of size $160\times192\times224$. In the training phase, boxes are generated from the segmentation ground truth such that more positive samples are generated. During inference, the predicted segmentation mask is utilised.

% -- Dong Y. ----------------------------

\subsection*{\crule[e4e400]{0.25cm}{0.25cm} Dong Y. et al.: Vertebra Labeling and Segmentation in 3D CT using Deep Neural Networks \citep{qihang19}}
\label{desc:dongy}

A U-shaped deep network is used for generating the vertebral segmentation masks and labels in the form of a model ensemble  followed by a post-processing module.

The problem is formulated as a 26-class segmentation task given 3D CT as input. The class information from prediction is able to provide labels (cervical $C1 \sim C7$, thoracic $T1 \sim T12$, lumbar $L1 \sim L6$) for different vertebrae. For vertebra localisation, the centroids of vertebrae are determined as the mass centres of segmentation masks.

We have adopted a U-shaped neural network for vertebral segmentation following the fashion of the state-of-the-art network for 3D medical image segmentation. The network architecture is nearly symmetric with an encoder and a decoder. After achieving the segmentation results, the vertebrae centroids are computed based on the mass centres of binary labels for each individual vertebra. To further help determine the vertebral body centre, several iterations of morphological erosion are conducted to remove the vertebral `wings'. The final prediction is from the ensemble of the five models.

% -- Hu Y. ----------------------------

\subsection*{\crule[baba00]{0.25cm}{0.25cm} Hu Y. et al.: Large Scale Vertebrae Segmentation Using nnU-Net}
\label{desc:huy}

The tasks at hand are posed as an application of the nnU-Net \citep{isensee2019}, a framework that automatically adapts the hyper-parameters to any given dataset. 

Generally, nnU-Net consists of three U-Net models (2D, 3D, and a cascaded 3D network) working on the images patch-wise. It automatically sets the training hyper-parameters such as the batch size, patch size, pooling operations etc. while keeping the GPU budget within a certain limit. If the selected patch size covers less than 25\% of the voxels in case, the 3D-Net cascade is additionally configured and trained on a downsampled version of the training data. Specific to \textsc{VerSe}`19, a sum of cross-entropy loss and Dice loss are used the training objective, minimised using the Adam optimiser. An initial rate of $3\times10^{-4}$ and $\ell_2$ weight decay of $3\times10^{-5}$ . The learning rate is dropped by a factor of 0.2 whenever the exponential moving average of the training loss does not improve within the last 30 epochs. Training is stopped when the learning rate drops below $10^{-6}$ or 1000 epochs are exceeded. The data is augmented using elastic deformations, random scaling, random rotations, and gamma augmentation. Note that in Phase 1, the nnU-Net ensemble did not include all its components. Included are a 3D U-Net operating at full resolution, a 3D U-Net at low resolution (as part of the cascade 3D), and a 2D U-Net. 

% -- Jiang T. ----------------------------

\subsection*{\crule[878700]{0.25cm}{0.25cm} Jiang T. et al.: SpineAnalyst: A Unified Method for Spine Identification and
Segmentation}
\label{desc:jiangt}

\begin{figure}%{\textwidth}
  \centering
  \includegraphics[width=0.75\textwidth]{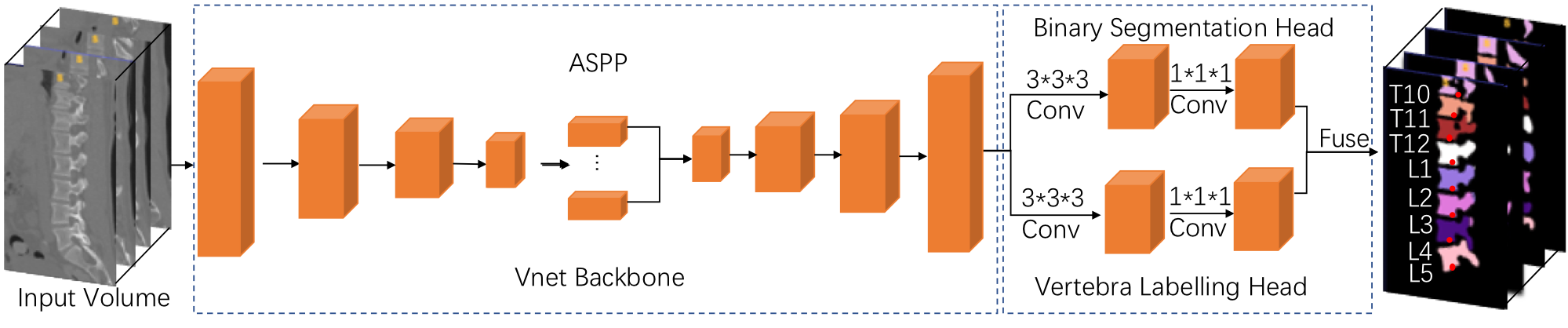}
  \caption{An overview of SpineAnalyst network, a contribution of \emph{Jiang T}.}
  \label{fig:alibabadamo}
 %\vspace{-110pt} % This removes the white box on the second page
\end{figure}

In contrast to most approaches that treat identification and segmentation as two separate steps, this work efficiently solves them simultaneously with a keypoint based instance segmentation framework applying anchor-free instance segmentation networks in a 3D setting. To the best of the participant's knowledge, this is a first. The proposed network adopts the encoder-decoder paradigm with two prediction heads attached to the shared decoder, as described in Fig. \ref{fig:alibabadamo}. The `binary segmentation head' distinguishes spine pixels resulting in a binary semantic map. The `vertebra labeling head' detects and labels all the vertebrae landmarks, while also predicting a vector field that associates vertebral pixels with their vertebrae centres. The predictions of two heads are fused together to produce
the final instance segmentation results

% \begin{wrapfigure}{r}{\textwidth}
%   \centering
%   \includegraphics[width=0.58\textwidth]{figures/appendix/alibabadamo.png}
%   \caption{An overview of SpineAnalyst network, a contribution of \emph{Jiang T}.}
%   \label{fig:alibabadamo}

%  %\vspace{-110pt} % This removes the white box on the second page
% \end{wrapfigure}

\noindent
\emph{Encoder \& Decoder.} A V-Net is used as the backbone with the encoder containing four cascaded blocks. Following this, the Atrous Spatial Pyramid Pooling (ASPP) method is applied to further increase the receptive field and capture multi-scale information effectively. In the decoder, the concatenated features of ASPP are passed through four cascaded up-sampling blocks recovering the original volume resolution .

\noindent
\emph{Binary Segmentation Head.}
A binary semantic segmentation head is trained to detect the spine as the foreground pixels. These pixels will further be assigned with vertebral labels in the subsequent fusion processing.

\noindent
\emph{Vertebra Labeling Head.}
This component performs two tasks: 1. Detect and label landmarks: For the former, the heatmap channels predict the probability that a pixel belongs to a vertebra centre. Pixels corresponding to high confidence are reserved as vertebral landmarks. Due to the similarity of the adjacent vertebrae, it is challenging to directly identify individual vertebrae. Instead, the reference vertebrae with obvious anatomical features, such as C2, L5 and C7, T12, are first identified. Other vertebrae labels are then inferred from the reference vertebrae. Following this, 2. a vector-field is predicted with each channel denoting the offsets relative to the corresponding vertebra centre. Each pixel is then labelled with the closest vertebra centre according to the long offset.

\noindent
\emph{Fusion Process.}
The final instance segmentation is obtained from binary semantic segmentation as follows: Each pixel within the semantic mask acquires its label from the centre point closest to its predicted centres, which is computed by pixel coordinates plus the vector field.

% -- Kirszenberg A. ----------------------------

\subsection*{\crule[545400]{0.25cm}{0.25cm} Kirszenberg A. et al.: }
A multi-stage approach is proposed involving a pseudo-3D U-Net architecture for segmentation and a template matching approach enabled by morphological operation.
\label{desc:kirszenberga}

\begin{figure}[H]
    \centering
  
  \begin{subfigure}[b]{0.4\textwidth}
        \includegraphics[width=0.8\textwidth]{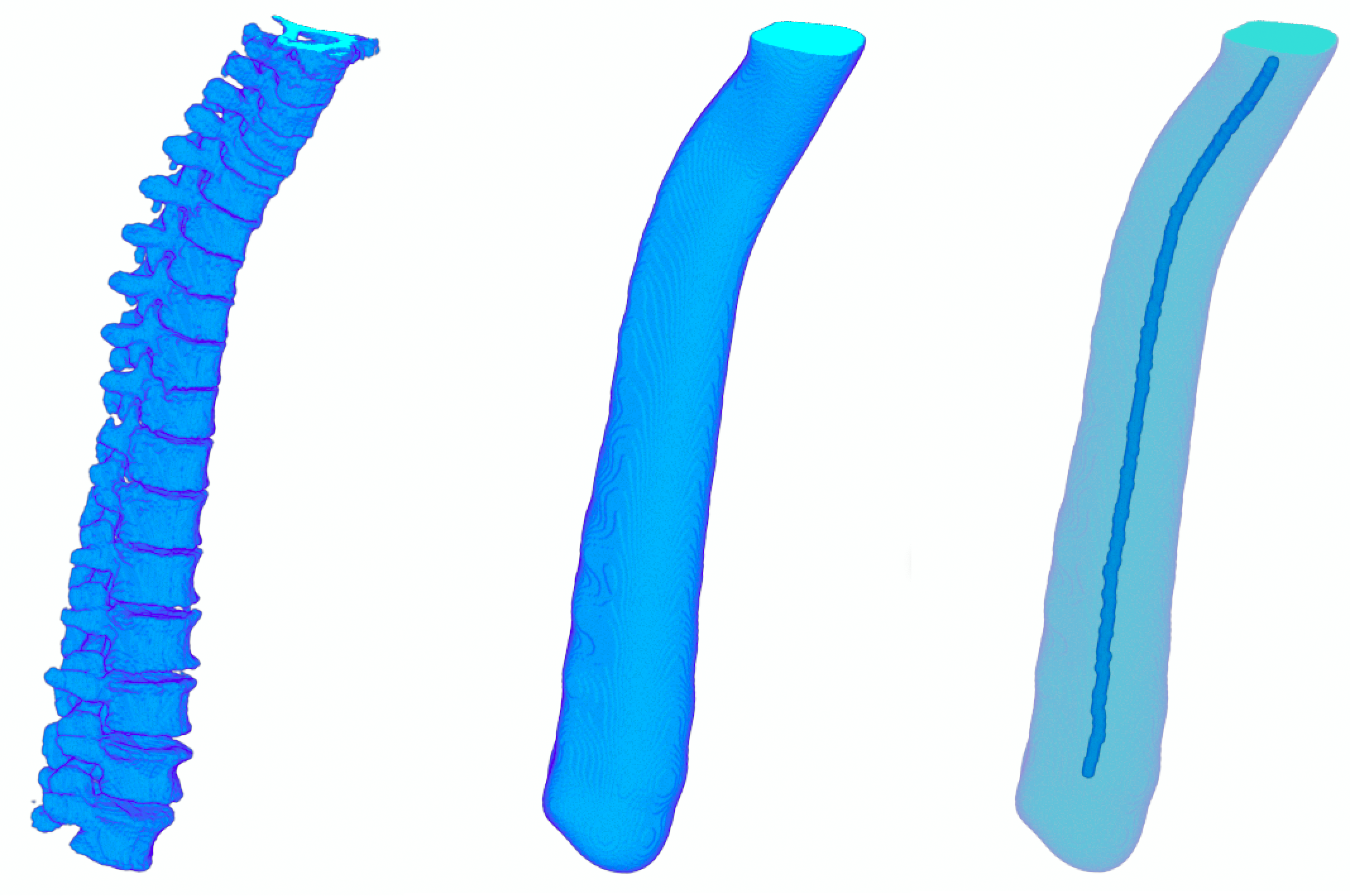}
        \caption{}
    \end{subfigure}
  ~
     \begin{subfigure}[b]{0.4\textwidth}
        \includegraphics[width=0.8\textwidth]{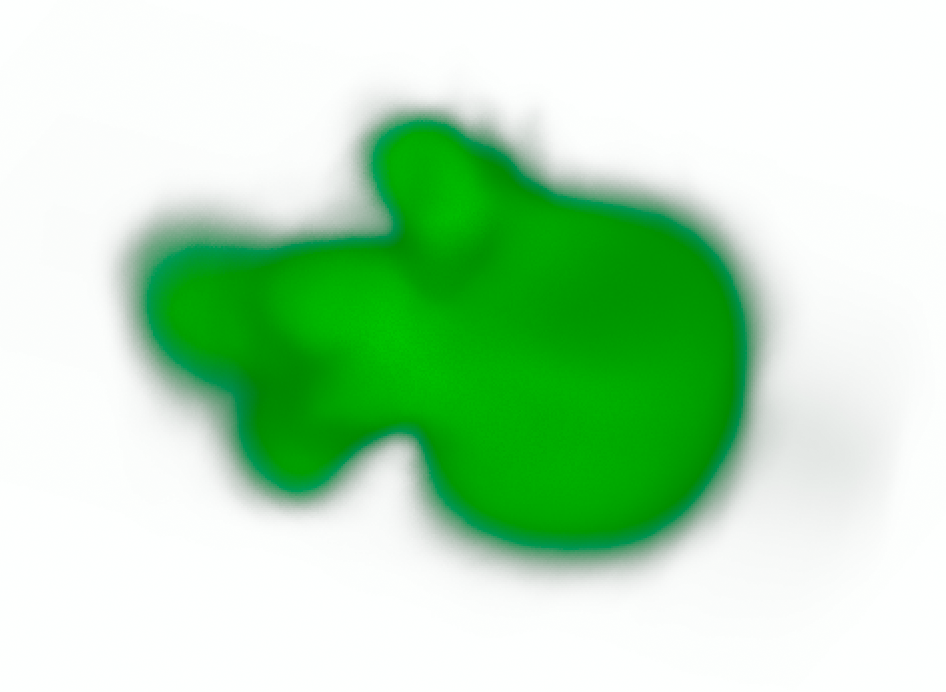}
        \caption{}
    \end{subfigure}
   
    \caption{Team \emph{Kirszenberg A.}'s contribution involving (a) detection of the spline passing through the vertebral column and (b) a sample template for L4 use for vertebra identification.}
	\label{fig:lrde}
\end{figure}

\noindent
\emph{Segmentation.} Three different U-Net models are trained in a `pseudo-3D' segmentation technique wherein, the 3D input is sliced into 3-voxel wide slices along the three axes. Prior to this, patches of size $80\times128\times128$ are extracted from the scan, resulting in sagittal, coronal, and axial slices of shapes $3\times123\times128$, $80\times3\times128$, and $80\times128\times3$, respectively. This step performs a binary segmentation of `spine vs. background'. The predicted masks of the three models are combined using majority voting and passed through a filtering operation for removal of stray segmentation and hole-filling (cf. Fig.~\ref{fig:lrde}a).

\noindent
\emph{Labelling.} This task is attempted as a combination of morphological operations and template matching, implemented as follows: 1. The predicted binary segmentation mask is blurred using a Gaussian kernel and skeletonised to obtain a skeleton of the vertebral column. Further clean-up is obtained by choosing the path connecting the voxels between two end-points using Dijkstra's algorithm. 2. The skeleton is then discretised into \siunit{1}{mm} distant points which are used as anchors for template matching. These templates were generated from the training data at a vertebra level by centring each vertebra at the centroid and averaging over a certain number of rotations as shown in Fig.~\ref{fig:lrde}b. For template matching, the five best {vertebrae, point} candidates are chosen and for every point its previous and next vertebrae are matched to the points before and after, respectively. Once no vertebrae can be matched, scores for each vertebrae are summed from each of the five vertebral columns and the one with the highest score is selected. Following this, each voxel of the column is labelled after the template with the highest score.

% -- Wang X. ----------------------------

\subsection*{\crule[008700]{0.25cm}{0.25cm} Wang X. et al.: Improved Btrfly Net and a residual U-Net for \textsc{VerSe}`19}
\label{desc:wangx}

Improved versions of Btrfly Net \citep{sekuboyina2018} and the U-Net \citep{ronneberger2015} are employed to address the tasks of labelling and segmentation, respectively. Of interest is the task-oriented pre- and post-processing employed in each task.

\noindent
\emph{Pre-processing.} A Single Shot MultiBox Detector (SSD) is implemented to localise the vertebrae in the sagittal and coronal projections and its predictions are used to crop the 3D scans. This is followed by re-sampling the crops to a \siunit{1}{mm} resolution and padding the projections to $610\times610$ pixels.

\noindent
\emph{Labelling.} The Btrfly Net is employed for this task with a major difference in the reconstruction of 3D coordinates from its 2D heatmap predictions. However, unlike obtaining the 3D coordinates from the outer product of the 2D channelled heat-maps followed by an \emph{argmax}, the authors propose an improved scheme resulting in a $4\%$ improvement of the identification rate. Specifically, 2D coordinates of the vertebra are obtained from the individual projections, denoted by $(x, z_s)$ from the sagittal and $(y, z_c)$ from the coronal heat maps. Notice the two variants of the $z$-coordinate. The final $z$-coordinate is then calculated as the weighted average of $z_s$ and $z_c$ with the maximum values of their corresponding heat maps as weights. Additionally, the missing predictions are \emph{filled in} with interpolation.

% \begin{wrapfigure}{r}{0.5\textwidth}
%   \centering
%   \includegraphics[width=0.45\textwidth]{figures/appendix/init.png}
%   \caption{Architecture of residual U-net employed by team \emph{Wang X.} for the segmentation task.}
% \label{fig:init}
%  %\vspace{-110pt} % This removes the white box on the second page
% \end{wrapfigure}

\begin{figure}%{\textwidth}
  \centering
  \includegraphics[width=0.6\textwidth]{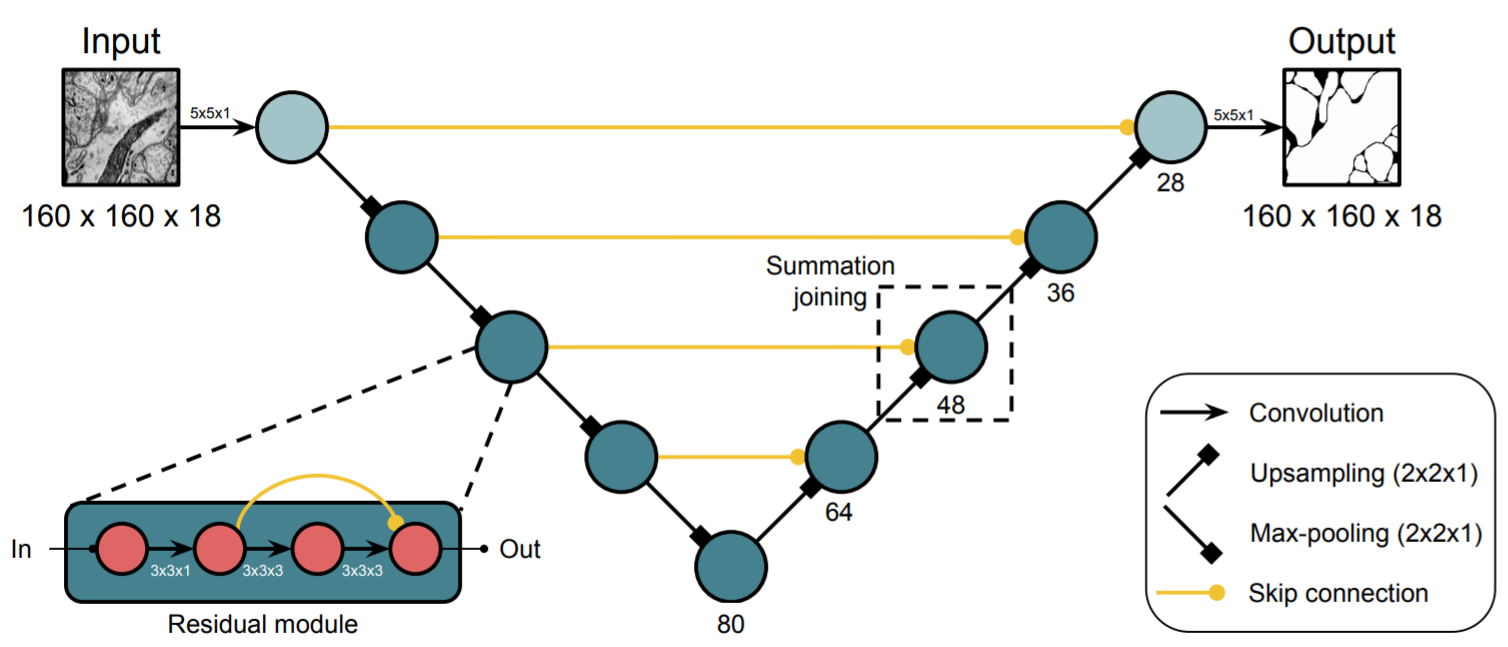}
  \caption{Architecture of residual U-net employed by team \emph{Wang X.} for the segmentation task.}
\label{fig:init}
 %\vspace{-110pt} % This removes the white box on the second page
\end{figure}

\noindent
\emph{Segmentation.} 
Since the vertebral centroids are now identified, the segmentation is tasked to segment one vertebra given its centroid position. For this, a 3D U-Net with residual blocks is chosen (Fig.~\ref{fig:init}). The network is trained with Dice loss and works with patches of size $96\times96\times96$ centred at the vertebral centroid in question. Once segmented, the vertebra is labelled according to its centroid's label and assigned back to the full scan. In case of a conflict, i.e: if a voxel labelled as $i$ is again labelled as $j$, the label with a higher logit is chosen.

% ================== VERSE 2020  ==================

% -- carpediem / Hou F ---------------------------- (malek)
\subsection*{\crule[00ffff]{0.25cm}{0.25cm}  Hou et al.: Fully Automatic Localisation and Segmentation of Vertebrae Based on Cascaded U-Nets }
\label{desc:houf}

\begin{figure*}%{\textwidth}
  \centering
  \includegraphics[width=0.8\textwidth]{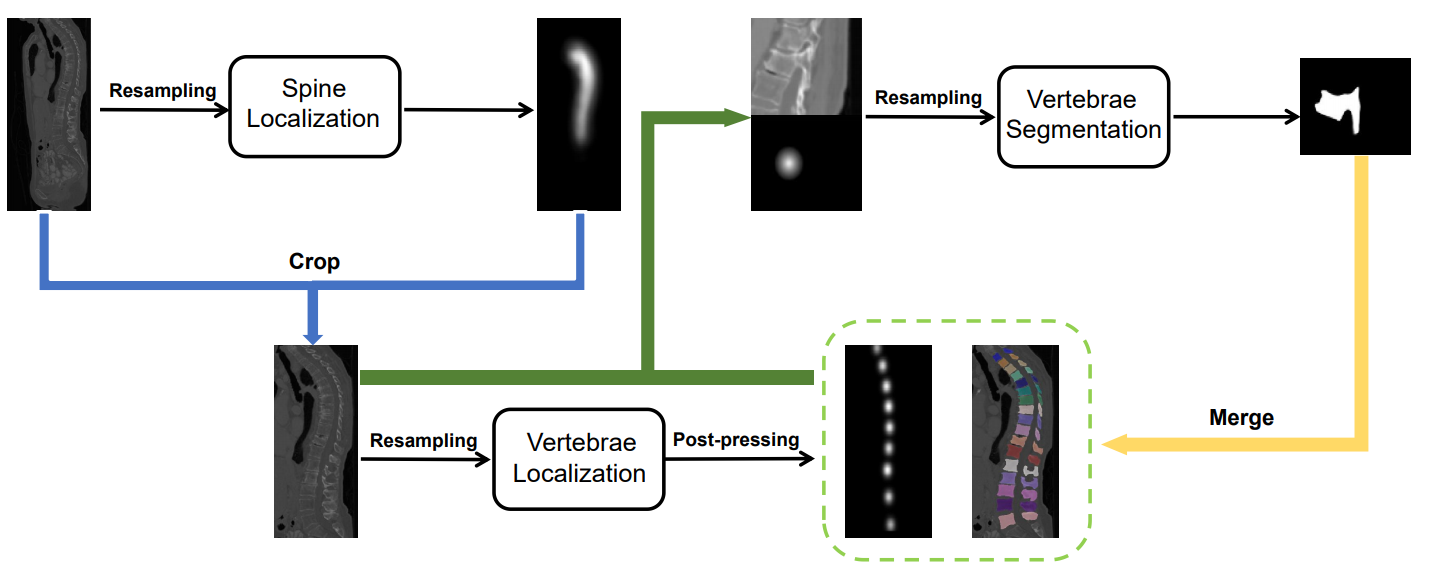}
  \caption{Hou et al. proposed pipeline}
\label{fig:houf}
 %\vspace{-110pt} % This removes the white box on the second page
\end{figure*}

The authors propose a multi-stage pipeline for vertebral localisation and segmentation based on a general U-net architecture. Firstly, the centre-line of the spine is inferred, and then the spine region is cropped to be fed as the input of the second stage. Accordingly, the second neural network predicts the centre coordinates and classes of all vertebrae. In the last stage, the segmentation network performs a binary segmentation of each of the cropped vertebrae. The full pipeline is illustrated in Fig.~\ref{fig:houf}

\noindent
\emph{Spine Localisation.} In the first stage, the authors use a variant of the U-Net \citep{ronneberger2015} to predict heat-maps that cover the whole spine. They set the filters of each convolutional layer to 64, which can significantly improve training speed while ensuring performance. The authors utilise the general $\ell_2$-loss to minimise the difference between the target and predicted heat-maps. As a pre-processing step, the CT images are sub-sampled to a uniform voxel spacing of 8mm, and then a patch size of 64$\times$64$\times$128 is fed into the network. The predicted coordinates of the centre of the spine help are used to crop the spine region as the input of the second stage.

\noindent
\emph{Vertebrae Localisation.} The authors deploy the general U-Net \citep{ronneberger2015} as a baseline. Both encoder and the decoder use five levels consisting of two convolution layers with a leaky-ReLU activation function. Due to the specific shape and fixed relative position of vertebrae, for most cases, the labels of the vertebrae are a continuous sequence despite their coordinates. It is important to localise and identify the first and the last vertebrae. The authors use a  weighted $\ell_2$-loss function to emphasise the contribution of the first and the last vertebrae in the loss. Similarly to the first stage, the CT images are re-sampled to uniform voxel spacing of 2mm, and then a patch size of 96 × 96 × 128 is fed into the network.

\noindent
\emph{Vertebrae Segmentation.} In this stage, the predicted coordinates of each vertebra are used to crop the individual vertebrae region. Similar to the localisation stage, the U-Net is used and the CT volumes are re-sampled to a uniform voxel spacing of \siunit{1}{mm}, the segmentation network with a patch size of 128 × 128 × 96 produces the individual predictions of each vertebra, and finally, the multi-label segmentation results are obtained by merging all binary segmentation results.

\noindent
\emph{Postprocessing.} Due to the partial vertebrae often in the top or bottom of volume, which has a bad influence on detecting the position of the first or last vertebrae, in this work, the landmark is abandoned if its distance from the top or the bottom of the volume is less than a threshold.

% -- poly / Huang Z ---------------------------- (malek)
\subsection*{\crule[00b0b0]{0.25cm}{0.25cm}  Huang et al.: A\textsuperscript{2}Unet: Attention and Aggregation UNet for Vertebrae Localisation and Segmentation }
\label{desc:huangz}
The authors formulate both tasks as a pixel-level prediction problem. Specifically, the landmark detection problem (task 1) is converted into a heat-map prediction format and the vertebrae segmentation problem (task 2) is converted into a multi-class semantic map prediction scheme. Both tasks generate full-scale outputs that enabled the authors to
utilise a U-net architecture \citep{ronneberger2015} to extract the features. In this work, the authors develop a new variation of 3D Unet, in which an attention and aggregation mechanisms are introduced to enhance the feature representation in both tasks. This new variant is called the A\textsuperscript{2}UNet.

\begin{figure*}%{\textwidth}
  \centering
  \includegraphics[width=0.9\textwidth]{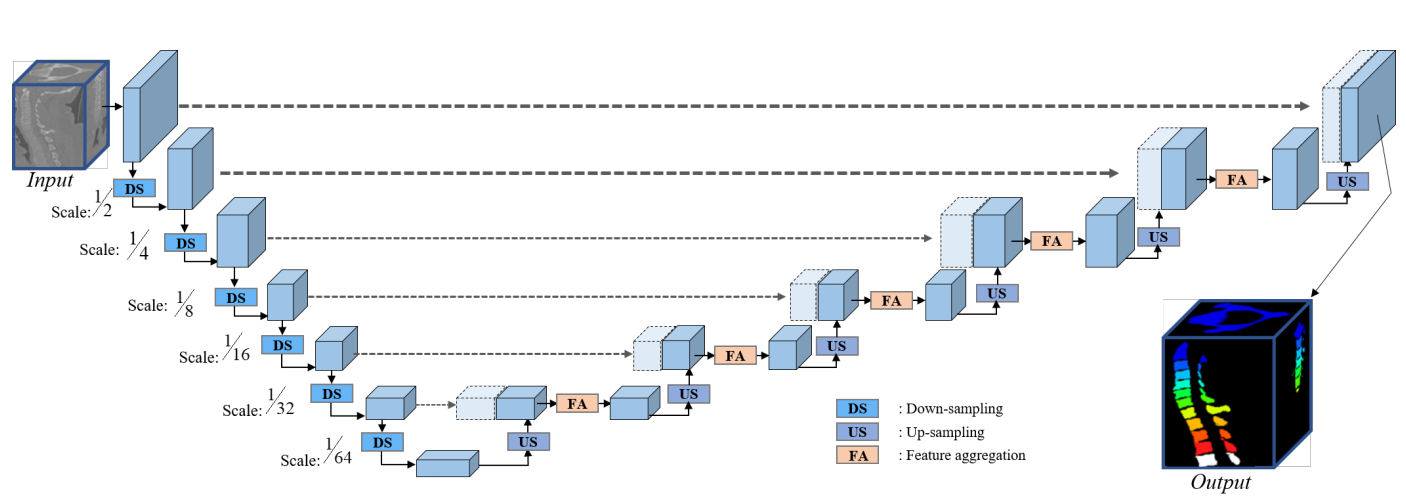}
  \caption{ A\textsuperscript{2}Unet's Architecture}
\label{fig:huangz}
 %\vspace{-110pt} % This removes the white box on the second page
\end{figure*}

\begin{figure*}%{\textwidth}
  \centering
  \includegraphics[width=0.6\textwidth]{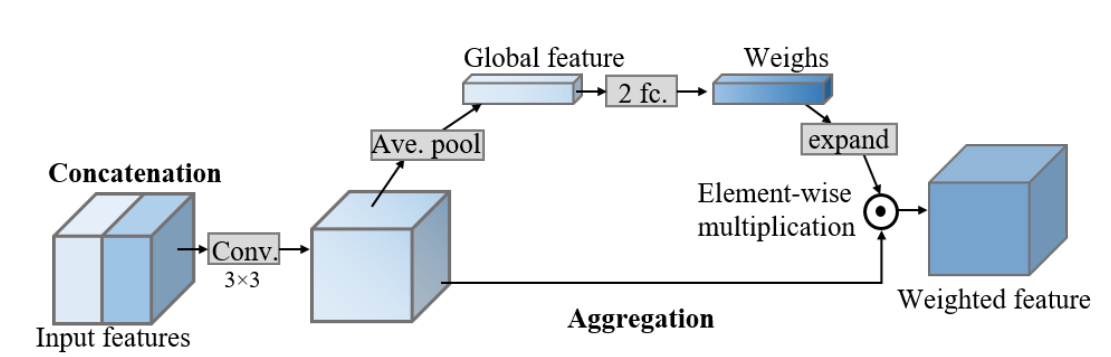}
  \caption{ Huang et al. Feature Aggregation block}
\label{fig:huangz1}
 %\vspace{-110pt} % This removes the white box on the second page
\end{figure*}

\noindent
\emph{Attention and Aggregation UNet (A\textsuperscript{2}UNet).} The proposed  A\textsuperscript{2}UNet, which is shown in Fig.~\ref{fig:huangz}, adopts the original U-Net structure that consists of a contracting path to downsample the inputs for global representation, and an expanding path to upsample the feature for detailed prediction. Several skip connections link the contracting and expanding paths that directly transfer the information from the shallow to deep layers. However, features from different convolution stages contain information from different semantic levels. In this work, the authors embed the efficient feature aggregation (FA) module shown in Fig.~\ref{fig:huangz1}, into the U-Net structure for channel-wise attention based on the Squeeze-and-Excitation (SE) block \citep{hu2019squeezeandexcitation}. It receives the two feature maps where one is from the contracting path, and the other is from the expanding path. The features are firstly sent to the average pooling process for global representation. Then, two fully connected layers are used to investigates the importance (weights) of different feature channels. By multiplying the weights to corresponding channels, the key features can be focused that will be used for the following process. 

\noindent
\emph{Heads for Vertebra Localisation and Segmentation.} The authors develop two sub-networks, or heads, to decode the backbone output into the feature format for each task. For the localisation task, a convolution layer is applied to generate a 26-channel output, each channel corresponds to one of 26 classes of the vertebra. Each channel is actually a heat map where the location information of a specific vertebra is encoded. To reason the vertebra location, the coordinates candidates are selected where the corresponding score in the heat-map is above 0:35. The final vertebra coordinates are determined by adopting the non-maximum suppression (NMS) algorithm towards those candidates in an adjacent vertebra region which has a distance between 12.5mm and
40mm. 

For the segmentation task, a convolutional layer is deployed to generate a single semantic map, each pixel contains 27 categorical value, indicating one of 26 anatomical classes or the background. The segmentation model is trained with Dice loss and CE loss. Since every voxel is classified only considering the channel score after obtaining the segmentation mask, outlier voxels that are not connected with the largest component will be removed.

% -- lrde / Hu`ynh L. D ---------------------------- (malek)
\subsection*{\crule[008787]{0.25cm}{0.25cm}  Hu\`{y}nh et al.: 3D Mask Retinanet for Vertebrae Instance Segmentation}
\label{desc:huynhl}

\begin{figure*}%{\textwidth}
  \centering
  \includegraphics[width=0.9\textwidth]{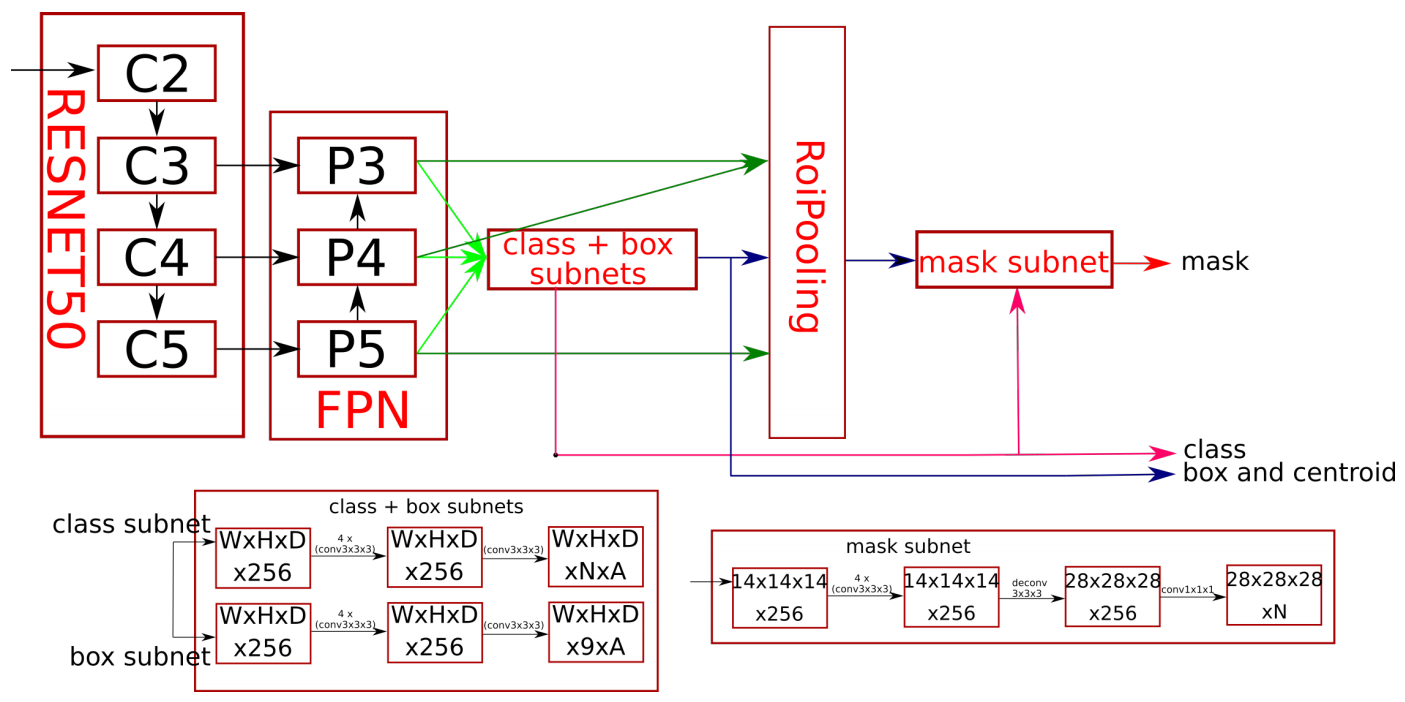}
  \caption{Mask RetinaNet's architecture, as employed by \emph{Hu\`{y}nh et al.}}
\label{fig:huynhl}
 %\vspace{-110pt} % This removes the white box on the second page
\end{figure*}

The authors propose a single model that performs both sub-tasks. A two-stage model is adopted inspired by Mask R-CNN \citep{he2018mask}. Mask R-CNN is a two-staged model, in which the first stage localises RoI while two sub-nets on the second stage classify and segment a subset of these RoIs. Since the Mask R-CNN is a heavy model, an extended version or Mask R-CNN for 3-D images will require significant memory, and as a result, it limits the number of RoIs that could be passed to the second stage. This problem makes the model more sensitive to class imbalance. For that reason, the authors propose a new two-stage model. They replace the first stage of Mask R-CNN with the Retinanet \citep{lin2018focal}. With this modification, the first stage is now responsible for both RoIs' localisation and classification. The first stage is more robust to class-imbalance than the original Mask R-CNN thanks to Focal Loss. This allows the authors to use a small, fully convolutional network on the second stage to performs the mask regression. Since only RoIs that contain objects will be passed through the second stage, training the model requires less memory. This model is called the Mask RetinaNet. Due to memory limitation, the authors are forced to train with a small batch size and they use Group Normalisation \citep{wu2018group} instead of Batch Normalisation in their network. The architecture of Mask RetinaNet is illustrated in Fig~\ref{fig:huynhl}

\noindent
\emph{The detector stage.} The authors adapt RetinaNet for 3D cases. Their version will also predict the object's centroid in addition to the axis-aligned bounding box (AABB). The backbone is constructed with a 3D version of the Resnet50 and Feature Pyramid Network \citep{lin2018focal}. For this dataset, the authors only use pyramid levels 3 to 5. They avoid level 2 because anchors defined on it are unnecessarily dense for this dataset, while anchors defined on levels greater than 5 are too sparse to distinguish nearby vertebrae. At each pyramid level, they use four anchors with two width/height/depth ratios of 1/1/0.625 and 1/0.74/0.42, and a width of 86 and 68 for level 3, 100 and 79 for level 4 and 120 and 94 for level 5. They are chosen by running a K-means clustering on the AABBs of training vertebrae similar to the algorithm described in \citep{redmon2016yolo9000} to ensure that for each vertebra, they could find at least one anchor so that the Intersection over Union (IoU) with its AABB is higher than 0.6. The classification and regression sub-nets are implemented as described in \citep{lin2018focal}. The classification subnet is responsible for the classification of anchors. It predicts a length $N$ one-hot classification vector for each anchor, with $N$ being the number of classes. The regression subnet performs AABB and centroid regression. For each positive anchor, it predicts a length nine regression vector, of which the first 6 encode the AABB (itscentre coordinate and size), and the last 3 encode the centroids' position. Instead of predicting these values directly, they adopt the coordinate parameterisations of \citep{ren2015faster} for their case.

\noindent
\emph{The mask regression stage.} To perform instance segmentation, the authors attach a second stage to their 3D-RetinaNet to output a binary mask for each RoIs detected by the first stage. This stage is implemented similar to Mask R-CNN: a 3D-ROIAlign layer extracts a fixed-size w $\times$ h $\times$ d feature map from the FPN for each RoI using trilinear interpolation, followed by simple fully convolutional networks. These subnets will produce D $\times$ 2w $\times$ 2h $\times$ 2d $\times$ N with D the number of detections provided by the first stage and N the number of classes.

% -- ubmi / Jakubicek R ---------------------------- (malek)
\subsection*{\crule[005555]{0.25cm}{0.25cm}  Jakubicek et al.: Approach for Vertebrae Localisation, Identification and Segmentation}
\label{desc:jakubicekr}

\begin{figure*}%{\textwidth}
  \centering
  \includegraphics[width=0.9\textwidth]{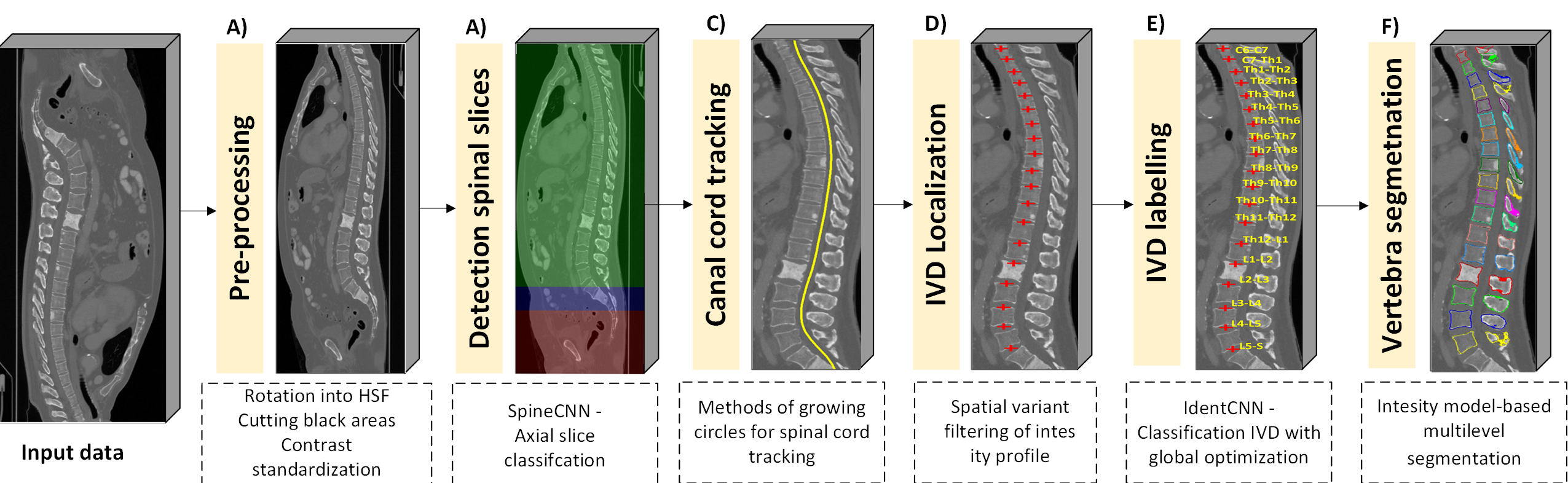}
  \caption{ An overview of the multi-stage framework proposed by \emph{Jakubicek R.}: Pre-processing, spinal slices detection, spinal canal tracking, inter-vertebral disc (IVD) localisation, IVD labelling, and vertebral segmentation.}
\label{fig:jakubicekr}
 %\vspace{-110pt} % This removes the white box on the second page
\end{figure*}

The authors propose a fully automatic multi-stage system as shown in Fig.~\ref{fig:jakubicekr}. Moreover, they provide an auxiliary semi-automatic mode that enables the inspection and possibly correction of automatically detected positions of the inter-vertebral discs (IVD) and their labels before the following segmentation step. Their approach combines modern deep-learning-based algorithms with more classical image and signal processing steps and with segmentation using the intensity vertebra models adaptation.

\noindent
\emph{Pre-processing.} The authors first attempt to cut the data from background and ``black" artefacts caused by geometrical shearing. The second step is the correction of the random rotation,which is not presented in the real CT data. For this purpose they provide the CTDeepRot algorithm \citep{jakubicek2020tool}, which predicts these rotational angles using a CNN and transforms the data into the standard Head First Supine (HFS) patient position.

\noindent
\emph{Detection of spinal cord centre-line}: First, each axial slice of the CT data is classified by a CNN into four categories (slices containing complex C1-2, slices with the main part of the spine from C3 to L6, slices containing the sacrum, and remaining areas feet, head, background). A pre-trained AlexNet \citep{krizhevsky2012imagenet} CNN is used for this purpose. In the slices containing the main part of the spine, the approximate position of a spinal canal is found \citep{simonyan2014very} architecture. Each detected centroid of a detected bounding box is taken as a potentially correct centre of the spinal canal in the appropriate axial slice. The whole spinal canal is then traced by the algorithm using the growing inscribed circles, where the detected centroids are taken as starting (seed) points of the tracing. The optimum spine centre-line is then chosen by the population-based optimisation process.

\noindent
\emph{Vertebra localisation and identification.} The spine CT data is geometrically transformed into the straightened data according to the spine centre-line curvature. In the straightened data, the centroids of the vertebral bodies are determined by morphological transforms, and the respective intensity profile along the z-axis is taken. This way, the obtained 1D signal is processed by an adaptive IIR (infinite impulse response) filter, which enables detection of the positions of the individual IVDs. Adaptation of the filter is controlled by a statistical model using knowledge about the anatomy of the spine. Finally, each detected IVD is classified into a category of the vertebral type (label) by a combination of a CNN (pre-trained Inception V3 \citep{guan2019deep}) and the dynamic programming optimisation. All used pre-trained CNN architectures were pre-trained on the ImageNet dataset \citep{russakovsky2015imagenet} and fine-tuned on the authors' database of CT image data.

\noindent
\emph{Vertebra segmentation.} The segmentation of the vertebrae is based on four-step vertebra intensity model registration. In the first step the mean model of the individual vertebra is scaled and deployed along the spine in accordance with the detected and labelled IVDs. The second step performs rigid registration of each vertebra, which aligns the model into an optimally precise position in the 3D CT data, followed by improvement via elastic registration of each vertebra. In the third step, the elastic registration is performed on the whole spine model, where the models fits the shapes of the vertebrae. In the last step, the final segmentation contours are slightly refined and smoothed by the graph-cut based algorithm. Elastix v.5.0.0 \citep{klein2009elastix,shamonin2014fast} is sued as the registration software.

% -- htic / Mulay S ---------------------------- (anjany)
\subsection*{\crule[b0b0ff]{0.25cm}{0.25cm} Supriti M. et al.: Vertebrae localisation and Segmentation using Mask-RCNN with Complete-IoU Loss}
\label{desc:supritim}
The authors propose to segment vertebrae using Mask-RCNN trained with Complete-IoU (CIoU) loss. The spine vertebrae segmentation process contains the following pipeline: 3D to 2D conversion, pre-processing, Mask-RCNN feature extraction with Complete IoU loss for geometric factor enhancement \citep{frosio2018statistical}, and 2D to 3D back conversion.

\noindent
\emph{3D to 2D conversion.} Reorientation of the image is done with flips and reordering the image data array so that the axes match the directions indicated in orientation required for spinal vertebrae segmentation. Reoriented images are resampled to get the balance between image smoothness and identify fine image details.

\begin{figure}%{\textwidth}
  \centering
  \includegraphics[width=0.8\textwidth]{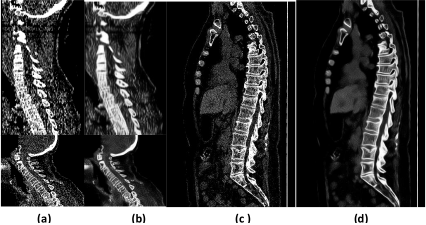}
  \caption{Enhancement of CT slices using the filtering algorithm proposed by \cite{frosio2018statistical} employed by \emph{Mulay S.}}
\label{fig:mulays}
 %\vspace{-110pt} % This removes the white box on the second page
\end{figure}

\noindent
\emph{Preprocessing.} CT images reconstructed from low-dose acquisitions may be severely degraded with noise and streak artefacts due to quantum noise, or with view-aliasing artefacts due to insufficient angular sampling. To improve CT image quality median filter along with non-local means (NLM) with Statistical Nearest Neighbors(SNN) by Frosio et al. \citep{frosio2018statistical} filtering algorithm is applied. Sampling neighbors with the nearest neighbour approach introduces a bias in the denoised patch which improves the CT image quality significantly. Fig.~\ref{fig:mulays}(a) and (c) shows the original slice of a spine CT while (b) and (d) shows the enhanced images.

\noindent
\emph{Segmentation using Mask-RCNN with CIoU loss.} Mask R-CNN predicts bounding boxes and corresponding object classes for each of the proposed region obtained using a backbone. Following this, a binary mask classifier generates a mask for every class. Bounding box regression is sometimes inaccurate due to overlapping areas. So a complete IoU (CIoU) loss is added in Mask R-CNN. 

A good loss for bounding box regression should consider three important geometric factors, i.e. overlap area, central point distance and aspect ratio. Zheng et al. \citep{zheng2020distance} proposed CIoU loss based on these requirements. The authors use an end-to-end pretrained Mask R-CNN-based detectron with CIoU loss model with Resnet x-152 backbone. An existing open-source implementation\footnote{\url{https://github.com/Zzh-tju/DIoU-pytorch-detectron}} using Pytorch is chosen. 

Once the scan is segmented slice-wise in 2D, the final segmentation is obtained by stacking the predicted masks and reorienting and resampling it back to the original image particulars.

% -- superpod / Netherton T. J ---------------------------- (anjany)
\subsection*{\crule[8484ff]{0.25cm}{0.25cm} Netherton T. et al.: A Multi-view Localisation and Deeply Supervised Segmentation Framework}
\label{desc:nethertont}
\begin{figure*}%{\textwidth}
  \centering
  \includegraphics[width=0.99\textwidth]{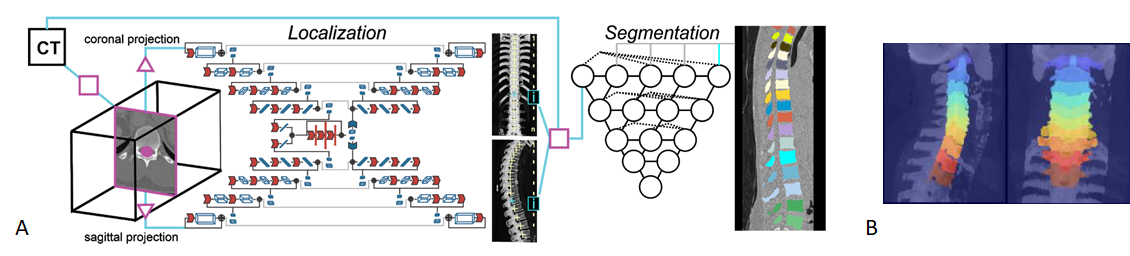}
  \caption{(A) An overview of the three-stage framework proposed by \emph{Netherton T.}: Sinal canal segmentation, localisation, and segmentation. (B) Ground truth sagittal and coronal intensity projection image pairs used in the training of the second stage. Each colored planar projection is housed in a separate channel. Centroids of each colored mask provide coordinates used in subsequent stages in this approach.}
\label{fig:nethertont}
 %\vspace{-110pt} % This removes the white box on the second page
\end{figure*}

The authors propose a framework that combines the use of a set of individual CNNs to accomplish 1) course spinal canal segmentation, 2) spine localisation via a multi-view network, and 3) automatic segmentation of individual vertebrae using a deeply supervised approach. A detailed description of steps (1) and (2) of the approach can be found in \cite{netherton2020evaluation}. Refer to Fig.~\ref{fig:nethertont} for an overview of the proposed approach.

\noindent
\emph{Data.} Data from \textsc{VerSe} 2019 and 2020 was used to train localisation and segmentation CNNs. All images and segmentations were resampled to have isotropic voxel sizes (1.0mm$^3$) and set to a common orientation. Ground truth localisation coordinates were not used in this approach. In total, 160 pairs of CT scans and segmentations were obtained and split into five groups for cross-validation.

\noindent
\emph{Spinal canal segmentation.} First, the spinal canal is segmented via a 2-dimensional FCN-8s with batch normalisation on the axial CT slices. Pairs of intensity projection images (sagittal and coronal) are then generated about a volume of interest (cropped from the CT scan) surrounding the spinal canal. These image pairs provide the network with sagittal and coronal views of the vertebral column and have a fixed width but variable length $l$ (where $l$ is the length of the CT scan); their corresponding ground-truth labels are then assigned individual channels (27 total) to account for each vertebral level. For the training stage, planar segmentation masks posterior to the spinal canal are removed to produce modified vertebral coordinates. In order to provide a large number of image augmentations, intensity projection image pairs (and corresponding ground truth masks) are incrementally cropped from the superior-inferior, inferior-superior, and medial-lateral directions.

\noindent
\emph{Multi-view spinal localisation.} X-Net \citep{netherton2020evaluation}, the localisation architecture, inputs the sagittal and coronal intensity projection pairs and outputs labeled, multi-dimensional sagittal and coronal arrays of individual vertebral column segmentations. X-Net, inspired by \cite{sekuboyina2020} and \cite{milletari2016v}, incorporates residual connections, pReLU activations, and is end-to-end trainable. By combining centre-of-mass coordinates from sagittal and coronal planar segmentations, 3-dimensional locations are obtained for each vertebral body. During training, the loss function, which incorporated the soft-Dice loss and cross-entropy loss, was applied to each view (i.e. coronal and sagittal, $L = L_s + L_c$). Augmentations were applied during training with a frequency of 0.7; coronal arrays were flipped left-right with a frequency of 0.5. Training was performed on a 16GB NVIDIA-V100 with batch size 8. Each model was trained for at least 26,000 iterations using early stopping.

\noindent
\emph{Deeply supervised vertebral body segmentation.} To perform vertebral body segmentation, a UNet++ architecture using skip connections, multi-class structure, and deep supervision is designed based on work by \citep{zhou2019unet++} . Using ground truth images and segmentations, three channel arrays
are formed for each vertebral level by cropping around the centre of mass of each vertebral level. Separate channels contained background, adjacent vertebral levels, and the central vertebral level, respectively. For each 3-dimensional coordinate (from the second stage), the CT scan is cropped to form separate volumes of interest (120$\times$96$\times$96mm$^3$). The top two most supervised outputs from each prediction are averaged to yield the vertebral body of interest.

% \noindent
% \textbf{Inference pipeline}
% During the test stage, the median coordinates across all five (cross-validated) X-Net models are passed to the segmentation stage. Unet++ output array from each of the five models used only the most supervised segmentations; predictions were also averaged across all five models from the cross validation split and thresholded ( > 0.5). The final volume mask was then returned to the original image space and saved for analysis.

% -- rigg / Paetzold J ---------------------------- (anjany)
\subsection*{\crule[4949ff]{0.25cm}{0.25cm} Paetzold J. et al.: A 2D-UNet on the \text{VerSe} data}
\label{desc:paetzoldj}
The authors implement a 2-D segmentation architecture for slices of the sagittal orientation of the 3-D dataset using a 2D U-Net \citep{ronneberger2015}. The encoder is made of a ResNet-34 backbone pre-trained on the ImageNet. The network is trained by optimising an equally weighted sum of the Dice loss and the binary cross-entropy loss (BCE) with data augmentations such as flipping, rotation, scaling, and shifting. The images are centre-cropped to 512 by 512 pixels to account for the irregular image sizes during training. All networks are implemented in Pytorch using the Adam optimiser and are trained for 1000 epochs. After prediction, the 2D slices are stacked together to reconstruct the 3D volume. The training was carried out on an NVIDIA QUADRO RTX 8000 GPU with a batch size of 52.

% -- fakereal / Xiangshang Z ---------------------------- (anjany)
\subsection*{\crule[ff00ff]{0.25cm}{0.25cm} Xiangshang Z. et al.: Vertebra Labelling and Segmentation using the Btrfly-net and the nnU-net}
\label{desc:xiangshangz}
The authors design an improved Btrfly Network \citep{sekuboyina2018} to detect the key points of the vertebrae and then build an nn-Unet \citep{isensee2019} to segment the vertebral regions. Both the labelling and segmentation tasks are handled independently.

\noindent
\emph{Vertebra Labelling.} Similar to \cite{sekuboyina2018}, the authors work with 2D sagittal and coronal MIP. Improving on it, changes were made to the model architecture and the training procedure. Two convolution layers for each layer of encoder and decoder in the network followed by batch-normalisation and ReLU non-linearity after each convolution layer. Kaiming-initialisation is used for the network parameters. In terms of data-based enhancements, the authors use horizontal and vertical flip for augmentation and with normalisation.

\noindent
\emph{Vertebra segmentation.} The preprocessing and training procedure of the nnU-Net is retained. On top of it, data augmentation is applied on the fly during training using the batch-generators framework \citep{isensee2020batchgenerators}. Specifically, elastic deformations, random scaling, and random rotations are used. If the data is anisotropic, the spatial transformations are applied in-plane as 2D transformations. Once trained, cases are predicted using a sliding window approach with half the patch size overlap between predictions.

% -- sitp / Yeah T ---------------------------- (bran)
 \subsection*{\crule[b000b0]{0.25cm}{0.25cm} Yeah T. \emph{et al.}: A Coarse-to-Fine Two-stage Framework for Vertebra Labeling and Segmentation.}
\label{desc:TimyoasYeah}
The author propose a two-stage network to achieve vertebra labeling and segmentation. Firstly, the low-resolution net determines the rough target location from downsampled CT images. Secondly, by feeding the first stage’s prediction results (upsampling before feeding) and high-resolution CT scans into a full resolution net, more accurate vertebra classification and segmentation are achieved. Considering the competition among different vertebra classes especially for adjacent vertebra, finally connected component analysis is applied to refine vertebrae segmentation results.
 
 The two-stage cascaded segmentation pipeline consists of two steps. Firstly a coarse location of spine RoI is obtained based on a lightweight low-resolution 3D U-Net from 3D CT scans with low resolution. Secondly the RoI and the accurate segmentation results are performed with a high-resolution 3D U-Net. Finally some post-processing methods are adopted to fill the holes inside each vertebrae and rule-based methods to recalibrate the vertebrae label. Both the low-resolution network and the high-resolution network have 26 output channels (C1-C7, T1-T13, L1-L6). 
 
The first stage preprocesses the training 3D CT scans to a larger spacing through downsampling and train the low-resolution 3D U-Net model with a patch size of 224$\times$128$\times$96. The second stage preprocesses 3D CT images to smaller spacing through upsampling and crops the RoI of spine regions as the training dataset for a  high-resolution U-Net model with a patch size of 256$\times$96$\times$80.

\noindent
\emph{Preprocessing and Augmentation.}
All input images are normalised zero mean and unit standard deviation (based on foreground voxels only). The data augmentation include elastic deformation, rotation transformation, gamma transformation, random cropping, etc.

\noindent
\emph{Loss and Optimisation.}
The low-resolution model with a classical combination of Dice loss and cross-entropy loss, while training the high-resolution model with a dynamic hybrid loss combining Dice loss and \emph{weighted} cross-entropy loss. A model with a dynamic hybrid loss combining Dice loss and \emph{Adam} optimiser with an initial learning rate of 10$^{-4}$ was used. During training, an exponential moving average of the validation and training losses is used. Whenever the training loss does not improve within the last 30 epochs, the learning rate is reduced by factor 5. The training is terminated automatically if validation loss does not improve within the last 50 epochs.
 
% -- aply / Zeng C. ---------------------------- (bran)

\subsection*{\crule[870087]{0.25cm}{0.25cm} Zeng C.: Two-stage Keypoint Location Pipeline for Vertebrae Location and Segmentation.}
\label{desc:ChanZeng}
The author proposes a two-stage keypoint detection pipeline for vertebral labeling based on the scheme of \cite{payer2019integrating} which uses Spatial-Configuration-Net and U-Net in \textsc{VerSe}`19 described in Section \ref{desc:payerc}.  

\noindent
\emph{Additional Data and Preprocessing.}
An additional 13 data sets from the \textsc{VerSe}`19 training set are used. The data is first preprocessed to the RAI direction. Data augmentation includes rotation, intensity shift, scaling and elastic deformation. The model is trained with all 113 cases.

\noindent
\emph{Localisation.}
To localize centres of the vertebrae, five keypoints location and global vertebrae location is performed separately. For the five keypoints, which contains the first and last two vertebral masses of the cervical spine, thoracic spine and lumbar spine, a network is designed of which the backbone is an HRNet \citep{sun2019deep} to regress the five keypoint heatmaps. The significance of the first stage is for better identification of several vertebral masses with obvious characteristics. The second stage follows \cite{payer2019integrating}, with a re-designed channel attention block in the network with a weighted loss function.

\noindent
\emph{Vertebrae Segmentation.}
For vertebrae segmentation, a binary segmentation network is trained based on the outcome of the labelling stage. A U-Net with an inputs size of 128$\times$128$\times$64 is used. The loss function is a mixture of Dice loss and binary cross-entropy loss.

% -- jdlu / Zhang A. ---------------------------- (bran)
\subsection*{\crule[550055]{0.25cm}{0.25cm} Zhang A. \emph{et al.}: A Segmentation-Based Framework for Vertebrae Localisation and Segmentation.}
\label{desc:AmberZhang}
In general, the vertebrae localisation and segmentation tasks are performed in a four-step approach: 1) spine localisation to obtain the region of interest, 2) single-class key point localisation to obtain the potential vertebrae candidates, 3) a triple-class vertebrae segmentation to obtain the individual mask and main category of each vertebrae, and 4) rule-based post-processing. 

A variant of V-Net with a mixture of Dice and binary cross-entropy loss is utilised in the first three steps and only a few hyperparameters are changed in each step, such as input/output shape, depth, width, etc. Step 2 and step 3 could be corrected by each other in an `intertwined' way as mentioned above: the proposed key-point candidates are used as input for step 3 to specify the vertebra to be segmented if the resulting segmentation result does not seem to be a mask (the volume is not large enough), then the proposed key-point can be regarded as a false positive.

\noindent
\emph{Spine Localisation.}
To obtain the spinal centerline, a variant of V-Net is used to regress a heatmap of the spinal centerline. The input is a 4-time downsampled single-channel 3D-patch with a size of 64$\times$64$\times$64). The sliding window approach is applied to serve the network with the specific size of local cubes. The heatmaps are generated using a Gaussian kernel by a kernel size (5, 5, 5) and sigma (6, 6, 6) on the downsampled mask to keep unique 3D connected domain. The output heatmap is converted to a binary mask by a threshold of 0.4 and resampled back to the origin image scale for later use.

\noindent
\emph{Keypoint Localization.}
A similar variant of V-Net is employed to regress a heatmap of the spine. The input in this step is a single-channel 3d-patch with the size of 64$\times$128$\times$128. Sliding window approach is applied as above. The heatmaps are generated by kernel size (7, 9, 9) and sigma (6, 6, 6) on the original scale based on the JSON label to ensure they are independent and disconnected. The proposed regression results are converted to a binary mask by a threshold of 0.4 and the centroid is calculated for each cluster.

\noindent
\emph{Vertebrae Segmentation.}
Considering half of the vertebrae account for a lack of samples for a 26-class classification, a triple-class segmentation task is defined to segment three categorises: `cspine', `tspine' or `lspine' for each vertebra in this step. A variant of V-Net is employed. The input is a cropped 3D patch around the localised centroid obtained from step 2.

\noindent
\emph{Rule-based Post-processing.}
In this step a simple post-preprocessing logic is applied to create the final multi-label result. If more than one category of vertebrae is found in one case, the two or four `split points' which is C7-T1 and T12-L1 can be localised. Then the others can be deduced based on these split points. If split points are not found in one case, then `cspine' vertebrae are deduced from C1 to the bottom and `lspine' ones are deduced from the bottom to the top.

\end{document}